\documentclass[sigconf, nonacm]{acmart}

\AtBeginDocument{%
  \providecommand\BibTeX{{%
    \normalfont B\kern-0.5em{\scshape i\kern-0.25em b}\kern-0.8em\TeX}}}

\setcopyright{acmcopyright}
\copyrightyear{2018}
\acmYear{2018}
\acmDOI{10.1145/1122445.1122456}

\usepackage{enumitem}
\usepackage{bm}
\usepackage{subcaption}
\usepackage[htt]{hyphenat}

\DeclareMathOperator*{\argmin}{arg\,min}

\begin{document}

\title{Importance of Tuning Hyperparameters of Machine Learning Algorithms}


\author{Hilde J.P. Weerts}
\authornote{Work done while intern at Columbia University in 2018.}
\affiliation{%
  \institution{Eindhoven University of Technology}
  \country{The Netherlands}}

\author{Andreas C. M\"{u}ller}
\affiliation{%
  \institution{Columbia University}
  \country{New York, U.S.A.}}

\author{Joaquin Vanschoren}
\affiliation{%
  \institution{Eindhoven University of Technology} 
  \country{The Netherlands}}

\renewcommand{\shortauthors}{Weerts et al.}

\begin{abstract}
  The performance of many machine learning algorithms depends on their hyperparameter settings. The goal of this study is to determine whether it is important to tune a hyperparameter or whether it can be safely set to a default value. We present a methodology to determine the importance of tuning a hyperparameter based on a non-inferiority test and \textit{tuning risk}: the performance loss that is incurred when a hyperparameter is not tuned, but set to a default value. Because our methods require the notion of a default parameter, we present a simple procedure that can be used to determine reasonable default parameters. We apply our methods in a benchmark study using 59 datasets from OpenML. Our results show that leaving particular hyperparameters at their default value is non-inferior to tuning these hyperparameters. In some cases, leaving the hyperparameter at its default value even outperforms tuning it using a search procedure with a limited number of iterations.
\end{abstract}




\settopmatter{printfolios=true}
\maketitle

\section{Introduction}
Many parameters of machine learning algorithms are derived through training. Additionally, most modern machine learning algorithms have parameters that need to be fixed before running them. Such parameters are referred to as \textit{hyperparameters} \citep{Vanrijn2018, Probst2018, Mantovani2015, Lavesson2006}. In many cases, the performance of an algorithm on a given learning task depends on its hyperparameter settings. In order to obtain the best performance, machine learning practitioners can \textit{tune} the hyperparameters. Methods for tuning hyperparameters for the problem at hand require the definition of a search space: the set of hyperparameters and ranges that need to be considered. Additionally, one needs to define the search heuristic that is used to search for good hyperparameters within the search space. Tuning is generally a computationally expensive procedure that becomes more expensive as the search space increases. Currently, there does not exist empirical evidence on which hyperparameters are important to tune and which hyperparameters result in similar performance when set to a reasonable default value. Hyperparameters that fall into the latter category might be eliminated from the search space entirely when computational resources are limited. In this work we study the importance of tuning hyperparameters and aim to provide empirical evidence of hyperparameters that might be eliminated from the search space.

We present a methodology to determine which hyperparameters are important to tune, based on empirical performance data derived from experiments across multiple datasets. Because our methodology requires the notion of a default parameter setting, we introduce a simple procedure to deduce reasonable default parameters from performance data. We apply our approach in a benchmark study of two popular classification algorithms: random forests and support vector machines. In particular, we analyze the importance of hyperparameter tuning based on the performance of these algorithms across 59 datasets that were taken from the OpenML-CC18 benchmark suite.

Our results suggest that leaving particular hyperparameters at our computed default value leads to non-inferior performance. Moreover, given a random search approach with a limited number of iterations, fixing the hyperparameter to our default value sometimes outperforms tuning all hyperparameters at once. Machine learning practitioners might want to leave these hyperparameters at their default value, rather than invoking an expensive random search procedure. For other hyperparameters, we observed a difference between the fixed and non-fixed conditions. This indicates that these hyperparameters should not be left at our computed default value.

The remainder of this work is structured as follows. In Section~\ref{sec:relwork}, we provide a short overview of related work. In Section~\ref{sec:methods}, we present our default parameter determination procedure as well as a methodology for determining the importance of tuning. Section~\ref{sec:exp} covers details of our benchmark experiments, including the used datasets, algorithms, hyperparameters, performance measures, and search strategy. In Section~\ref{sec:results}, the results of the experiments are presented. Finally, the conclusions and limitations are discussed in Section~\ref{sec:concl}.

\section{Related Work}
\label{sec:relwork}

\subsection{Hyperparameter Default Values}
In the work of \citet{Probst2018}, default values are determined using surrogate models as follows. First, for each of the 38 binary classification tasks in the study, a surrogate model that predicts the performance of the algorithm based on hyperparameter settings is learned from empirical performance data. Second, a large number of hyperparameter configurations are sampled and their performance is estimated using the surrogate models. Finally, the default parameters are determined by minimizing the average risk. In contrast to this method, we will use a simple, intuitive heuristic to determine default parameters directly from the performance data.

\subsection{Hyperparameter Importance}
The importance of hyperparameter tuning has been studied in the work of \citet{Mantovani2015}. The authors use properties of the learning task, known as meta-features, to predict whether hyperparameter tuning will lead to better results than models obtained using default parameter settings. Additionally, \citet{Lavesson2006} show that tuning hyperparameters is often more important than the choice of the machine learning algorithm. In contrast to these works, we study the importance of tuning particular hyperparameters, rather than the importance of tuning at all.

\citet{Vanrijn2018} study the importance of particular hyperparameters as well as interactions between hyperparameters. The authors define important hyperparameters as parameters that explain most variance in performance across multiple datasets. The explained variance is determined through a functional ANOVA framework. A limitation of this approach is that the results do not directly translate into guidelines on which particular hyperparameters are important to tune. For example, a hyperparameter for which one specific setting always performs well is considered important, as it explains variance in performance. However, such a parameter does not require extensive tuning. Instead, it can simply be set to the setting that always results in good performance.

This issue is alleviated by \citet{Probst2018}, who introduce the \textit{tunability} of a hyperparameter: the performance gain that can be achieved by tuning the hyperparameter. The authors compare the performance of leaving all hyperparameters at their default values to tuning the hyperparameter of interest while leaving all other hyperparameters at their defaults. Their approach can be used directly to determine the importance of tuning a hyperparameter. 

The approach taken in this work is similar to the work of \citet{Probst2018}. As opposed to the aforementioned work, we are interested in the \textit{tuning risk} of a hyperparameter: the performance loss that is incurred when a hyperparameter is set to a default value instead of being tuned, while all other hyperparameters are tuned. \citet{Probst2018} describe the method used in this work and mention that ``the two alternatives can be seen as similar to forward and backward selection of variables in stepwise regression'' (p. 7). The difference between the two approaches is reflected in the way default parameters are chosen. The default parameters determined by \citet{Probst2018} are chosen in a joint fashion, while the default parameters used in this work are determined univariately. That is, \citet{Probst2018} determine a specific set of default hyperparameter values at the same time, whereas we determine the default setting of a hyperparameter irrespective of the values of other hyperparameters. Consequently, the approach of \citet{Probst2018} takes into account possible interaction effects between hyperparameters. This is beneficial in their forward selection approach, as most of the hyperparameters are left at their default value. In our backward selection approach, however, potential interaction effects are optimized by tuning the interacting hyperparameter, allowing us to choose the default parameters using a univariate approach. In future research, these interaction effects can be investigated further.

\section{Methods}
\label{sec:methods}

\subsection{Notation}
Let $A$ be an algorithm with $H$ hyperparameters with domains $\Theta_1, ..., \Theta_h$ and configuration space $\bm{\Theta} = \Theta_1 \times ... \times \Theta_h$. Let $\bm{\theta} = \langle \theta_1, ..., \theta_h \rangle$ with $\theta_i \in \Theta_i$ denote a setting of all hyperparameters of $A$.

The performance of an algorithm with hyperparameter settings $\bm{\theta}$ on a specific learning task can be expressed as a loss function $L(Y, \hat{f}(X,\bm{\theta})$. Here, $Y$ is the target variable, $X$ a vector of features and $f(.,\bm{\theta})$ the predictive model resulting from training algorithm $A$ with hyperparameter settings $\bm{\theta}$. For example, classification accuracy can be expressed as the mean of a 0/1 loss function defined as follows:
\begin{equation}
   L(y, \hat{f}(x,\bm{\theta})) =\begin{cases}
    0,& \text{if } y = \hat{f}(x,\bm{\theta})\\
    1,              & \text{otherwise}
\end{cases}
\end{equation}
Usually, one is interested in estimating the expected loss \citep{Probst2018}. Let Let $R_j(\bm{\theta})$ denote the expected loss (risk) of a model trained with hyperparameter settings $\bm{\theta}$ on dataset $j$. Let $\theta_i^{j}$ denote the default setting of hyperparameter $i$ for dataset $j$. Moreover, let $R_j(\theta_i^j)$ denote the risk for dataset $j$ of fixing hyperparameter $i$ to $\theta_i^j$ while tuning all other $H -1$ hyperparameters. Finally, let $\theta_i^{*j}$ denote the best setting for parameter $i$ for dataset $j$, i.e.:

\begin{equation}
\theta_i^{*j} = \argmin_{\theta_i \in \Theta_i} R_j(\theta_i)
\end{equation}

\noindent Given an algorithm $A$ and hyperparameter $i$, we aim to determine 1) a reasonable default setting $\theta_i^{j}$, for $i = 1, ..., h$ and $j = 1, ..., m$, and 2) whether $R_j(\theta_i^{j})$ is non-inferior to $R_j(\theta_i^{*j})$ across datasets.

\subsection{Hyperparameter Default Values}
The default setting problem can be formalized as follows. Given
\begin{itemize}
    \item an algorithm $A$ with configuration space $\bm{\Theta}$
    \item a hyperparameter $i$ with domain $\Theta_i$
    \item $M$ datasets $\mathcal{D}^{(1)}, ..., \mathcal{D}^{(m)}$
    \item for each dataset $j$, a set of $K$ empirical risk measurements with different hyperparameter settings $\langle R_j(\bm{\theta}_k) \rangle_{k=1}^{K}$, where $\bm{\theta}_k \in \bm{\Theta}$
\end{itemize}
we aim to determine a default setting $\theta_i^j$ for hyperparameter $i$, for dataset $j$. 

\subsubsection{Approach}
We use a simple heuristic procedure to determine default values. The intuition behind our method is to find a setting that, across many datasets, most often results in good performance.

It is important to note that if we were to use a single default value in the hyperparameter importance experiment, we leak information on the best hyperparameter value of a specific dataset. Therefore, we determine a default setting using a leave-one-out approach. That is, we determine $\theta_i^{j}$ using only performance data related to the other $M - 1$ datasets.

First, we take a subset of all empirical performance measurements that represent good performance. That is, for each dataset, we pick the $n$ hyperparameter settings that resulted in the best performance, i.e. lowest risk. 
Note that the top $n$ performance data consists of $M \cdot n$ empirical performance measurements. This is similar to the first step of the prior distribution estimation method proposed by \citet{Vanrijn2018}. The choice of $n$ is potentially important. On the one hand, picking $n$ too small might give a large weight to outliers. On the other hand, picking $n$ too large could result in the inclusion of bad performance measurements in the subset.

Second, we determine which parameter setting occurs most frequently within this subset. The intuition is that a setting that worked well on most datasets is a good candidate for a default value. Moreover, if there exists a single setting that works best for most datasets, our approach is certain to find it. Note that this would not be the case when using, for example, the median. For hyperparameters with a nominal or boolean domain, the setting that worked well most often is simply the mode. Since hyperparameter settings are picked randomly across the domain, parameters with a continuous domain or large integer domain, e.g. more than 50 possible values, must first be discretized. In this work, we will determine the bin size $b$ using the Freedman-Diaconis rule \citep{Freedman1981}, presented in Equation~\ref{eq:fd}.

\begin{equation}
\label{eq:fd}
b = 2 \frac{IQR}{\sqrt[3]{N}}
\end{equation}
where IQR is the interquartile range of the data and $N$ is the number of observations. The Freedman-Diaconis rule is known to be resilient to outliers. An alternative of this rule is Sturges formula \citep{Sturges1926}. The automatic histogram binning, as implemented in the python library \texttt{numpy} \citep{oliphant2006guide}, uses the maximum of the Freedman-Diaconis rule and Sturges formula. However, Sturges formula assumes that the data follows a Gaussian distribution. Since we have no reason to assume that this holds for our data, we only use the Freedman-Diaconis rule.

The advantage of our method is that it is very simple and intuitive. In future work, it would be interesting to compare our simple, univariate heuristic with more sophisticated methods, such as default parameter estimation through surrogate models \citep{Probst2018}.

\subsubsection{Meta-feature dependent default values}
In the open source machine learning library scikit-learn \citep{sklearn}, the default values of some hyperparameters depend on properties (or meta-features) of the dataset. In particular, \texttt{max\_features} of the random forest and \texttt{gamma} of the support vector machine depend on the number of features in the dataset. The default values of all other hyperparameters in scikit-learn that are involved in this study are not meta-feature dependent. As there is no indication of what functions might be reasonable for these hyperparameters, we will only apply a meta-feature dependent approach for \texttt{max\_features} and \texttt{gamma}. Let $p$ denote the number of features of a dataset. Then, the default parameters are $\sqrt{p}$ and $1/p$ for \texttt{max\_features} and \texttt{gamma} respectively. For these hyperparameters, we investigate an alternative default parameter estimation approach. Rather than choosing the value that occurs most frequently, we use non-linear least squares to fit several functions to the top 10 performance data, as implemented in the \texttt{curve\_fit} function of the python library \emph{scipy} \citep{scipy2001}.

\subsection{Tuning Risk of Hyperparameters}
In this section, we explain the experiment design and methods that are used to determine the importance of tuning a hyperparameter.

\subsubsection{Experiment Design}
We adhere to a repeated measures experiment design, where the unit of analysis is a dataset. The dependent variable in our study is the cross validated \textit{performance} of an instantiation of an algorithm. The two independent variables in our study are \textit{condition} and \textit{random search seed}. The \textit{condition} variable indicates whether a hyperparameter is \textit{fixed} or \textit{non-fixed}. That is, for some algorithm $A$, hyperparameter $i$, and dataset $j$, we have:
\begin{itemize}
    \item \textit{Fixed:} fix hyperparameter $i$ to default setting $\theta_i^{j}$ and tune all other $H - 1$ hyperparameters;
    \item \textit{Non-fixed:} tune hyperparameter $i$ as well as all other $H - 1$ hyperparameters.
\end{itemize}
The hyperparameters will be tuned using a random search strategy \citep{Bergstra2012} within a nested cross validation procedure. In the outer cross validation loop, we split the data into training and test sets. In the inner cross validation loop, we split each training set further into training and validation sets. In each iteration of the random search, the average validation set performance is determined to assess the performance of a hyperparameter setting. The best hyperparameter setting is used to determine the test set performance. The final performance of the algorithm is the average test set performance over all outer folds.

Although the random seeds are the same for fixed and non-fixed conditions, they are not shared among datasets. For example, the first random seed of task 1 is not necessarily identical to the first random seed of task 2.

\subsubsection{Approach}
We evaluate the results of the experiment through both an absolute measure (tuning risk) and a statistical measure (non-inferiority). \\

\noindent \textbf{Tuning Risk -} To understand to what extent the \textit{condition} affects the performance of the algorithm, we compute the difference in risk between the fixed and non-fixed conditions. Analogous to the \textit{tunability} measure introduced by \citet{Probst2018}, we define \textit{tuning risk} of hyperparameter $i$ for dataset $j$ and seed $s$ as follows:

\begin{equation}
d_{i,j,s} = R(\theta_i^{j}) - R(\theta_i^{*j,s})
\end{equation}

\noindent The tuning risk quantifies the risk of not tuning a particular hyperparameter. Larger values of $d_{i,j,s}$ indicate that the risk of the fixed condition is higher than the risk of the non-fixed condition. To summarize the differences over all $M$ datasets and seeds, we need an aggregating function. An obvious function is to compute the simple average. Other examples would be to use the median or some weighting scheme, e.g. based on properties of the dataset. Note that using the median would give more weight to intermediate risk values. Because we do not have evidence for the justification of a particular weighting scheme, we compute the simple average tuning risk of hyperparameter $i$: 

\begin{equation}
d_i = \frac{1}{S \cdot M} \sum_{s = 1}^{S} \sum_{j = 1}^{M} d_{i,j,s}
\end{equation}

\noindent We also investigate the standard deviation of $d_{i,j,s}$, which is denoted by $s_i$.

An assumption of this definition of tuning risk is that every unit of risk is equally relevant. In practice, however, this is often not the case. For example, when considering the missclassification rate, an increase from 0.01 to 0.02, might be considered worse than an increase from 0.49 to 0.50, as in the former case the number of misclassified instances is doubled. Hence, it depends on the learning task whether the absolute tuning risk is a relevant improvement or not. To alleviate this issue, we define the \textit{relative tuning risk} as the relative difference in risk between the fixed and non-fixed conditions. The measure is formalized as follows:

\begin{equation}
d_{i,j,s}^R = \frac{R(\theta_i^{j}) - R(\theta_i^{*j,s})}{R(\theta_i^{*j,s})}
\end{equation}

\noindent Again, we aggregate the relative tuning risk by computing the average:

\begin{equation}
d_i^R = \frac{1}{S \cdot M} \sum_{s=1}^{S} \sum_{j = 1}^{M} d^R_{i,j,s}
\end{equation}
\noindent Additionally, we compute the standard deviation of $d_{i,j,s}^R$, which we will denote by $s^R_i$. \\

\noindent \textbf{Non-inferiority -} We use the statistical measure of non-inferiority to determine whether the performance of the fixed condition is comparable to the performance of the non-fixed condition. This measure differs from traditional hypothesis testing, where the goal is to show that one group is different from another group with respect to some variable. Instead, we wish to determine whether an effect observed in one group is non-inferior to the effect in another group. Two one-sided non-inferiority tests can be combined into a method referred to as equivalence testing. This method is often applied in clinical and psychological research \citep{Walker2010, Lakens2017}. For example, one might want to determine whether some new treatment has similar benefits compared to an existing treatment. 

An important consideration in non-inferiority studies is the non-inferiority margin: the maximum difference in effect size that is considered to be irrelevant in practice. The non-inferiority margin is also known as smallest effect size of interest. It is more meaningful to define relevant improvements in performance on a relative scale than on an absolute scale. Therefore, we will only investigate the relative tuning risk in a non-inferiority test. We are mostly interested in the case where the observed risk in the fixed condition is higher than the risk observed in the non-fixed condition. We formulate our research question as follows:

\begin{quote}
    \textit{Is the observed risk in the fixed condition non-inferior to the risk observed in the non-fixed condition?}
\end{quote}

\noindent To answer our research question, we perform a non-parametric one sided test of non-inferiority for paired samples. We use a one-sided version of the Two One-Sided Tests (TOST) procedure described in the work of \citet{Constance2012}. The authors show that the non-parametric TOST procedure is more powerful for non-normal distributions than the TOST procedure that uses a paired t-test, because the former is less sensitive to outliers. Since we have no reason to assume that our data follows a normal distribution, we use the non-parametric alternative. The procedure is very similar to the Wilcoxon signed ranks tests, the biggest difference being the inversion of the null and alternative hypotheses. In the remainder of this section, we describe the procedure in more detail. 

We define the relative risk of observations as $\frac{x_{f} - x_{nf}}{x_{nf}}$, where $x_f$ denotes an observation of the fixed condition and $x_{nf}$ an observation of the non-fixed condition. Let $\delta$ be the non-inferiority margin for the relative risk. Additionally, let $M_{rr}$ be the population median of the relative risk. Then the null and alternative hypotheses are as follows:

\begin{align}
    H_0 & : {M_{rr}} \geq \delta \\
    H_1 & : {M_{rr}} < \delta
\end{align}

\noindent We compute signed ranks for observations $\frac{x_{f} - x_{nf}}{x_{nf}} - \delta$. Let $s_{nr}$ denote the absolute value of the sum of the negative ranks and $N$ the number of observations. Then the test statistic $z$ is defined as follows:
\begin{equation}
    z = \frac{s_{nr} - \left( \frac{N(N+1)}{4}\right)}{\sqrt{\frac{N(N+1)(2N+1)}{24}}}
\end{equation}
For a single test, $H_0$ is rejected if $z \geq z_{1-\alpha}$, where $z_{1-\alpha}$ is the value of the standard normal distribution at significance level $\alpha$. The experiment will be performed multiple times in this study, each time for a different hyperparameter. We use the Holm-Bonferroni method \citep{holm1979} to control the family-wise error.

\section{Experiment details}
\label{sec:exp}
In this section we give an overview of the datasets, algorithms, performance measures, and experiment setups. All experiments are executed using dockerized Python applications\footnote{See \url{https://github.com/hildeweerts/hyperimp}.} on the Azure Cloud Computing platform.

\subsection{Datasets}
We use data from the open machine learning environment OpenML \citep{Vanschoren2014}. In particular, we use datasets from the OpenML-CC18, a curated machine learning benchmark suite of 73 classification datasets. A description of these datasets can be found in Appendix~\ref{app:cc18}. In this study, we used 59 of these 73 datasets. One dataset was not included in this study because of technical issues arising from the large number of nominal features in this dataset. Additionally, due to time constraints, we limited the training time of a single algorithm run on a dataset to 3 hours. Datasets for which more than 10\% of the default value experiment runs lasted more than 3 hours are left out of the analysis. The 14 datasets that are excluded from the study are listed in Appendix~\ref{app:cc18ni}. It is important to note that all excluded datasets had a relatively high number of features, instances, or both. This might bias our results towards smaller datasets when considering meta-feature dependent default values.

\subsection{Algorithms and Hyperparameters}
In this work we consider two algorithms as implemented in  scikit-learn. We consider a Support Vector Machine (SVM) with a Radial Basis Function (RBF) kernel and the Random Forest. The parameter ranges that are considered are taken from the automatic machine learning package {auto-sklearn} \citep{NIPS2015_5872}. The ranges for SVM and random forest can be found in Table~\ref{tab:svm-ranges} and Table~\ref{tab:rf-ranges} respectively.

\begin{table}[ht]
\centering
\caption{Hyperparameter ranges for \texttt{SVC}.}
\label{tab:svm-ranges}
\begin{tabular}{lll}
\midrule
\textbf{Hyperparameter} & \textbf{Type} & \textbf{Range}  \\ \midrule
gamma                   & continuous    & $[2^{-15}, 2^3]$ (log-scale)     \\ 
C                       & continuous    & $[2^-5, 2^15]$ (log-scale)     \\ 
tol                     & continuous    & $10^{-5}, 10^{-1}]$ (log-scale)  \\ 
shrinking               & boolean       & \{True, False\} \\ 
kernel                 & discrete      & rbf \\ \bottomrule

\end{tabular}
\end{table}

\begin{table}[ht]
\centering
\caption{ Hyperparameter ranges for \texttt{RandomForestClassifier}. Parameters annotated with * are fixed to a value different than the default of scikit-learn.}
\label{tab:rf-ranges}
\begin{tabular}{lll}
\toprule
\textbf{Hyperparameter} & \textbf{Type} & \textbf{Range}  \\ \midrule
bootstrap               & boolean       & \{True, False\} \\ 
criterion               & nominal       & \{gini, entropy\} \\ 
max\_features           & continuous    &    [0,1]    \\ 
min\_samples\_leaf      & integer       & {[}1, 20{]}     \\ 
min\_samples\_split     & integer       & {[}2, 20{]}     \\ 
n\_estimators*          & integer       & 500             \\ \bottomrule
\end{tabular}
\end{table}

The complete pipeline consists of simple pre-processing steps and the algorithm. The pre-processing pipeline includes missing value imputation by the mean (continuous features) or mode (categorical features), one hot encoding for categorical features, and removal of features with zero variance. Because SVM's are known to be sensitive to different feature scales, a scaling step is added to the SVM pre-processing pipeline. 

\subsection{Performance Measures}
\label{sec:perfmeasures}
We compare two performance measures, as implemented in OpenML, at the default parameter determination stage: accuracy and macro-averaged Area under the ROC Curve (AUC). In macro-averaging, the measure is computed locally over each category first, then the average over all categories is taken, weighted by the number of instances of that class. The macro-averaged AUC is computed using the approach of \citet{Provost00:}. In this approach, the ROC curve of each class is constructed by comparing the class of interest to the union of all other classes. Note that these curves can be sensitive to changes in prevalence of the classes within the union. At this point, multi-class AUC is not yet implemented in scikit-learn. Therefore, we will use accuracy as a performance measure in the random search procedure of the hyperparameter importance experiment.

\subsection{Experiment 1: Default Values}
In order to determine good default values, we need empirical risk measurements $\langle R(\bm{\theta}_k) \rangle_{k=1}^{K}$ for each of the $M$ datasets. In this experiment $M = 59$ and we choose $K = 1000$. In other words, for each dataset, we evaluate 1,000 random configurations $\bm{\theta}_i$ of each algorithm, i.e. 59,000 evaluations per algorithm. Each hyperparameter will be picked uniformly at random from the corresponding hyperparameter range and scale. The performance of a configuration is calculated using 10-fold cross validation. Recall that we use the best $n$ settings for each task to determine the default parameters. In accordance to the work of \citet{Vanrijn2018}, we pick $n = 10$.

\subsection{Experiment 2: Importance of Tuning}
We use nested cross validation to determine the performance of both the fixed and non-fixed conditions. In the outer loop, we use 10-fold cross validation to split the data in training and test sets. In the inner loop, we use 5-fold cross validation to split the training data further into training and validation sets. For each fold in the inner loop, a random search strategy with 100 iterations is applied to tune hyperparameters. Note that the hyperparameter settings that are used to determine test set performance can be different for each of the 10 folds in the outer cross validation. We repeat the experiment $S = 10$ times, using a different seed for the random search each time. 

\section{Results}
\label{sec:results}
In this section we discuss the results of our experiments. First, we discuss the results of the default value experiment. Second, we discuss the results from the hyperparameter tuning importance experiments. All performance data that was used in this study is publicly available on OpenML\footnote{See \url{https://www.openml.org/s/98/}.}.

\subsection{Experiment 1: Default Values}
In this section, we first discuss the collected performance data. Subsequently, we discuss the computed default values. In particular, we compare our computations with the default values computed by \citet{Probst2018} and the default values used in scikit-learn. Finally, we discuss the meta-feature dependent default parameters.

\subsubsection{Performance data}
\begin{figure}[ht]
\captionsetup[subfigure]{justification=centering}
\centering
\begin{subfigure}[t]{.75\linewidth}
    \centering
    \includegraphics[width=\textwidth]{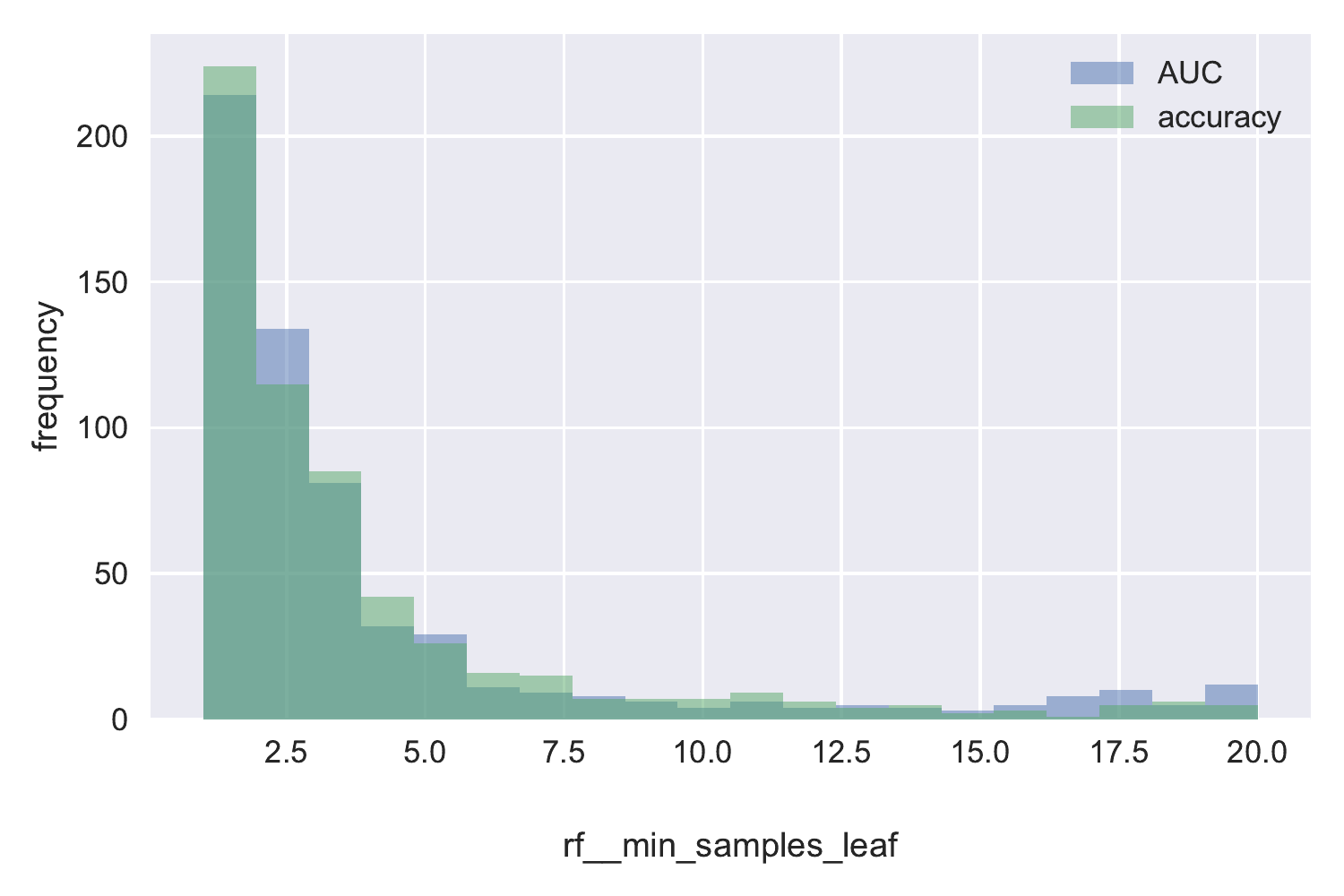}
    \caption{Random Forest - \texttt{min\_samples\_leaf}}\label{fig:histminsamples}
\end{subfigure}

    \bigskip
    
\begin{subfigure}[t]{.75\linewidth}
    \centering
    \includegraphics[width=\textwidth]{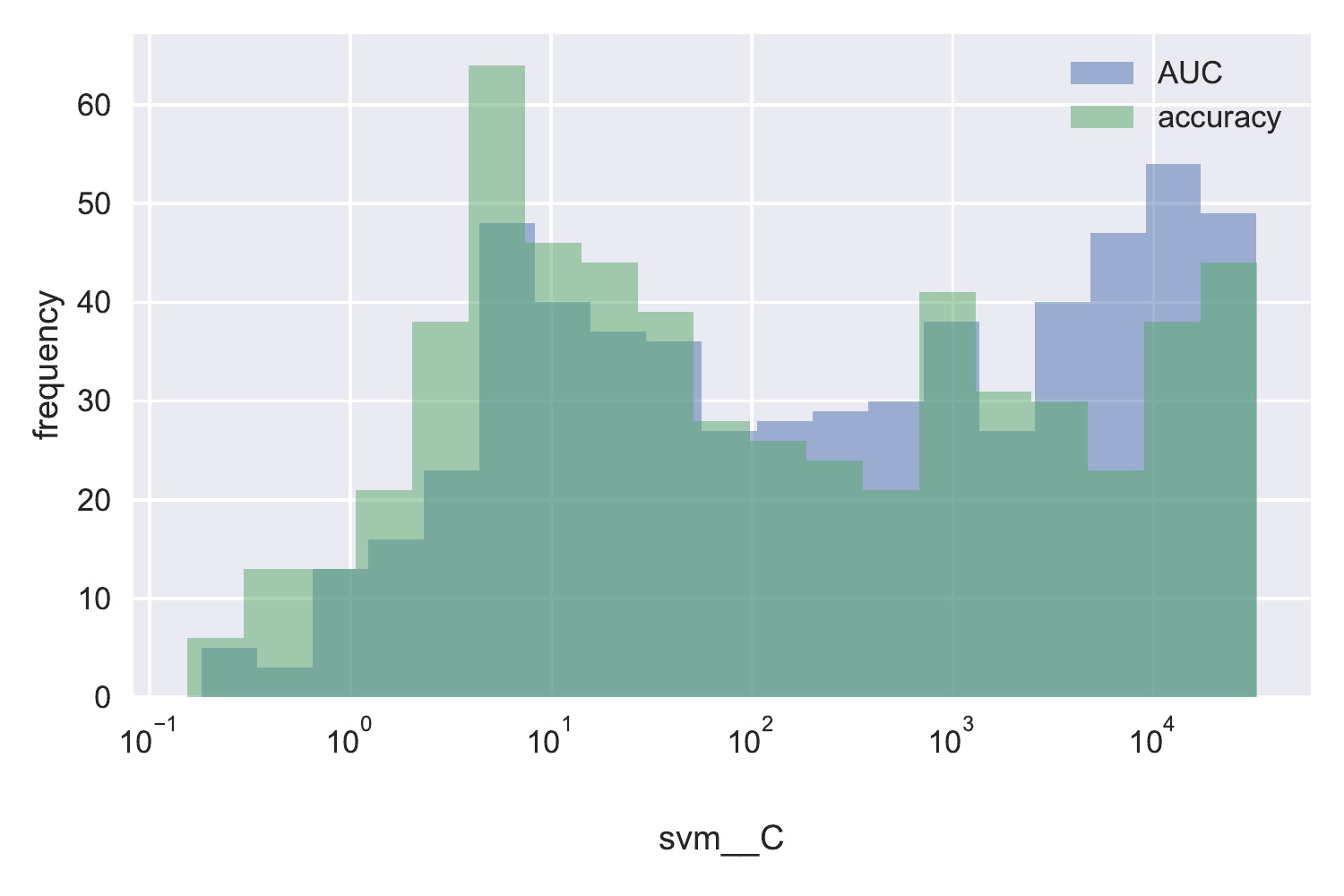}
    \caption{SVM - \texttt{C}}\label{fig:C}
    \end{subfigure}
\caption{ The distribution of \texttt{min\_samples\_leaf} and \texttt{C} within the top 10 highest performing hyperparameter settings per dataset, derived from either accuracy or AUC-based performance data.}
\label{fig:top10example}
\end{figure}

After selecting the top 10 performance data, we visualized the distributions of the hyperparameter settings in a histogram, as illustrated in Figure~\ref{fig:top10example}. From the distribution depicted in Figure~\ref{fig:histminsamples}, we can conclude that fixing \texttt{min\_samples\_leaf} to 1 resulted in good performance most often. For most hyperparameters, the distribution of accuracy- and AUC-based data is very similar. For others, one may arrive at a different default parameter depending on the performance measure that was used. For instance, consider the distribution of \texttt{C} shown in Figure~\ref{fig:C}. An accuracy-based default value for \texttt{C} would be in the order of magnitude 1, whereas an AUC-based default value would be in the order of magnitude 4. The histograms of all other hyperparameters can be found in Appendix~\ref{app:top10hists}.

\subsubsection{Default values}
Recall that the default values are computed for each task separately, using a leave-one-out approach. The average and standard deviation of the computed default values across all datasets are displayed in Table~\ref{tab:defaults}.

For most hyperparameters, the standard deviation of the computed default values across datasets is zero or relatively small, considering the scaling. For example, the default values of \texttt{C} using an accuracy-based approach range between 3 and 11, with a standard deviation of 3.224. This is negligible on a logarithmic scale. The distribution of default values with non-zero variance are displayed in Appendix~\ref{app:defval}.

\begin{table*}[ht]
\centering
\caption{ Average/Mode (numerical/categorical) and standard deviation of our calculated defaults over all datasets, defaults of scikit-learn, and defaults as calculated by \citet{Probst2018}. Hyperparameters that are not computed in the work of \citet{Probst2018} (e.g. because they do not exist in the R implementation) are indicated with N/A. The standard deviation that is shown for \texttt{max\_features}, annotated with *, is the standard deviation of the exponent. The standard deviation that is shown for boolean variables, annotated with **, is the standard deviation when considering \texttt{True} as 1 and \texttt{False} as 0. The standard deviation shown for \texttt{criterion}, annotated with ***, is the standard deviation when considerint \texttt{entropy} as 1 and \texttt{gini} as 0.}
\label{tab:defaults}
\begin{tabular}{lrrrrrr}
\toprule
\multicolumn{1}{l}{\textbf{Hyperparameter}}      & \multicolumn{1}{c}{\textbf{\begin{tabular}[c]{@{}c@{}}Average/Mode\\ (accuracy)\end{tabular}}} & \multicolumn{1}{c}{\textbf{\begin{tabular}[c]{@{}c@{}}Standard \\ deviation \\ (accuracy)\end{tabular}}} & \multicolumn{1}{c}{\textbf{\begin{tabular}[c]{@{}c@{}}Average/Mode\\ (AUC)\end{tabular}}} & \multicolumn{1}{c}{\textbf{\begin{tabular}[c]{@{}c@{}}Standard\\ deviation\\ (AUC)\end{tabular}}} & \multicolumn{1}{c}{\textbf{scikit-learn}} & \multicolumn{1}{c}{\textbf{\citet{Probst2018}}} \\ \midrule

\textbf{Random Forest}                             &                                                                                                 &                                                                                                          &                                                                                            &                                                                                                    &                                       &                                                                   \\ \midrule
\multicolumn{1}{l}{\textbf{bootstrap}}           & \multicolumn{1}{r}{False}                                                                      & \multicolumn{1}{r}{**0}                                                                                   & \multicolumn{1}{r}{False}                                                                 & \multicolumn{1}{r}{**0}                                                                             & \multicolumn{1}{r}{True}             & False                                                             \\ 
\multicolumn{1}{l}{\textbf{criterion}}           & \multicolumn{1}{r}{entropy}                                                                    & \multicolumn{1}{r}{***0}                                                                                   & \multicolumn{1}{r}{entropy}                                                               & \multicolumn{1}{r}{***0}                                                                             & \multicolumn{1}{r}{gini}             & N/A                                                               \\ 
\multicolumn{1}{l}{\textbf{max\_features}}       & \multicolumn{1}{r}{n\textasciicircum{}0.737}                                                   & \multicolumn{1}{r}{*0.009}                                                                              & \multicolumn{1}{r}{n\textasciicircum{}0.714}                                              & \multicolumn{1}{r}{*0.006}                                                                        & \multicolumn{1}{r}{sqrt(n)}          & 0.284                                                             \\ 
\multicolumn{1}{l}{\textbf{min\_samples\_leaf}}  & \multicolumn{1}{r}{1}                                                                          & \multicolumn{1}{r}{0}                                                                                   & \multicolumn{1}{r}{1}                                                                     & \multicolumn{1}{r}{0}                                                                             & \multicolumn{1}{r}{1}                & 0                                                                \\ 
\multicolumn{1}{l}{\textbf{min\_samples\_split}} & \multicolumn{1}{r}{5}                                                                          & \multicolumn{1}{r}{0}                                                                                   & \multicolumn{1}{r}{2.3}                                                                   & \multicolumn{1}{r}{0.8}                                                                           & \multicolumn{1}{r}{2}                & N/A                                                               \\ \midrule

\textbf{SVM}                                       &                                                                                                 &                                                                                                          &                                                                                            &                                                                                                    &                                       &                                                                   \\ \midrule
\multicolumn{1}{l}{\textbf{C}}                   & \multicolumn{1}{r}{8.888}                                                                      & \multicolumn{1}{r}{3.224}                                                                               & \multicolumn{1}{r}{16646}                                                                 & \multicolumn{1}{r}{63}                                                                            & \multicolumn{1}{r}{1.000}            & 475.504                                                           \\ 
\multicolumn{1}{l}{\textbf{gamma}}               & \multicolumn{1}{r}{1.2 e-02}                                                                   & \multicolumn{1}{r}{1.5 e-04}                                                                            & \multicolumn{1}{r}{8.1 e-03}                                                              & \multicolumn{1}{r}{1.5 e-03}                                                                      & \multicolumn{1}{r}{1/n}              & 5 e-03                                                             \\ 
\multicolumn{1}{l}{\textbf{shrinking}}           & \multicolumn{1}{r}{True}                                                                       & \multicolumn{1}{r}{**0}                                                                                   & \multicolumn{1}{r}{True}                                                                  & \multicolumn{1}{r}{**0.36}                                                                        & \multicolumn{1}{r}{True}             & N/A                                                               \\ 
\multicolumn{1}{l}{\textbf{tol}}           & \multicolumn{1}{r}{4.6 e-05}                                                                   & \multicolumn{1}{r}{8.8 e-08}                                                                            & \multicolumn{1}{r}{5.4 e-02}                                                              & \multicolumn{1}{r}{1.3 e-02}                                                                      & \multicolumn{1}{r}{1.0 e-03}         & N/A                                                               \\ \bottomrule
\end{tabular}
\end{table*}

The differences between defaults resulting from the accuracy and AUC approach are in line with the histograms shown in the previous section. For \texttt{bootstrap}, \texttt{criterion}, \texttt{min\_samples\_split}, \texttt{C}, \texttt{tol} we find different default parameters than the ones currently used in scikit-learn, although it should be noted that the default parameter of \texttt{min\_samples\_split} in scikit-learn was returned often by the AUC-based approach. For \texttt{bootstrap} and \texttt{gamma}, we find similar values as \citet{Probst2018}. On the other hand, the default values for \texttt{C} differ a lot between the accuracy-based defaults, AUC-based defaults, the default used in scikit-learn, and the defaults computed by \citet{Probst2018}.

\subsubsection{Meta-feature dependent hyperparameters}
\label{sec:metadep}
\begin{figure}[ht]
\captionsetup[subfigure]{justification=centering}
\centering
\begin{subfigure}[t]{.75\linewidth}
    \centering
    \includegraphics[width=\textwidth]{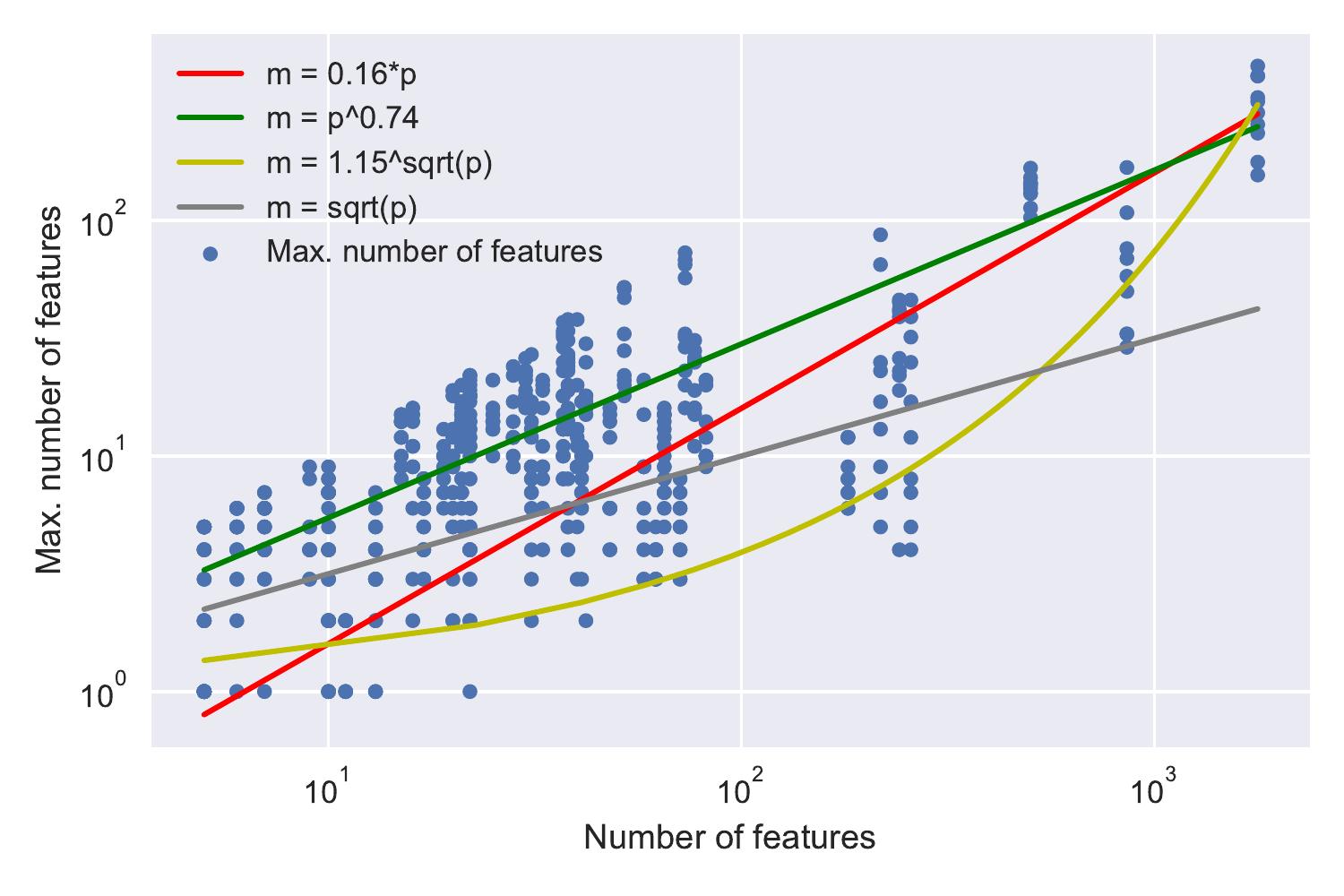}
    \caption{Random Forest -\texttt{ max\_features}}\label{fig:mdmaxfeatures}
\end{subfigure}

    \bigskip
    
\begin{subfigure}[t]{.75\linewidth}
    \centering
    \includegraphics[width=\textwidth]{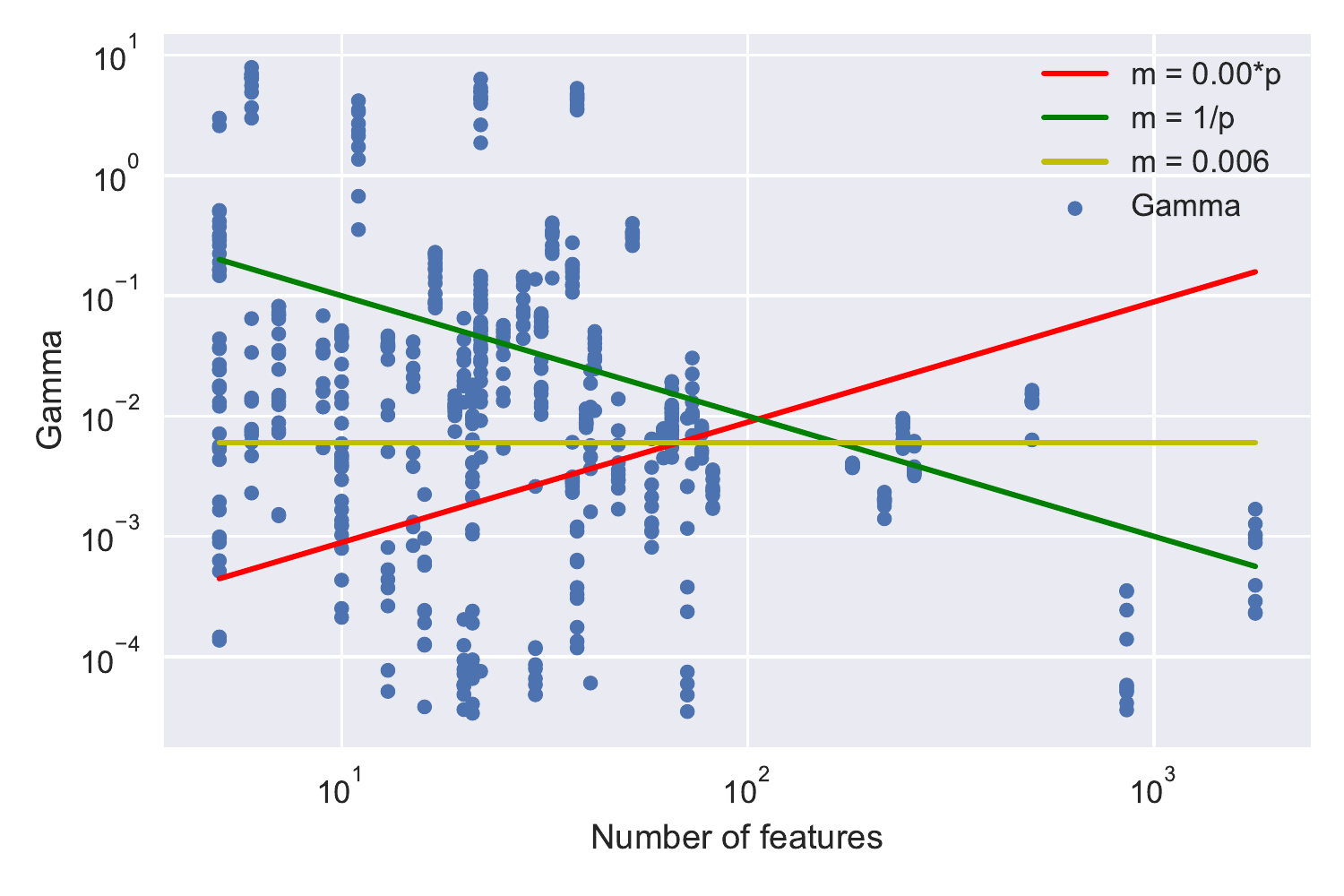}
    \caption{SVM - \texttt{gamma}}\label{fig:mdgamma}
    \end{subfigure}
\caption{ Meta-feature dependent parameter values of the top 10 performance data (accuracy-based only) against the number of features ($p$) of the corresponding dataset.}
\label{fig:metadependent}
\end{figure}

As stated in Section~\ref{sec:methods}, we will determine the hyperparameters for \texttt{max\_features} and \texttt{gamma} using an approach that considers the number of features of a dataset. Figure~\ref{fig:metadependent} shows a scatter plot of the hyperparameter value versus the number of features, including the functions that that were fitted. The Root Mean Squared Error (RMSE), R$^2$, Root Mean Squared Logarithmic Error (RMSLE), and logarithmic R$^2$ (LR$^2$) for each of the functions can be found in TableZ\ref{tab:meta}.

\begin{table}[ht]
\centering
\caption{ RMSE, R$^2$, RMSLE, and LR$^2$ of several functions that depend on the number of features of the dataset, fitted on the accuracy-based top 10 performance data for \texttt{max\_features} and \texttt{gamma}.}
\label{tab:meta}
\begin{tabular}{lrrrr}
\toprule
\multicolumn{1}{l}{\textbf{function}}                 & \multicolumn{1}{c}{\textbf{RMSE}}           & \multicolumn{1}{c}{\textbf{R$^2$}} & \multicolumn{1}{c}{\textbf{RSMLE}}          & \multicolumn{1}{c}{\textbf{LR$^2$}}             \\ \midrule

{\textbf{max\_features}}                         & \multicolumn{1}{l}{} & \multicolumn{1}{l}{}       & \multicolumn{1}{l}{} & \multicolumn{1}{l}{} \\ \midrule
\multicolumn{1}{l}{m = 0.16*p}                        & \multicolumn{1}{r}{21.64}                   & \multicolumn{1}{r}{0.77}                          & \multicolumn{1}{r}{0.90}                    & 0.13                                          \\ 
\multicolumn{1}{l}{m = p \textasciicircum 0.74}       & \multicolumn{1}{r}{23.13}                   & \multicolumn{1}{r}{0.73}                          & \multicolumn{1}{r}{0.73}                    & 0.42                                          \\ 
\multicolumn{1}{l}{m = 1.15\textasciicircum{}sqrt(p)} & \multicolumn{1}{r}{24.33}                   & \multicolumn{1}{r}{0.70}                          & \multicolumn{1}{r}{1.24}                    & -0.66                                         \\ 
\multicolumn{1}{l}{m = sqrt(p)}       & \multicolumn{1}{r}{41.38}                   & \multicolumn{1}{r}{0.14}                         & \multicolumn{1}{r}{0.81}                    & 0.28                                         \\ \midrule

{\textbf{gamma}}                                 & \multicolumn{1}{l}{} & \multicolumn{1}{l}{}       & \multicolumn{1}{l}{} & \multicolumn{1}{l}{} \\ \midrule
\multicolumn{1}{l}{m = 0.00574*p}                     & \multicolumn{1}{r}{1.25}                    & \multicolumn{1}{r}{-0.07}                         & \multicolumn{1}{r}{0.46}                    & -0.12                                         \\ 
\multicolumn{1}{l}{m = 1/p}                           & \multicolumn{1}{r}{1.24}                    & \multicolumn{1}{r}{-0.05}                         & \multicolumn{1}{r}{0.44}                    & -0.04                                         \\ 
\multicolumn{1}{l}{m = 0.006}                         & \multicolumn{1}{r}{1.27}                    & \multicolumn{1}{r}{-0.09}                         & \multicolumn{1}{r}{0.46}                    & -0.14                                         \\ \bottomrule
\end{tabular}
\end{table}
 From Figure~\ref{fig:mdmaxfeatures}, it becomes clear that for our datasets, the default value in scikit-learn, $\sqrt{p}$, seems to underestimate \texttt{max\_features}. In fact, on a linear scale, it performs worse than all other functions that we considered, whereas $m = 0.16^p$ performs best. However, RMSE and R$^2$ are slighty biased, because there are relatively few large datasets. When we consider the RSMLE and LR$^2$ instead, $m = p^{0.74}$ performs best, followed by $m = \sqrt{p}$. In Figure~\ref{fig:mdgamma}, we do not observe a clear pattern for \texttt{gamma}. Additionally, we observe negative values for both R$^2$ and LR$^2$ for all functions in Table~\ref{tab:meta}. The default value in scikit-learn, $1/p$, performs slightly better than the other two functions.

In experiment 2, where we investigate the importance of tuning, we will consider $\sqrt{p}$ and $p^{0.74}$ for \texttt{max\_features}, and $1/p$ and 0.006 for \texttt{gamma}.

\subsection{Experiment 2: Importance of Tuning}
\label{sec:imptun}
\begin{figure}[ht]
    \centering
    \includegraphics[width=0.75\linewidth]{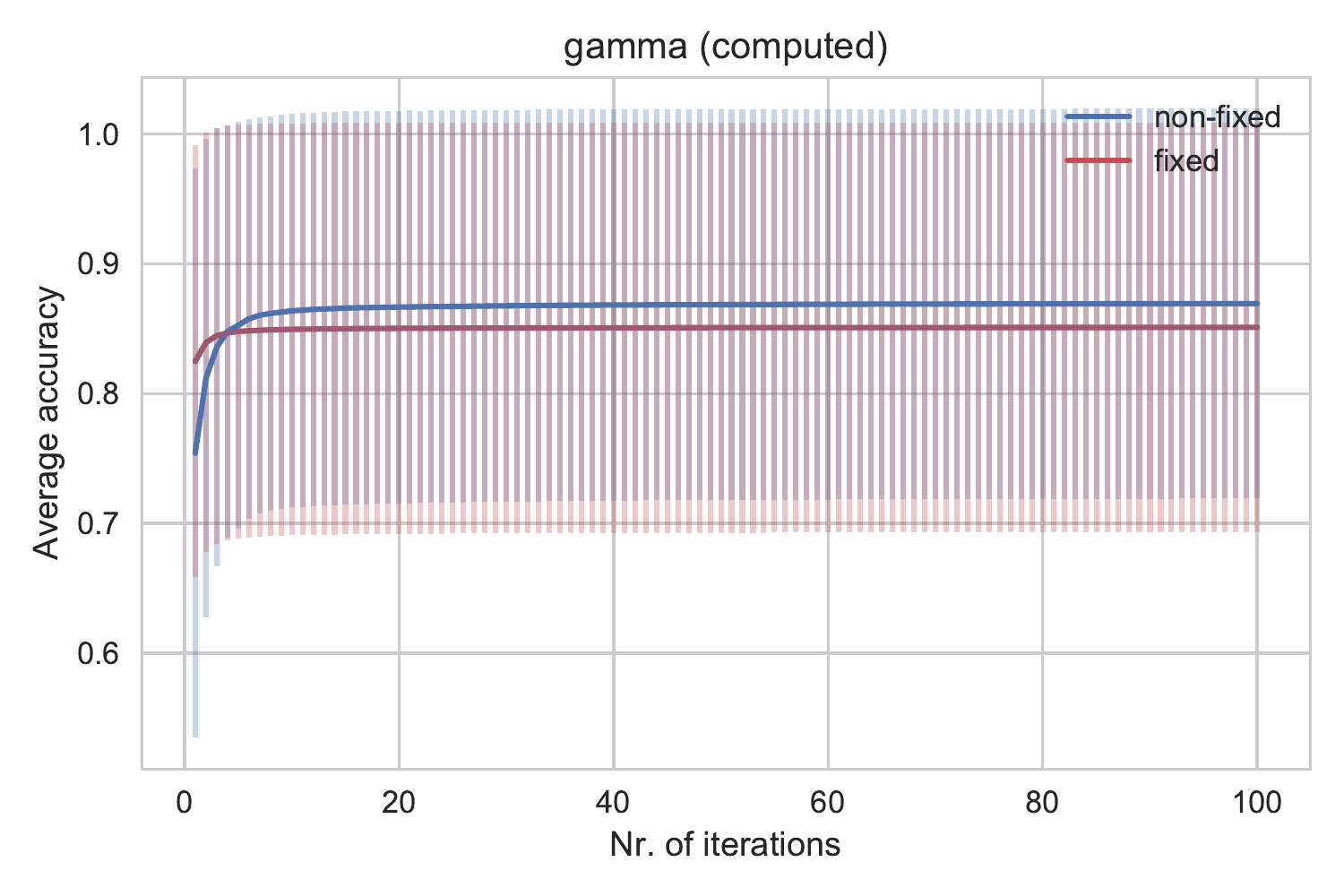}
    \caption{Average accuracy (+/- standard deviation) of the the fixed and non-fixed conditions over number of iterations for hyperparameter \texttt{gamma} (computed default setting).}
    \label{fig:accuracyexamples}
\end{figure}

\begin{figure}[ht]
    \centering
    \includegraphics[width=0.9\linewidth]{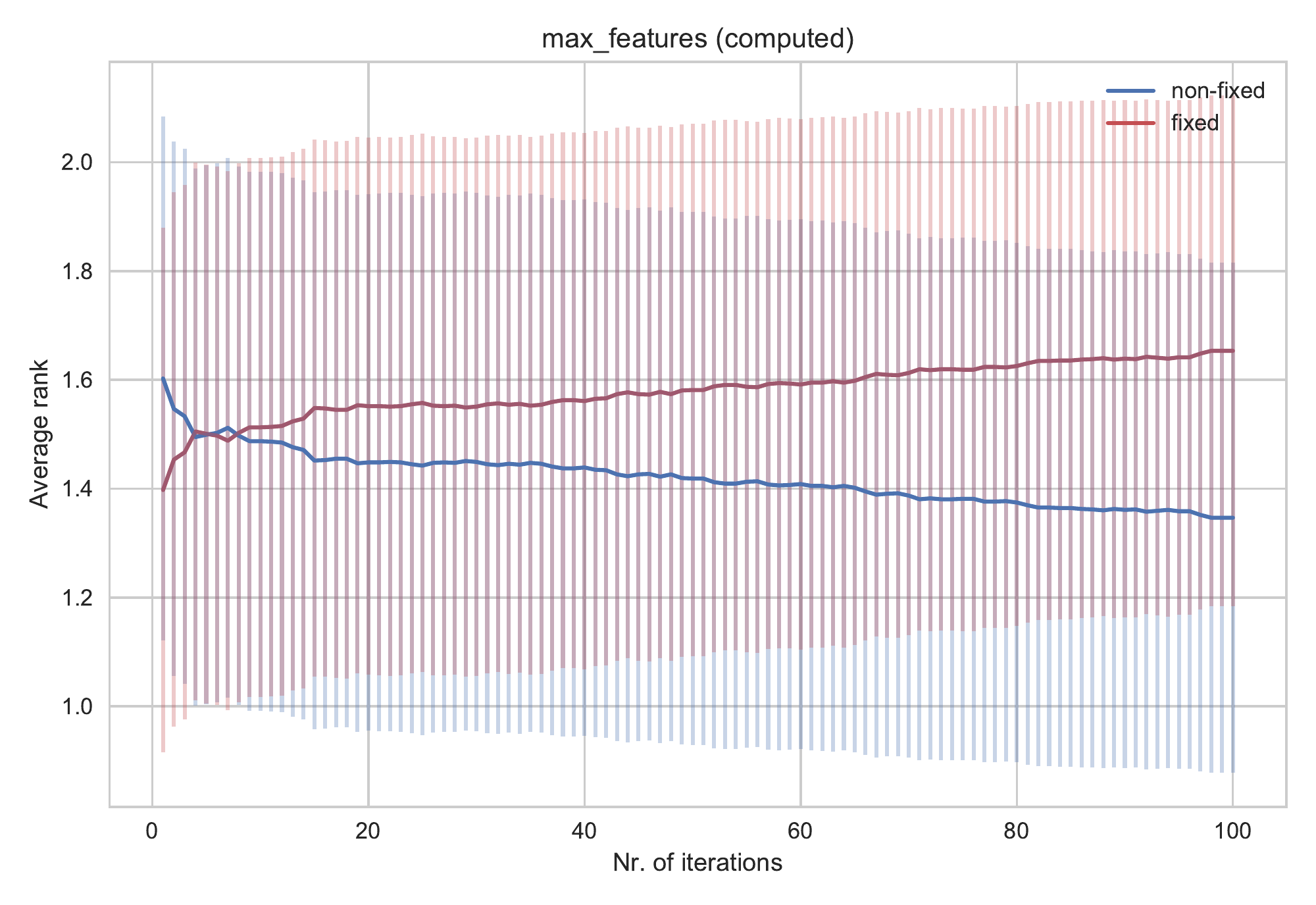}
    \caption{Average rank (+/- standard deviation) of the the fixed and non-fixed conditions over number of iterations for hyperparameter \texttt{max\_features} (computed default setting).}
    \label{fig:ranksexamples}
\end{figure}

In order to investigate the tuning process, we visualize the average accuracy and rank of the fixed and non-fixed conditions at each iteration. As an example, Figure~\ref{fig:accuracyexamples} depicts the maximum average validation set accuracy observed up until a certain iteration of the random search for \texttt{gamma}, using our computed default value. Figure~\ref{fig:ranksexamples} depicts the average rank over the number of iterations for \texttt{max\_features}, using our computed default value. The ranks are computed as follows. For each measurement, for each iteration, the condition with the highest validation accuracy receives rank 1 and the other rank 2. The visualizations of the other hyperparameter experiments can be found in Appendix~\ref{app:accuracy} and Appendix~\ref{app:ranks}.

For all hyperparameters except SVM's \texttt{shrinking}, the average rank of the fixed condition is lower than the average rank of the non-fixed condition in the first iteration. For many hyperparameters, the fixed condition, on average, outperforms the non-fixed condition for all 100 iterations. This indicates that 100 iterations was not enough to find the best possible hyperparameter settings of the algorithm. For SVM's \texttt{shrinking} and \texttt{tol}, we observe that the average rank is consistently close to 1.5, which indicates that neither the fixed nor the non-fixed condition was superior. For random forest's \texttt{max\_features} (using the computed default value), SVM's \texttt{C}, SVM's \texttt{gamma} (both computed default value and scikit-learn's default value), the non-fixed condition, on average, outperforms the fixed condition after 15 to 30 iterations.

\subsubsection{Tuning Risk}
\begin{figure*}[ht]
\captionsetup[subfigure]{justification=centering}
\centering
\begin{subfigure}[t]{\linewidth}
    \centering
    \includegraphics[width=\textwidth]{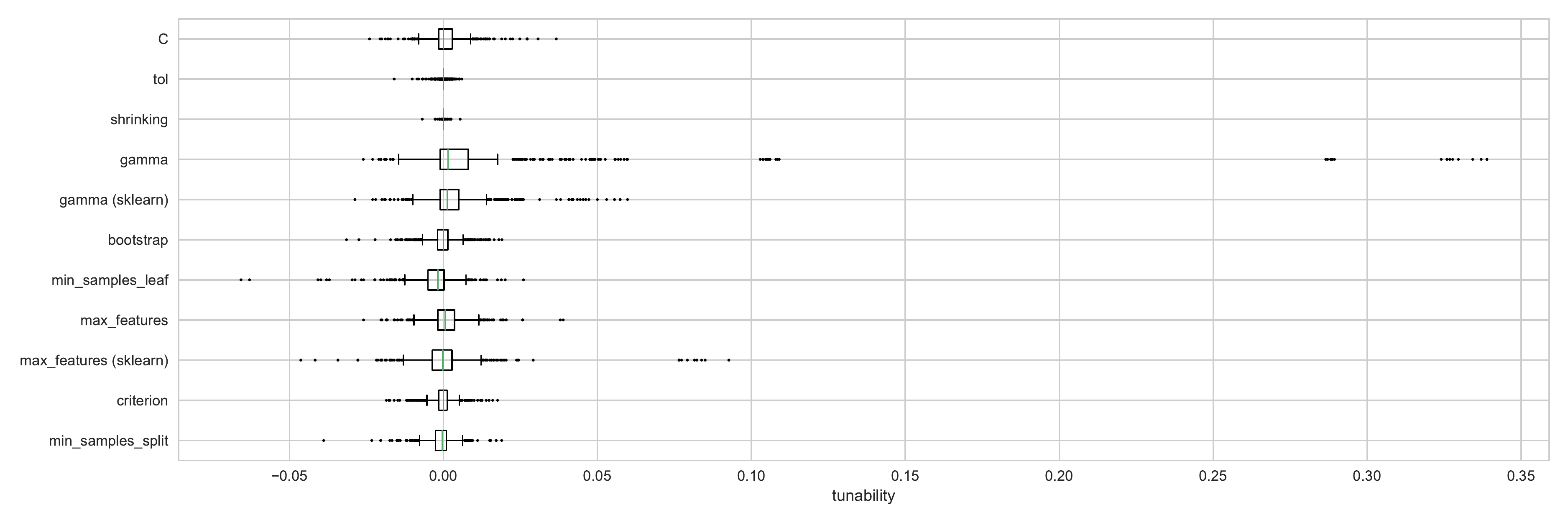}
    \caption{Tuning risk}\label{fig:tunability}
\end{subfigure}

    \bigskip
    
\begin{subfigure}[t]{\linewidth}
    \centering
    \includegraphics[width=\textwidth]{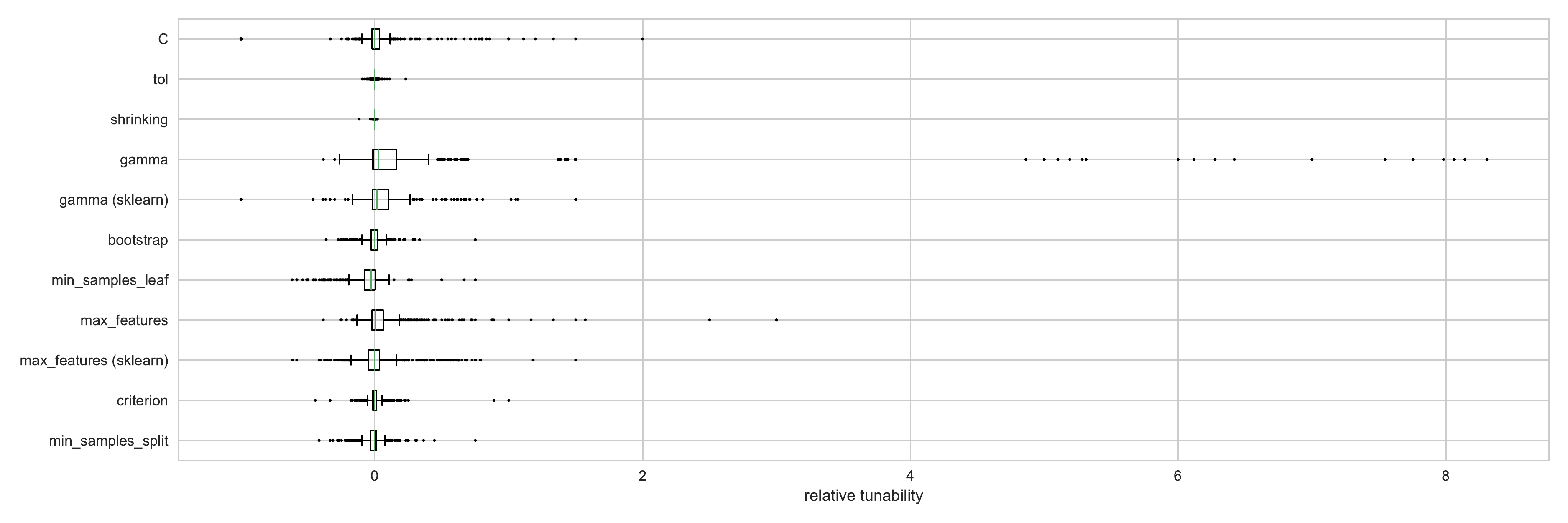}
    \caption{Relative tuning risk}\label{fig:reltunability}
    \end{subfigure}
\caption{ The distribution of tuning risk and relative tuning risk measurements ($d_{i,j,r}$) per hyperparameter.}
\label{fig:tunabilityboxplots}
\end{figure*}

The average and standard deviation of both tuning risk and relative tuning risk are shown in Table~\ref{tab:tunability}. Additionally, Figure~\ref{fig:tunabilityboxplots} shows the distribution of tuning risk for each of the hyperparameters. There exist three datasets for which the non-fixed SVM almost always resulted in perfect accuracy: OpenML task 11, 49, and 10093. Because relative tuning risk is undefined when the non-fixed risk is equal to 0, these tasks are left out of the tuning risk analysis. Note that the tuning risk is often negative. A negative tuning risk indicates that the hyperparameters in the non-fixed condition have not been tuned to their optimal value. If they were, the performance of the non-fixed condition should always be at least as good as the performance of the fixed condition, as an exhaustive search would include the default value of the fixed condition. This is related to the length of our random search procedure, which we limited to 100 iterations. In future research, this issue could be alleviated by increasing the number of iterations or using a smarter search procedure such as the prior based approach suggested by \cite{Vanrijn2018}.

\begin{table}[ht]
\centering
\caption{ Average and standard deviation of tuning risk ($d_i$, $s_i$) and relative tuning risk ($d_i^R$, $s_i^R$).}
\label{tab:tunability}
\begin{tabular}{lrrrr}
\toprule
\multicolumn{1}{l}{\textbf{Hyperparameter}}                                            & \multicolumn{1}{c}{\bm{$d_i$}}          & \multicolumn{1}{c}{\bm{$s_i$}}          & \multicolumn{1}{c}{\bm{$d_i^R$}}        & \multicolumn{1}{c}{\bm{$s_i^R$}}         \\ \midrule

\textbf{Random Forest}                                                                   & \multicolumn{1}{l}{} & \multicolumn{1}{l}{} & \multicolumn{1}{l}{} & \multicolumn{1}{l}{} \\ \midrule

\multicolumn{1}{l}{\textbf{bootstrap}}                                                          & \multicolumn{1}{r}{-0.0002}                 & \multicolumn{1}{r}{0.0053}                  & \multicolumn{1}{r}{-0.0053}                 & 0.0820                                        \\ 
\multicolumn{1}{l}{\textbf{criterion}}                                                          & \multicolumn{1}{r}{-0.0002}                 & \multicolumn{1}{r}{0.0041}                  & \multicolumn{1}{r}{0.0035}                  & 0.0793                                        \\ 
\multicolumn{1}{l}{\textbf{max\_features}}                                                      & \multicolumn{1}{r}{0.0011}                  & \multicolumn{1}{r}{0.0068}                  & \multicolumn{1}{r}{0.0742}                  & 0.2822                                        \\ 
\multicolumn{1}{l}{\begin{tabular}[c]{@{}l@{}}\textbf{max\_features} \\ \textbf{(scikit-learn)}\end{tabular}} & \multicolumn{1}{r}{0.0010}                  & \multicolumn{1}{r}{0.0134}                  & \multicolumn{1}{r}{0.0309}                  & 0.2037                                        \\ 
\multicolumn{1}{l}{\textbf{min\_samples\_leaf}}                                                 & \multicolumn{1}{r}{-0.0030}                 & \multicolumn{1}{r}{0.0077}                  & \multicolumn{1}{r}{-0.0505}                 & 0.1280                                        \\ 
\multicolumn{1}{l}{\textbf{min\_samples\_split}}                                                & \multicolumn{1}{r}{-0.0009}                 & \multicolumn{1}{r}{0.0048}                  & \multicolumn{1}{r}{-0.0110}                 & 0.0829                                        \\ \midrule

\textbf{SVM}                                                                             & \multicolumn{1}{l}{} & 
\multicolumn{1}{l}{} & \multicolumn{1}{l}{} & \multicolumn{1}{l}{} \\ \midrule

\multicolumn{1}{l}{\textbf{C}}                                                                  & \multicolumn{1}{r}{0.0008}                  & \multicolumn{1}{r}{0.0062}                  & \multicolumn{1}{r}{0.0252}                  & 0.2415                                        \\ \midrule
\multicolumn{1}{l}{\textbf{gamma}}                                                              & \multicolumn{1}{r}{0.0180}                  & \multicolumn{1}{r}{0.0599}                  & \multicolumn{1}{r}{0.4595}                  & 1.4687                                        \\ 
\multicolumn{1}{l}{\textbf{gamma (scikit-learn)}}                                                    & \multicolumn{1}{r}{0.0039}                  & \multicolumn{1}{r}{0.0115}                  & \multicolumn{1}{r}{0.0503}                  & 0.2498                                        \\ 
\multicolumn{1}{l}{\textbf{shrinking}}                                                          & \multicolumn{1}{r}{0.0000}                  & \multicolumn{1}{r}{0.0005}                  & \multicolumn{1}{r}{-0.0003}                 & 0.0057                                        \\ 
\multicolumn{1}{l}{\textbf{tol}}                                                                & \multicolumn{1}{r}{-0.0001}                 & \multicolumn{1}{r}{0.0016}                  & \multicolumn{1}{r}{0.0006}                  & 0.0179                                        \\ \bottomrule
\end{tabular}
\end{table}

From Table~\ref{tab:tunability}, we observe that for \texttt{bootstrap}, \texttt{criterion}, \texttt{min\_samples\_leaf}, \texttt{min\_samples\_split}, \texttt{shrinking}, and \texttt{tol}, both the tuning risk and relative tuning risk are close to or lower than zero. This is in line with the average ranks shown in the previous section.

In Figure~\ref{fig:tunability}, we observe a few high tuning risk scores for \texttt{gamma} (computed default) and \texttt{max\_features} (scikit-learn default). This indicates that there exist a few datasets for which the fixed condition performed much worse than the non-fixed condition. This effect is even larger for \texttt{gamma} (computed default) when considering relative tuning risk. The tuning risk of \texttt{gamma} when using scikit-learn's default parameter is much better than the tuning risk of \texttt{gamma} when using our non meta-feature dependent default setting. This indicates that it is important to take into account the number of features when optimizing \texttt{gamma}. 

For \texttt{max\_features}, our computed default value resulted in slightly better tuning risk but worse relative tuning risk compared to using scikit-learn's default value. In particular, there are several outliers with a relative tuning risk higher than 2. This is not as expected, given the analysis on meta-feature dependent parameter values in Section~\ref{sec:metadep}. It indicates that the top 10 performance data gives a decent representation of absolute performance, but not necessarily of relative performance. This might be caused by the fact that all data points are weighted equally when determining the default value, disregarding the fact that some learning tasks are much easier than others.

\subsubsection{Non-inferiority test}
The test statistics and p-values associated with non-inferiority tests with a non-inferiority margin of $\delta = 0.01$ are shown in Table~\ref{tab:noninfer}. To determine which null hypotheses are rejected, we used the Holm-Bonferroni method with a significance level of $\alpha = 0.05$. 

\begin{table}[ht]
\centering
\caption{ Test statistics, p-values, and conclusions of non-inferiority tests. Conclusions are obtained through the Holm-Bonferroni method \citep{holm1979}.}
\label{tab:noninfer}
\begin{tabular}{lrrr}
\toprule
\multicolumn{1}{l}{\textbf{Hyperparameter}}          & \multicolumn{1}{c}{\textbf{z}}              & \multicolumn{1}{c}{\textbf{p}}              & \multicolumn{1}{c}{\bm{$H_0$}}            \\ \midrule

\textbf{Random Forest}                                 & \multicolumn{1}{l}{} & \multicolumn{1}{l}{} & \multicolumn{1}{l}{} \\ \midrule
\multicolumn{1}{l}{\textbf{bootstrap}}               & \multicolumn{1}{r}{7.93}                    & \multicolumn{1}{r}{1.11 e-15}               & rejected                                      \\ 
\multicolumn{1}{l}{\textbf{criterion}}               & \multicolumn{1}{r}{8.16}                    & \multicolumn{1}{r}{1.11 e-16}               & rejected                                      \\ 
\multicolumn{1}{l}{\textbf{max\_features}}           & \multicolumn{1}{r}{-2.02}                   & \multicolumn{1}{r}{9.78 e-01}               & not rejected                                  \\ 
\multicolumn{1}{l}{\textbf{\begin{tabular}[c]{@{}l@{}}max\_features \\ (scikit-learn)\end{tabular}}} & \multicolumn{1}{r}{3.59}                    & \multicolumn{1}{r}{1.63 e-04}               & rejected                                      \\ 
\multicolumn{1}{l}{\textbf{min\_samples\_leaf}}      & \multicolumn{1}{r}{15.04}                   & \multicolumn{1}{r}{0.00 e+00}               & rejected                                      \\ 
\multicolumn{1}{l}{\textbf{min\_samples\_split}}     & \multicolumn{1}{r}{10.06}                   & \multicolumn{1}{r}{0.00 e+00}               & rejected                                      \\ \midrule

\textbf{SVM}                                           & \multicolumn{1}{l}{} & \multicolumn{1}{l}{} & \multicolumn{1}{l}{} \\ \midrule
\multicolumn{1}{l}{\textbf{C}}                       & \multicolumn{1}{r}{1.83}                    & \multicolumn{1}{r}{3.40 e-02}               & not rejected                                  \\ 
\multicolumn{1}{l}{\textbf{gamma}}                   & \multicolumn{1}{r}{-8.24}                   & \multicolumn{1}{r}{1.00 e+00}               & not rejected                                  \\ 
\multicolumn{1}{l}{\textbf{gamma (scikit-learn)}}         & \multicolumn{1}{r}{-4.61}                   & \multicolumn{1}{r}{1.00 e+00}               & not rejected                                  \\ 
\multicolumn{1}{l}{\textbf{shrinking}}               & \multicolumn{1}{r}{20.49}                   & \multicolumn{1}{r}{0.00 e+00}               & rejected                                      \\ 
\multicolumn{1}{l}{\textbf{tol}}                     & \multicolumn{1}{r}{16.32}                   & \multicolumn{1}{r}{0.00 e+00}               & rejected                                      \\ \bottomrule
\end{tabular}
\end{table}

The results show that for all hyperparameters, except for \texttt{max\_features} (computed default), \texttt{C}, and \texttt{gamma} (both computed and scikit-learn default) the relative risk is not more than 1\% higher in the fixed condition compared to the non-fixed condition. This indicates that for these hyperparameters, setting the hyperparameter to the default value is non-inferior to tuning the parameter through a random search using 100 iterations. For \texttt{C}, \texttt{gamma}, and \texttt{max\_features} (computed default) this is not the case.

\section{Conclusions}

\label{sec:concl}

In this study, we have first presented a simple heuristic to find default hyperparameters. In our experiments, we touched upon determining meta-feature dependent default parameters. Additionally, we presented a methodology for determining the importance of tuning a hyperparameter empirically. In contrast to previous work, our approach can be used to determine the loss incurred when one of the hyperparameters is not tuned but set to a default value. This is different from the study of \citet{Probst2018}, who investigate the performance gain of leaving all hyperparameters to a default value and tuning one hyperparameter. We have applied our methodology in a benchmark study using 59 different datasets. In this way, we provide empirical evidence that can be consulted by machine learning practitioners before they start a computationally and time intensive tuning process.

Our results show that using our computed default value often results in non-inferior performance compared to tuning the hyperparameter. It should be noted that the number of iterations turned out to be too low to tune several configurations of the algorithms to the best possible performance in the non-fixed condition. Although this indicates that our default parameters were reasonable, the number of iterations could be increased in future work. For other hyperparameters, such as random forest's \texttt{max\_features} and SVM's \texttt{gamma} and \texttt{C}, we observed a high tuning risk and relative tuning risk, which suggests that it is important to tune these hyperparameters.

A limitation of our work is that the default parameters are only determined once and are not validated separately. As a result, it is unclear how our simple default parameter estimation method affects the results of the second experiment. In particular, it is unclear whether our choice of $n$ results in a good representation of hyperparameters with good performance. This could be resolved in future work by comparing different methods for finding default parameters. Moreover, it would be interesting to compare the effect of particular performance measures, e.g. accuracy and AUC, on computed default values.

Another limitation is that we excluded several datasets from the OpenML-CC18 that turned out to have a high number of instances and/or features. It is unclear how this affects our conclusions, in particular for the determination of meta-feature dependent default values. A related issue is that it is unclear whether the OpenML-CC18 is a good representation of datasets that machine learning practitioners encounter in real life. Our conclusions might not hold for datasets that are very different from the ones analyzed in this work.

Finally, it is important to note that we have only considered leaving a single hyperparameter at a default value. In future work, interactions between hyperparameters could be further investigated.

\begin{acks}
We thank Microsoft Azure for providing the computational resources for this study.
\end{acks}

\bibliographystyle{ACM-Reference-Format}
\bibliography{references}

\onecolumn
\appendix
\clearpage

\section{Datasets}
\subsection{Properties OpenML-CC18}
\label{app:cc18}
The OpenML-CC18 is a curated benchmark suite consisting of 73 classification tasks. For each dataset, the following properties hold:

\begin{itemize}
    \item The number of observations is larger than 500;
    \item The number of observations is smaller than 100,000;
    \item The ratio of the minority class and the majority class is larger than 0.05;
    \item The number of values for categorical features most not exceed 100;
    \item The classification problem is not trivial (a model based on 1 feature does not result in perfect performance);
    \item Each target class contains at least 20 instances;
    \item The dataset does not belong to one of the following categories:
    \begin{itemize}
        \item Artificial dataset
        \item Simulated dataset
        \item Time series dataset
        \item Text data
        \item Multilabel data
        \item Derived versions of another dataset
        \item Dataset where the intended classification target is unclear
        \item Binarized regression problem
        \item Dataset of unknown origin
        \item Grouped data
    \end{itemize}
\end{itemize}

\subsection{Excluded tasks}
\label{app:cc18ni}
\begin{figure}[ht]
    \centering
    \includegraphics[width = 0.3\linewidth]{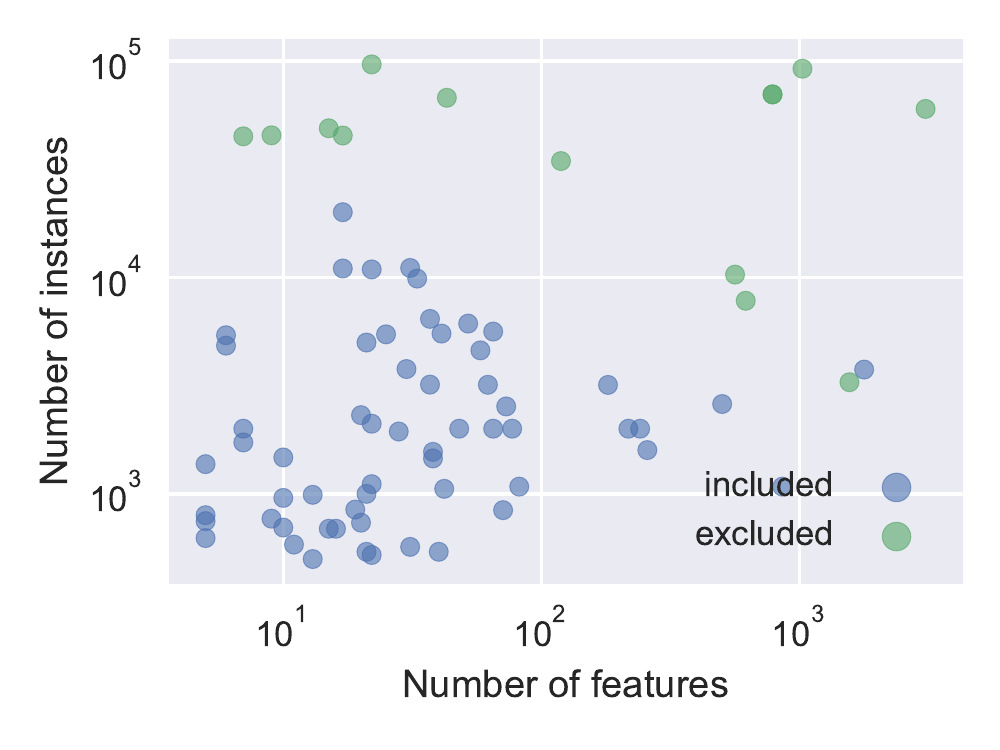}
    \caption{Number of features against number of instances of datasets of the OpenML-CC18 that were included and excluded from this study.}
    \label{fig:inexcl}
\end{figure}

The tasks of OpenML-CC18 that were not included in this study, due to either time constraints or technical issues, are listed in Table~\ref{tab:excl}. Additionally, the number of features and number of instances of the datasets are shown in Figure~\ref{fig:inexcl}. It is clear that the datasets that were excluded are all datasets with relatively many features or instances.

\begin{table}[ht]
\centering

\caption{ OpenML-CC18 tasks excluded from this study. Tasks for which only the random forest or SVM timed out too often are annotated with respectively * and **.}
\label{tab:excl}
\begin{tabular}{ll}
\midrule
\textbf{OpenML task id} & \textbf{Reason} \\ \midrule
167125                  & technical issue \\ \midrule
219**                    & time constraint \\ \midrule
3481*                    & time constraint \\ \midrule
3573                   & time constraint \\ \midrule
7592                    & time constraint \\ \midrule
9977                    & time constraint \\ \midrule
14965**                 & time constraint \\ \midrule
14970*                   & time constraint \\ \midrule
146195                  & time constraint \\ \midrule
146825                  & time constraint \\ \midrule
167119**                  & time constraint \\ \midrule
167120                  & time constraint \\ \midrule
167121                  & time constraint \\ \midrule
167124                  & time constraint \\ \midrule
\end{tabular}
\end{table}

\subsection{Datasets previous work}
\label{app:cc18co}

\begin{figure}[!ht]
\captionsetup[subfigure]{justification=centering}
\centering

\begin{subfigure}[t]{0.25\textheight}
\vskip 0pt
    \centering
    \includegraphics[width=\textwidth]{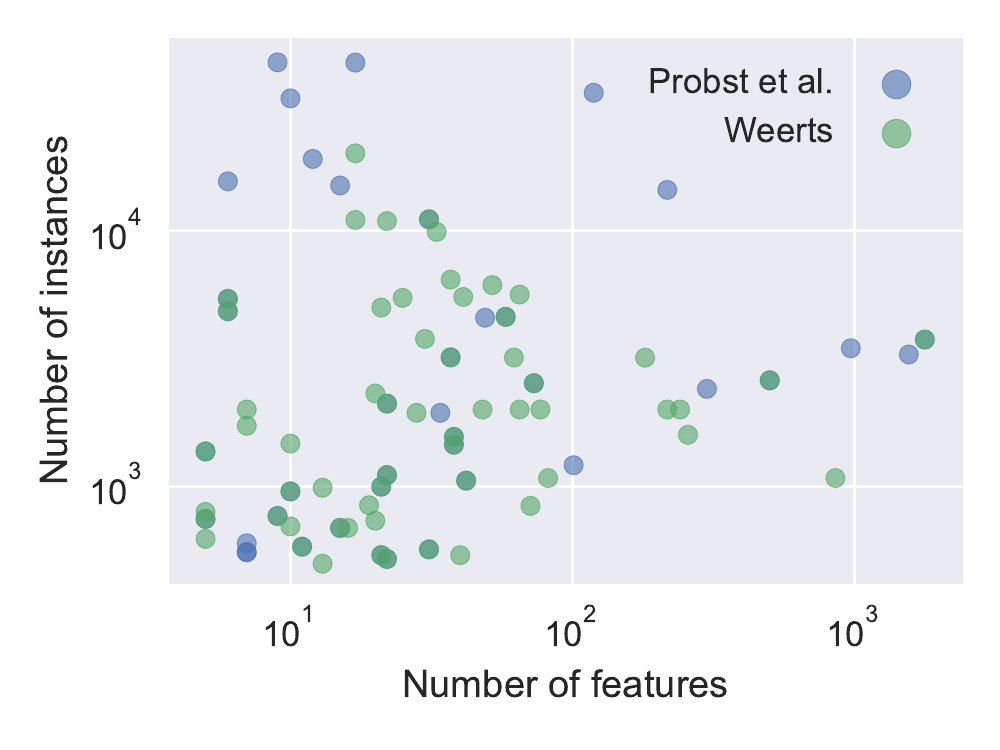}
    \caption{ The number of features against number of instances for this work and \citet{Probst2018}.}\label{fig:scatterfeatures}
\end{subfigure}
    ~
\begin{subfigure}[t]{0.25\textheight}
\vskip 0pt
    \centering
    \includegraphics[width=\textwidth]{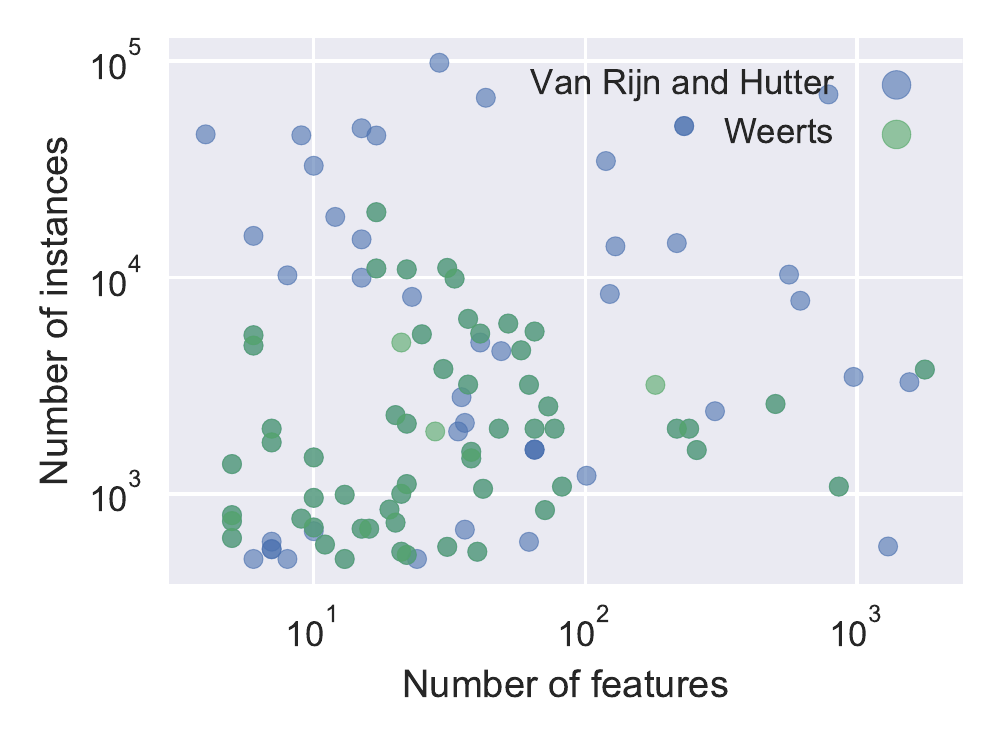}
    \caption{ The number of features against number of instances for this work and \citet{Vanrijn2018}.}\label{fig:scatterfeatures2}
\end{subfigure}
~
\begin{subfigure}[t]{.25\textheight}
\vskip 0pt
    \centering
    \includegraphics[width=\textwidth]{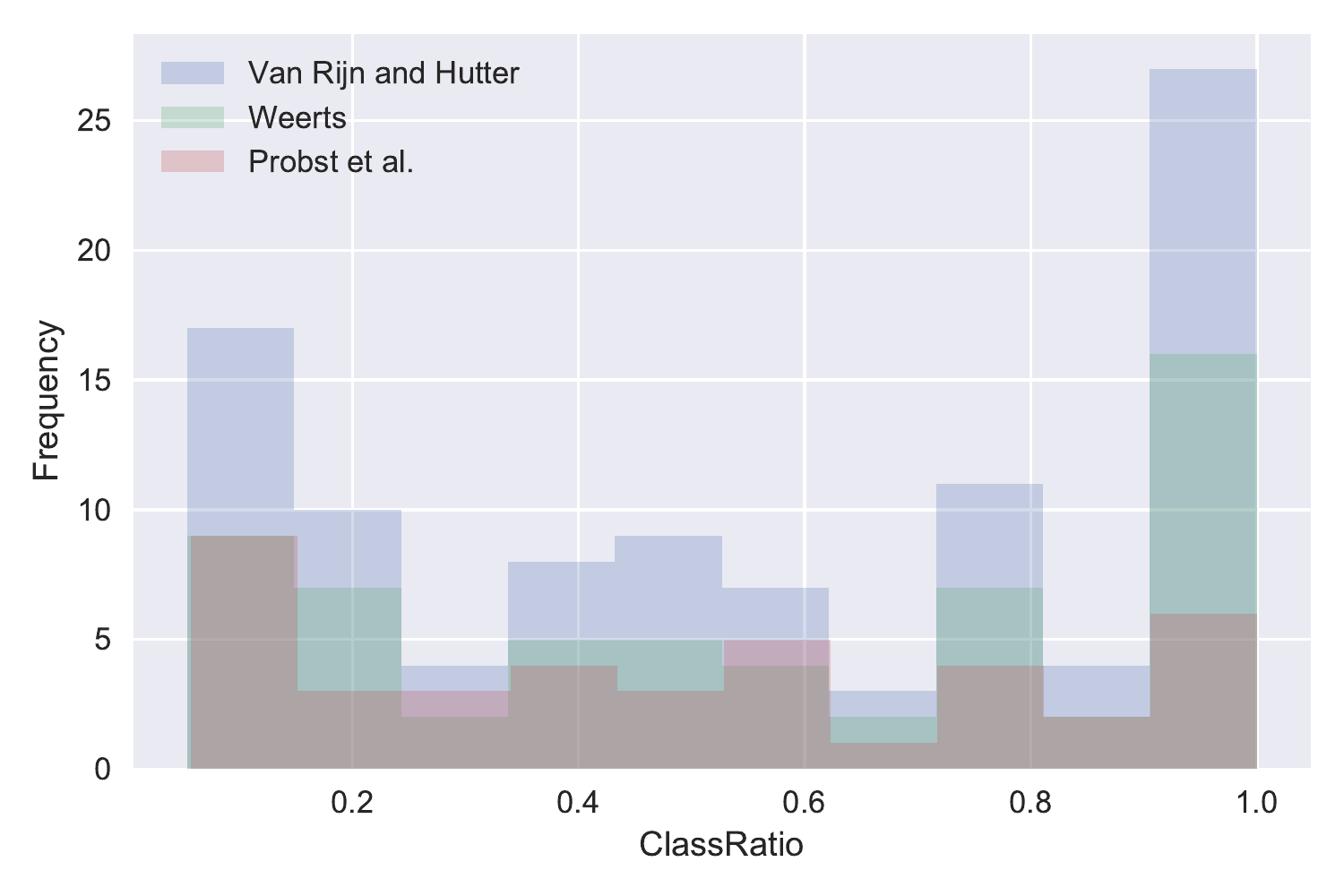}
    \caption{ Distribution of the ratio of the number of instances in the minority class compared to the majority class.}
    \label{fig:weertsprobstratio}
    \end{subfigure}
    ~

\caption{ Meta-features for the datasets included in the work of \citet{Probst2018}, \citet{Vanrijn2018} and this study.}
\label{fig:previouswork}
\end{figure}

In Figure~\ref{fig:scatterfeatures} we observe that both the study of \citet{Probst2018} and \citet{Vanrijn2018} contain more datasets with a large number of features and instances. In Figure~\ref{fig:weertsprobstratio} we observe that the data of \citet{Probst2018} and \citet{Vanrijn2018} contain relatively more imbalanced dataset than ours. This makes sense, since these datasets are taken from the OpenML-100, for which the class imbalance was not yet one of the criteria (as opposed to the OpenML-CC18).

\clearpage
\section{Distribution of top 10 performance data}
\label{app:top10hists}
\begin{figure}[h]
\captionsetup[subfigure]{justification=centering}
\centering
\begin{subfigure}{.3\linewidth}
    \centering
    \includegraphics[width=\textwidth]{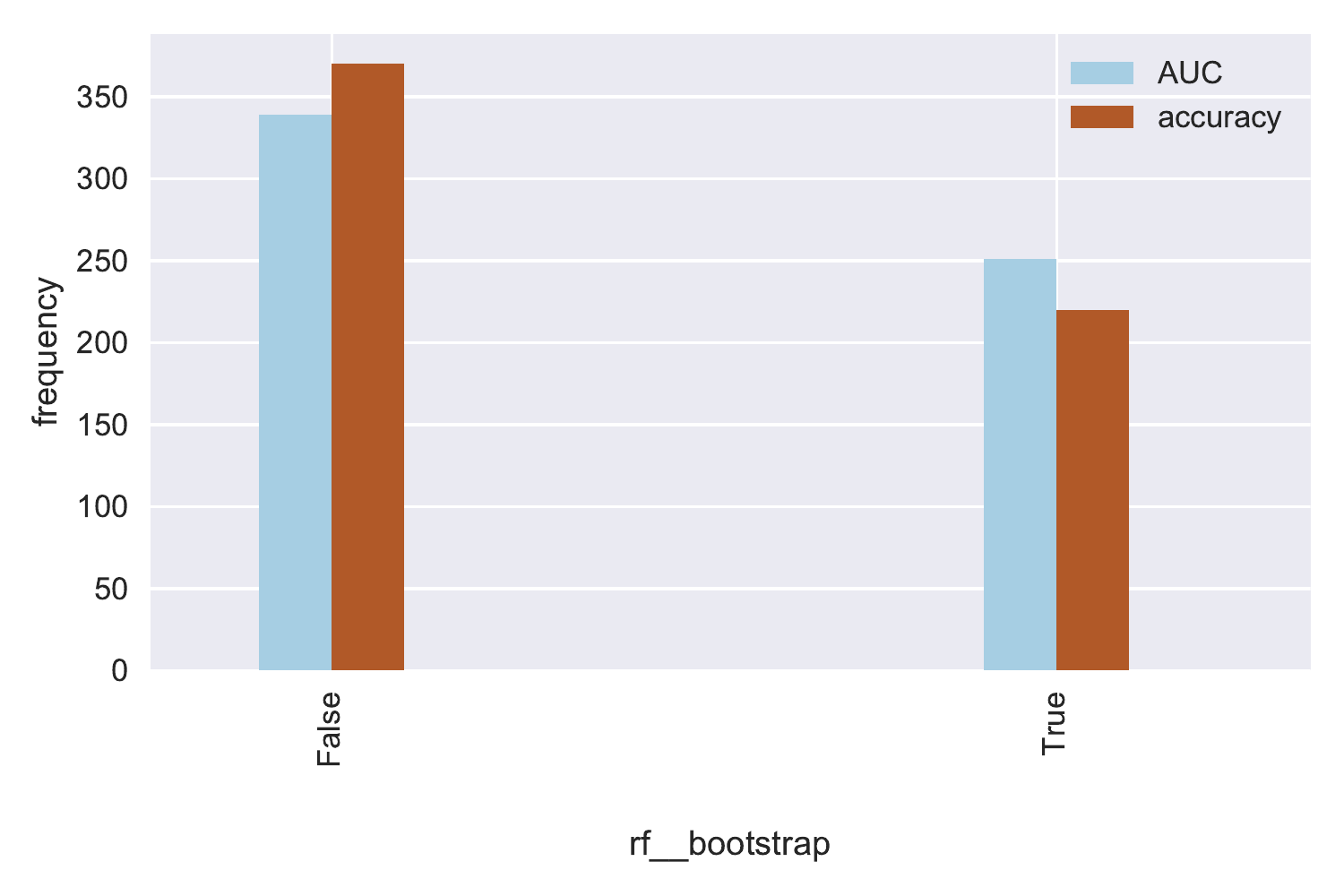}
    \caption{Random Forest - \texttt{bootstrap}}\label{fig:bootstrap}
\end{subfigure}
    \hfill
\begin{subfigure}{.3\linewidth}
    \centering
    \includegraphics[width=\textwidth]{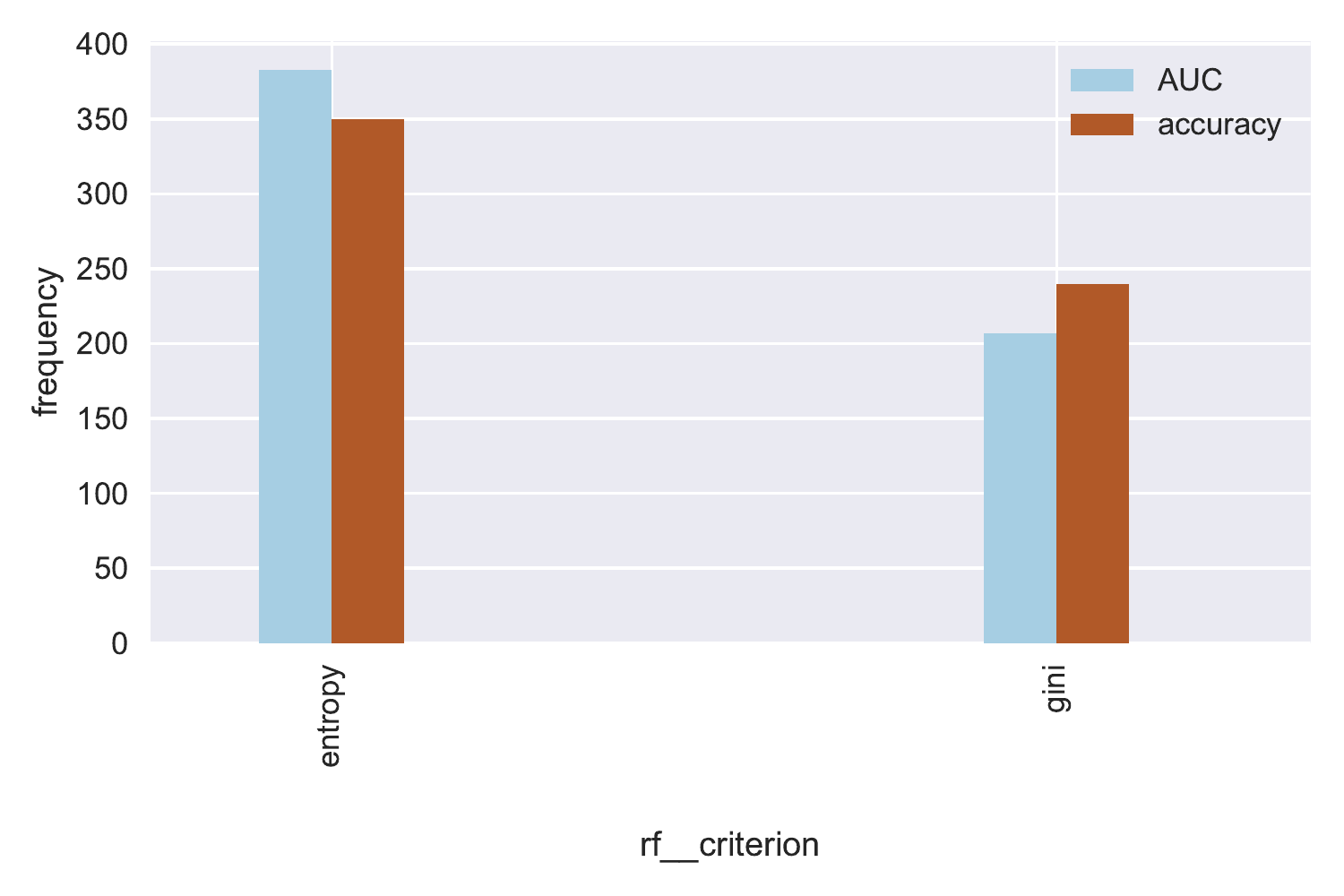}
    \caption{Random Forest - \texttt{criterion}}\label{fig:criterion}
\end{subfigure}
   \hfill
\begin{subfigure}{.3\linewidth}
    \centering
    \includegraphics[width=\textwidth]{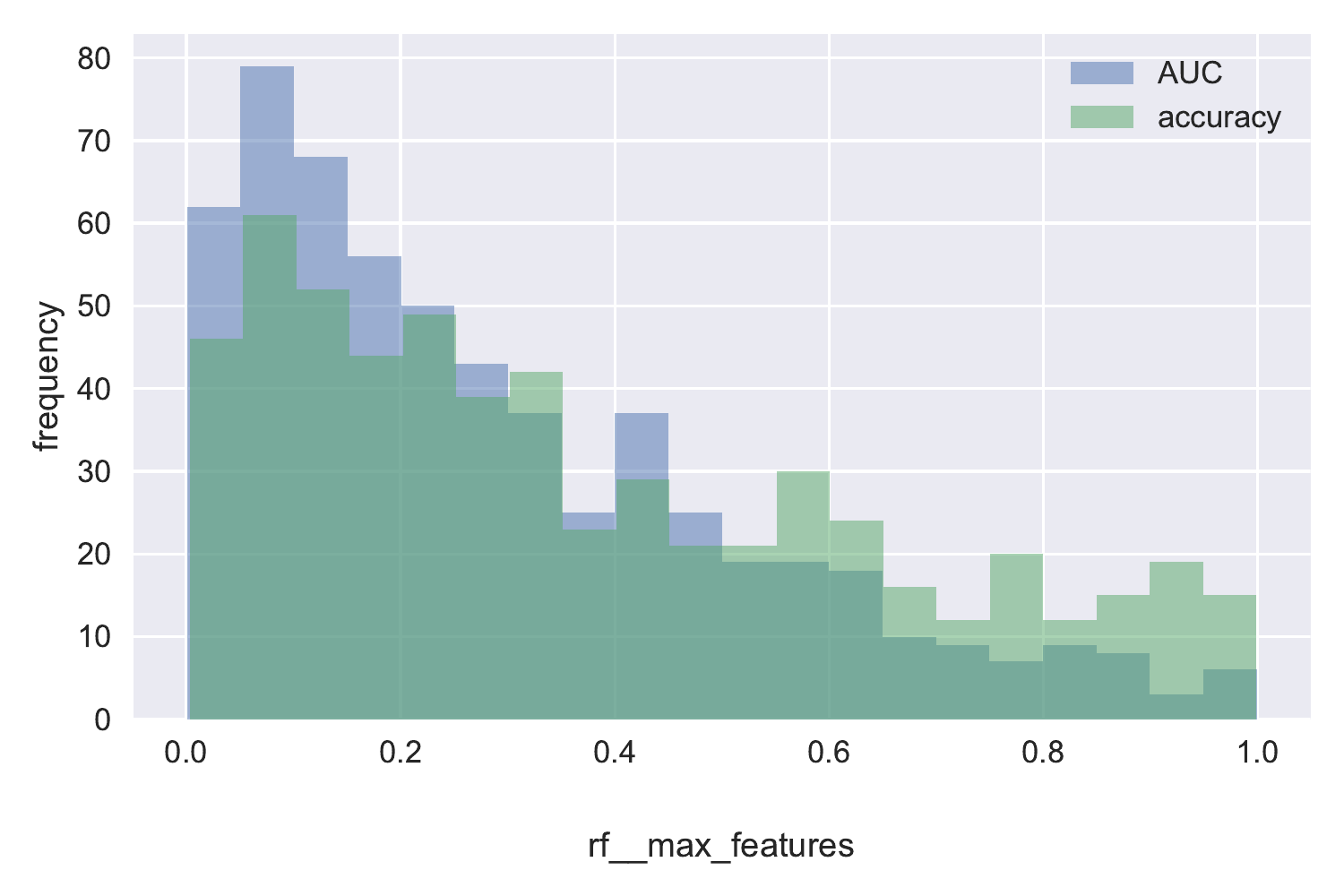}
    \caption{Random Forest - \texttt{max\_features}}\label{fig:maxfeatures}
\end{subfigure}

\bigskip

\begin{subfigure}{.3\linewidth}
    \centering
    \includegraphics[width=\textwidth]{figures/hist__rf__min_samples_leaf.pdf}
    \caption{ Random Forest - \texttt{min\_samples\_leaf}}\label{fig:minsamplesleaf}
\end{subfigure}
    \hfill
\begin{subfigure}{.3\linewidth}
    \centering
    \includegraphics[width=\textwidth]{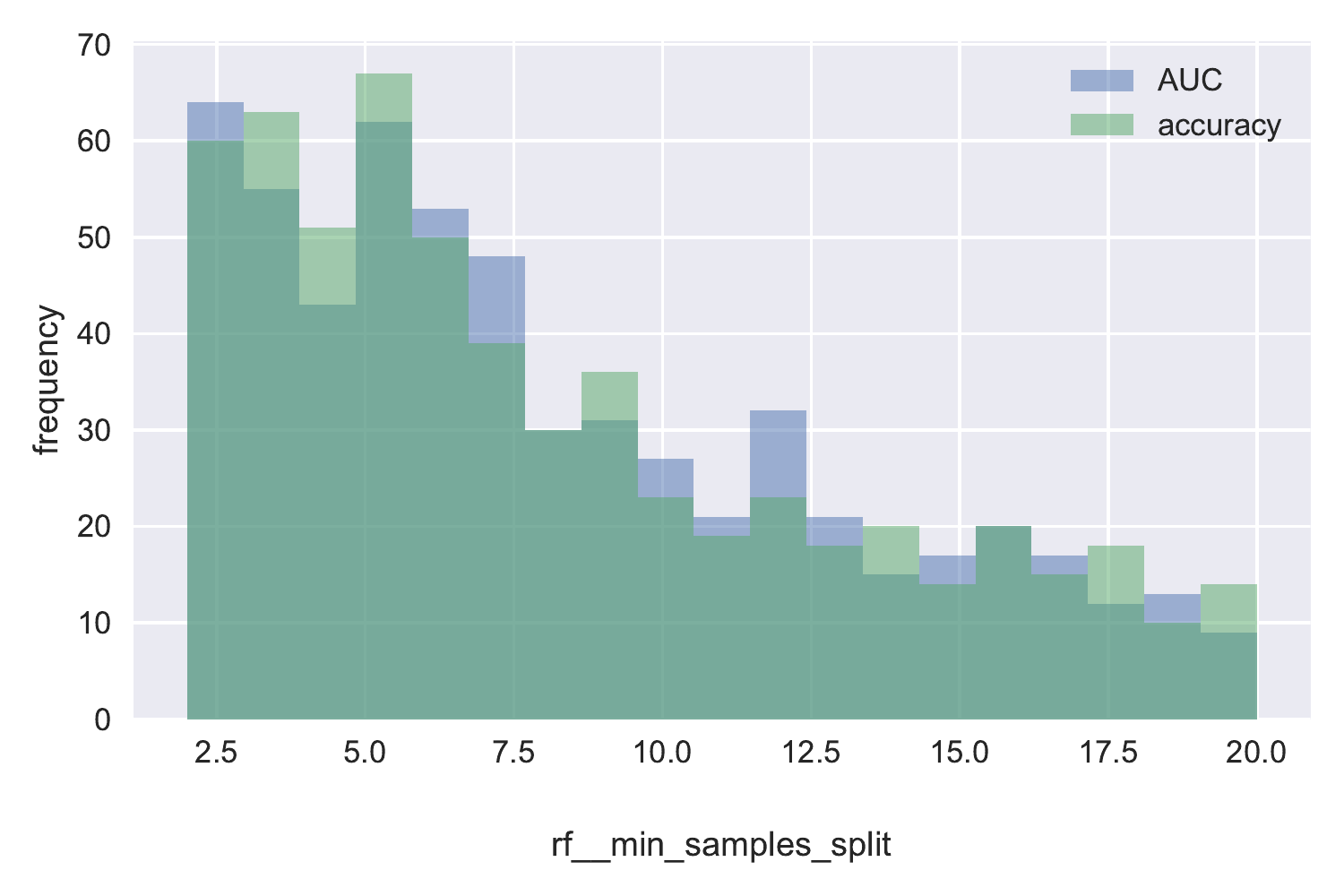}
    \caption{Random Forest - \texttt{min\_samples\_split}}\label{fig:minsamplessplit}
\end{subfigure}
   \hfill
\begin{subfigure}{.3\linewidth}
    \centering
    \includegraphics[width=\textwidth]{figures/hist__svm__C.pdf}
    \caption{SVM - \texttt{C}}\label{fig:c}
\end{subfigure}

\bigskip

\begin{subfigure}{.3\linewidth}
    \centering
    \includegraphics[width=\textwidth]{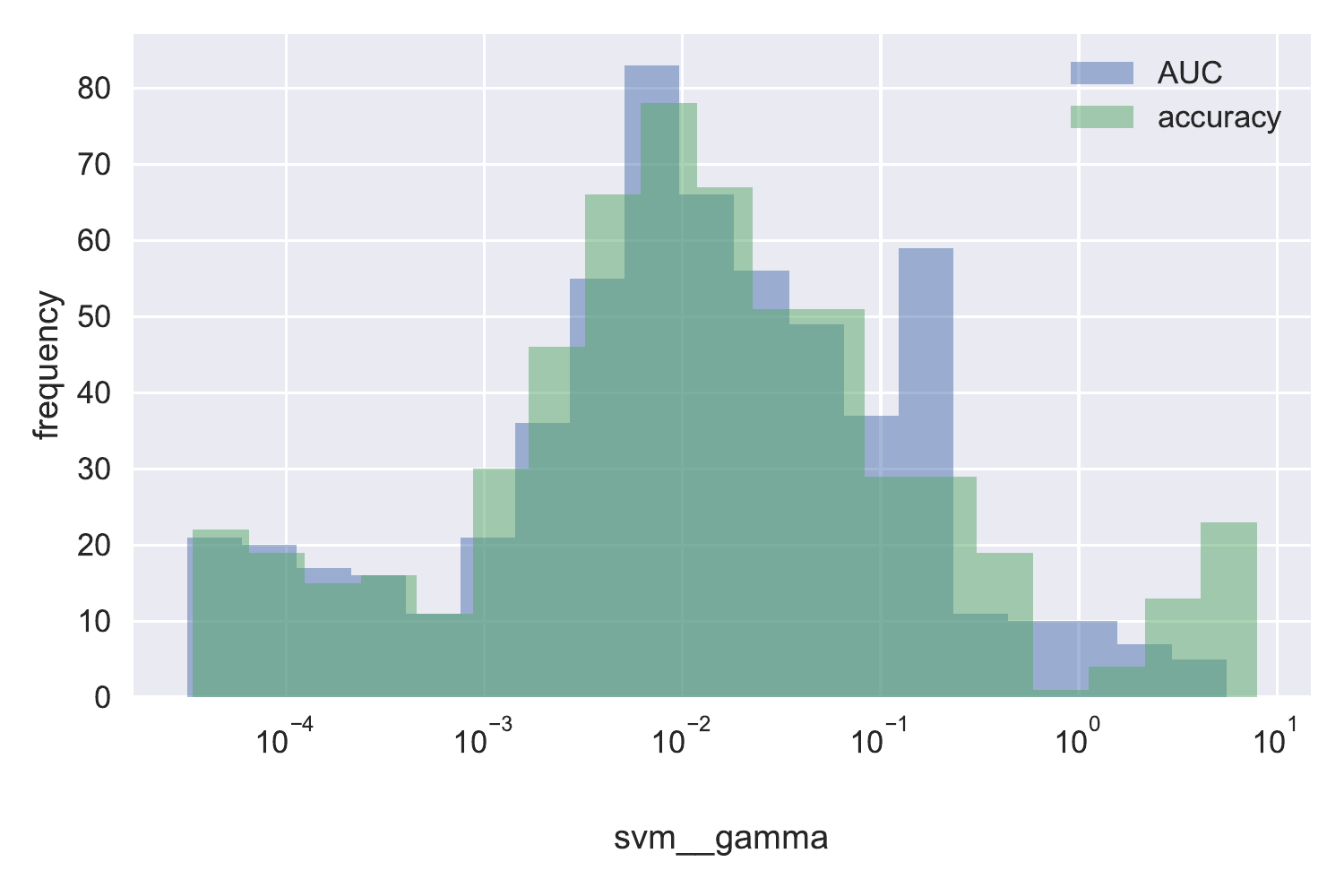}
    \caption{SVM - \texttt{gamma}}\label{fig:gamma}
\end{subfigure}
    \hfill
\begin{subfigure}{.3\linewidth}
    \centering
    \includegraphics[width=\textwidth]{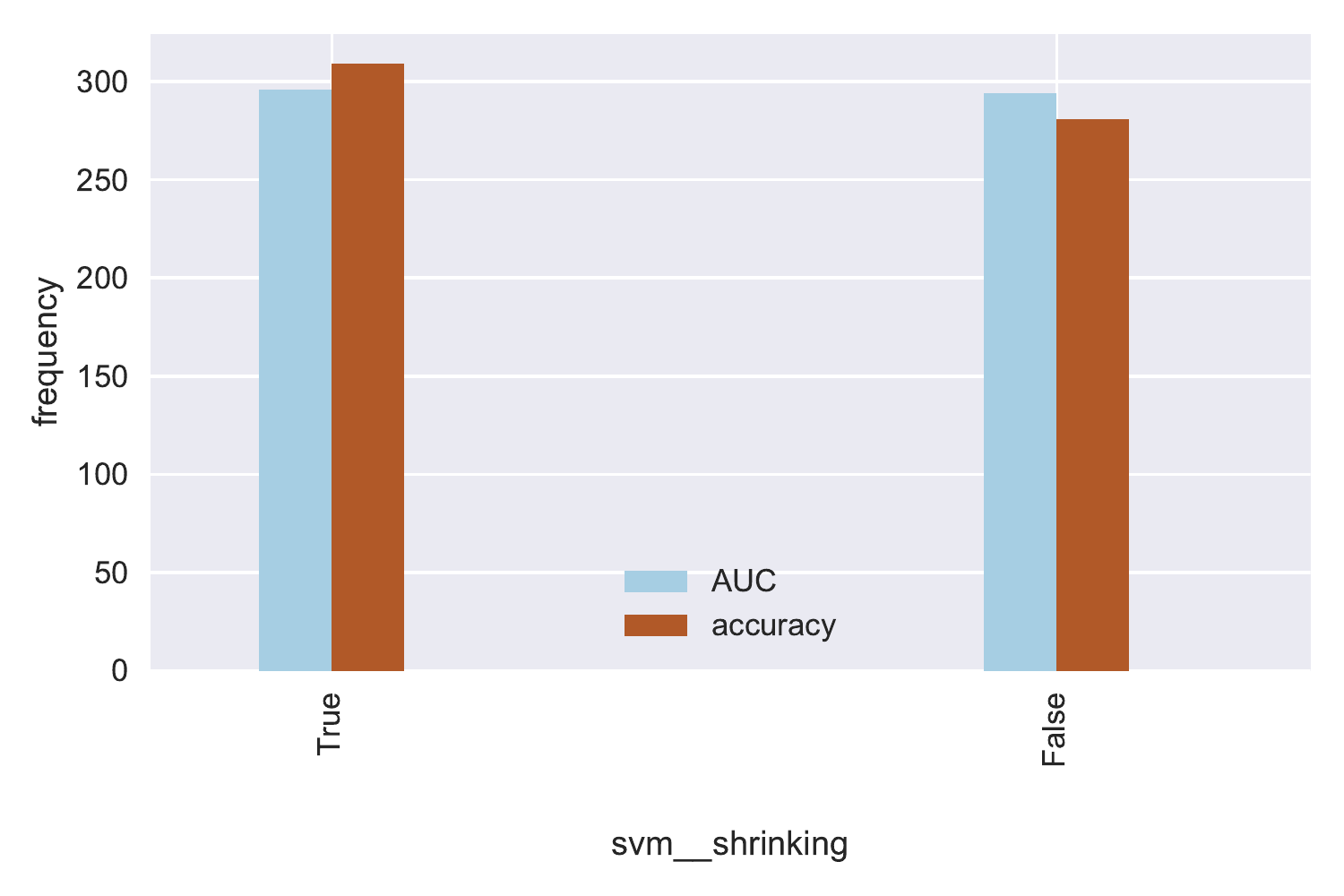}
    \caption{SVM - \texttt{shrinking}}\label{fig:shrinking}
\end{subfigure}
   \hfill
\begin{subfigure}{.3\linewidth}
    \centering
    \includegraphics[width=\textwidth]{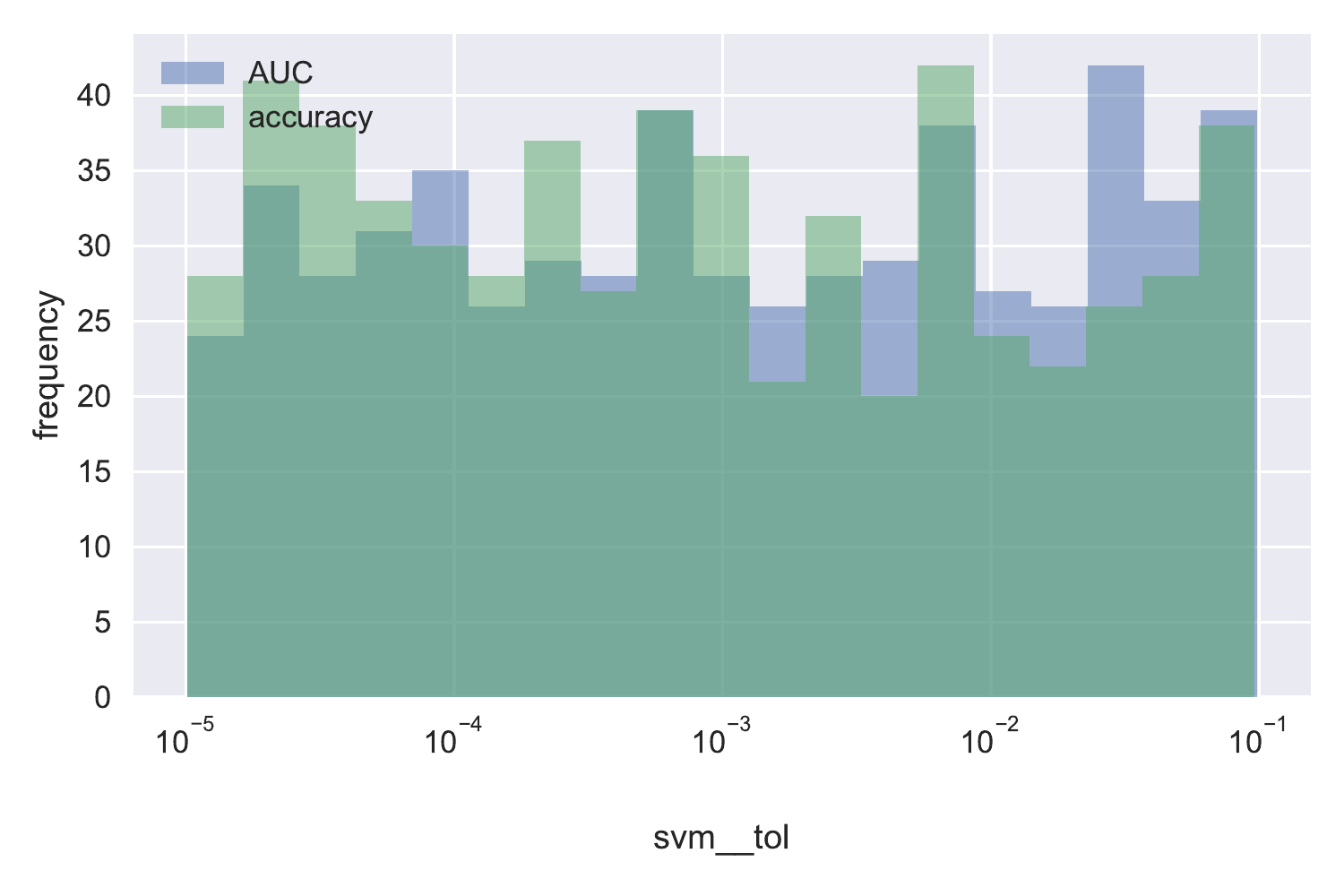}
    \caption{SVM - \texttt{tol}}\label{fig:tol}
\end{subfigure}

\caption{ The distribution of a hyperparameters within the top 10 highest performing hyperparameter settings per dataset, measured by either accuracy or AUC.}
\label{fig:histograms}
\end{figure}

In Figure~\ref{fig:histograms} histograms of the distribution of hyperparameters in the top 10 performance data are displayed for both accuracy and AUC.

\clearpage
\section{Distribution of default values across tasks}
\label{app:defval}
\begin{figure}[!ht]
\captionsetup[subfigure]{justification=centering}
\centering

\begin{subfigure}{0.3\linewidth}
    \centering
    \includegraphics[width=\textwidth]{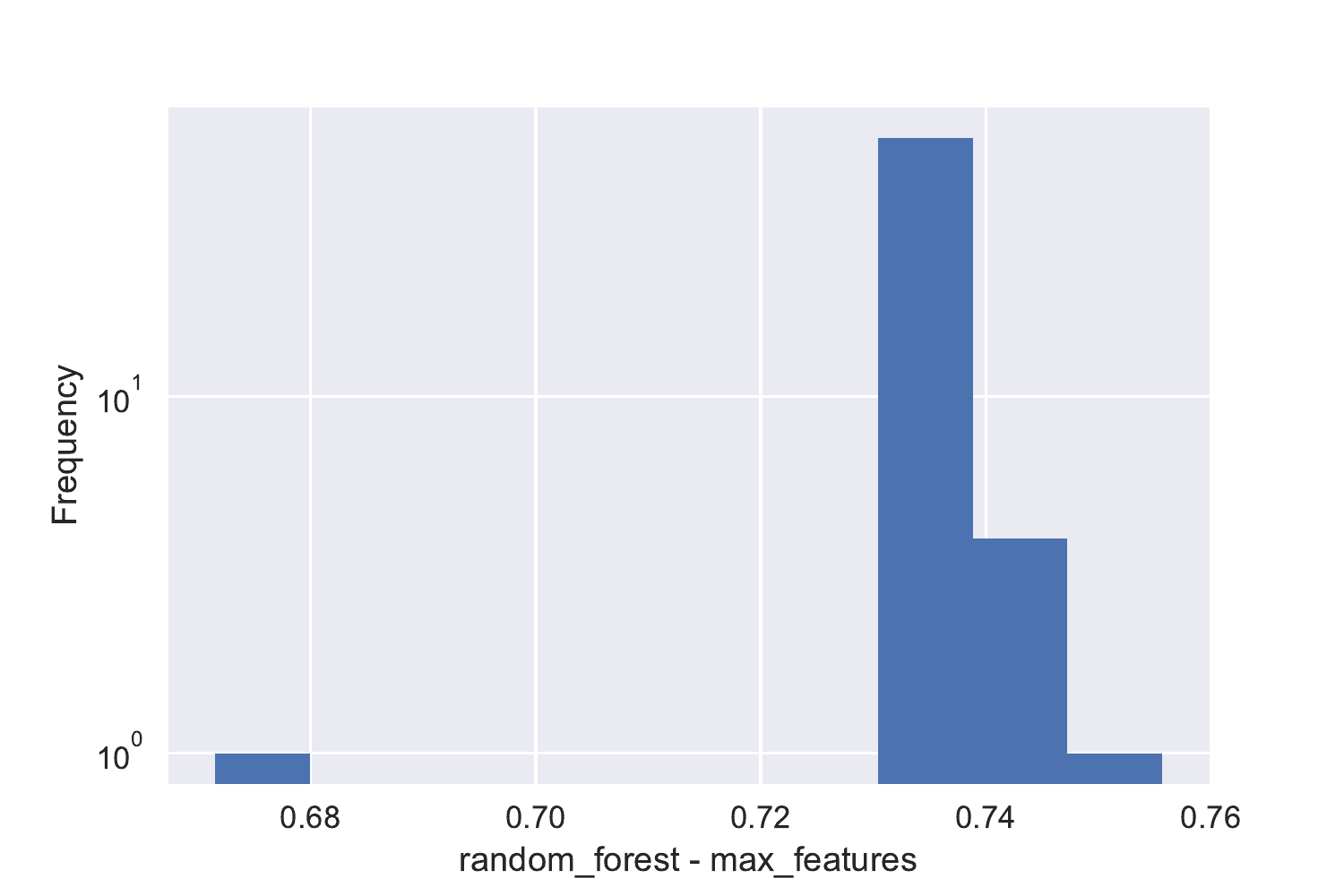}
    \caption{Random Forest - \texttt{max\_features}}\label{fig:histmaxfeatures}
\end{subfigure}
    ~
\begin{subfigure}{.3\linewidth}
    \centering
    \includegraphics[width=\textwidth]{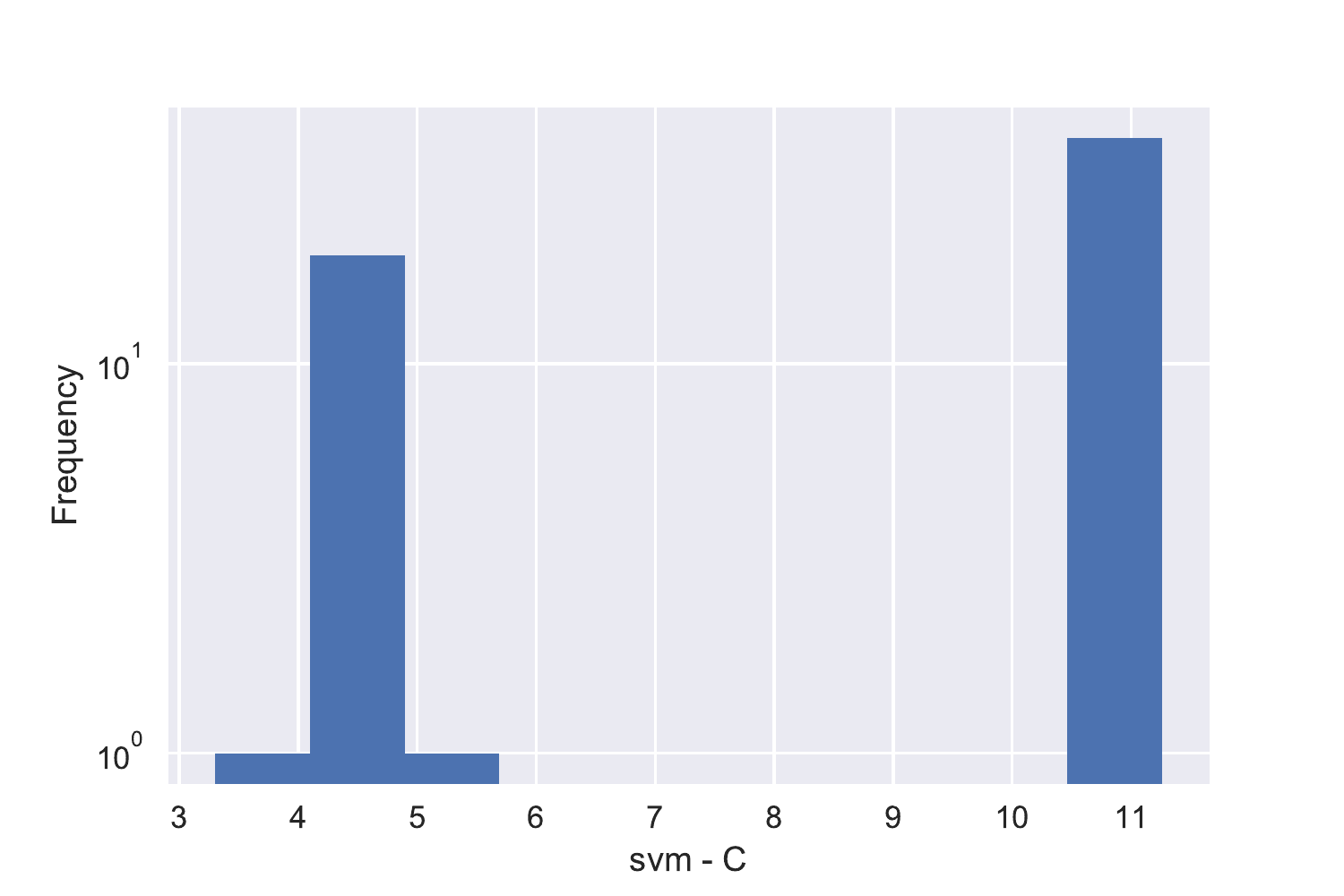}
    \caption{SVM - \texttt{criterion}}\label{fig:histC}
    \end{subfigure}
    
\bigskip

\begin{subfigure}{.3\linewidth}
    \centering
    \includegraphics[width=\textwidth]{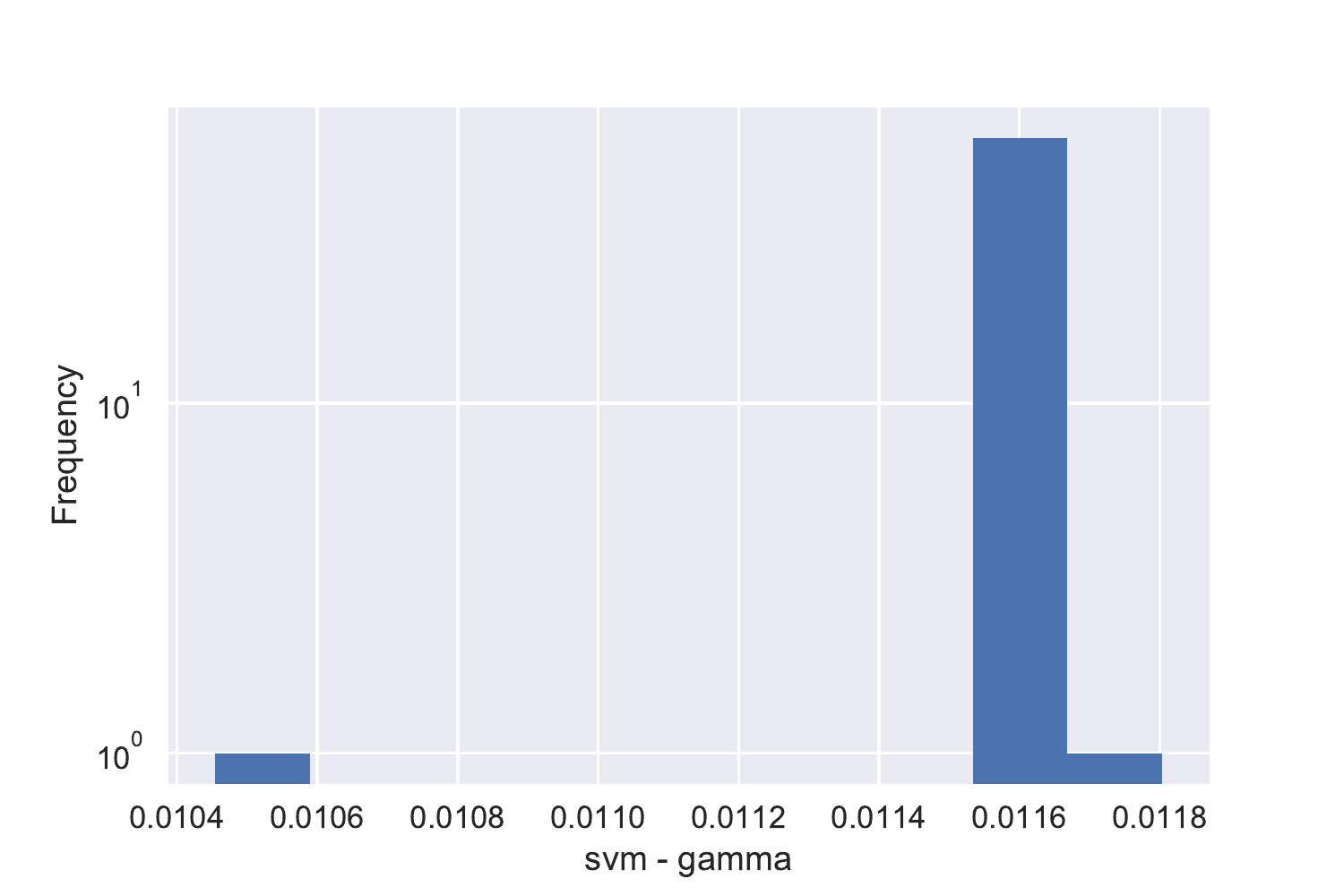}
    \caption{ SVM - \texttt{gamma}}\label{fig:histgamma}
\end{subfigure}
    ~
\begin{subfigure}{.3\linewidth}
    \centering
    \includegraphics[width=\textwidth]{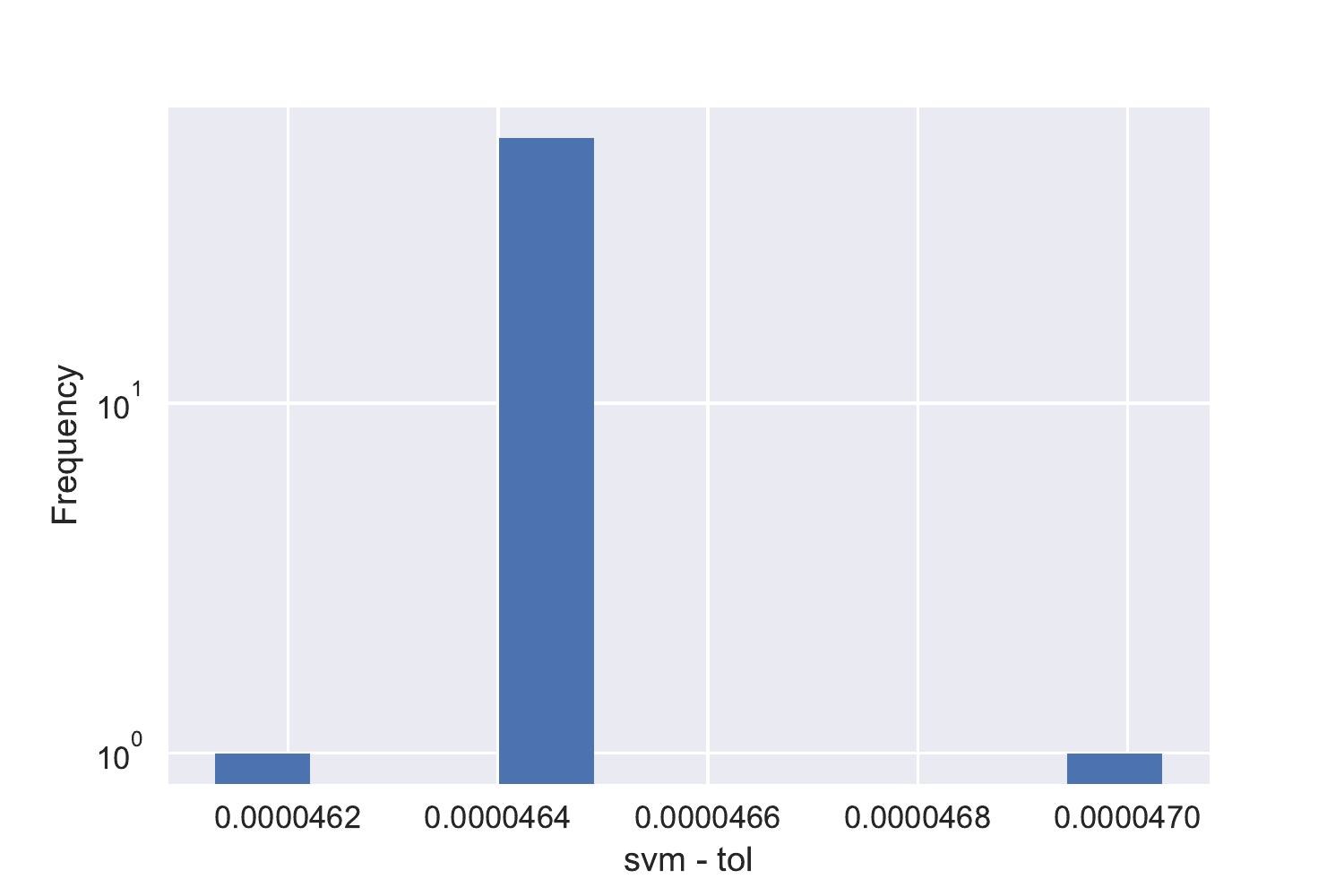}
    \caption{SVM - \texttt{tol}}\label{fig:histtol}
\end{subfigure}

\caption{ The distribution of the computed default values, using \textbf{accuracy} based performance data, across different tasks for hyperparameters where the default values have a non-zero standard deviation.}
\label{fig:distrdefvalacc}
\end{figure}

\begin{figure}[!ht]
\captionsetup[subfigure]{justification=centering}
\centering

\begin{subfigure}[t]{0.3\linewidth}
    \centering
    \includegraphics[width=\textwidth]{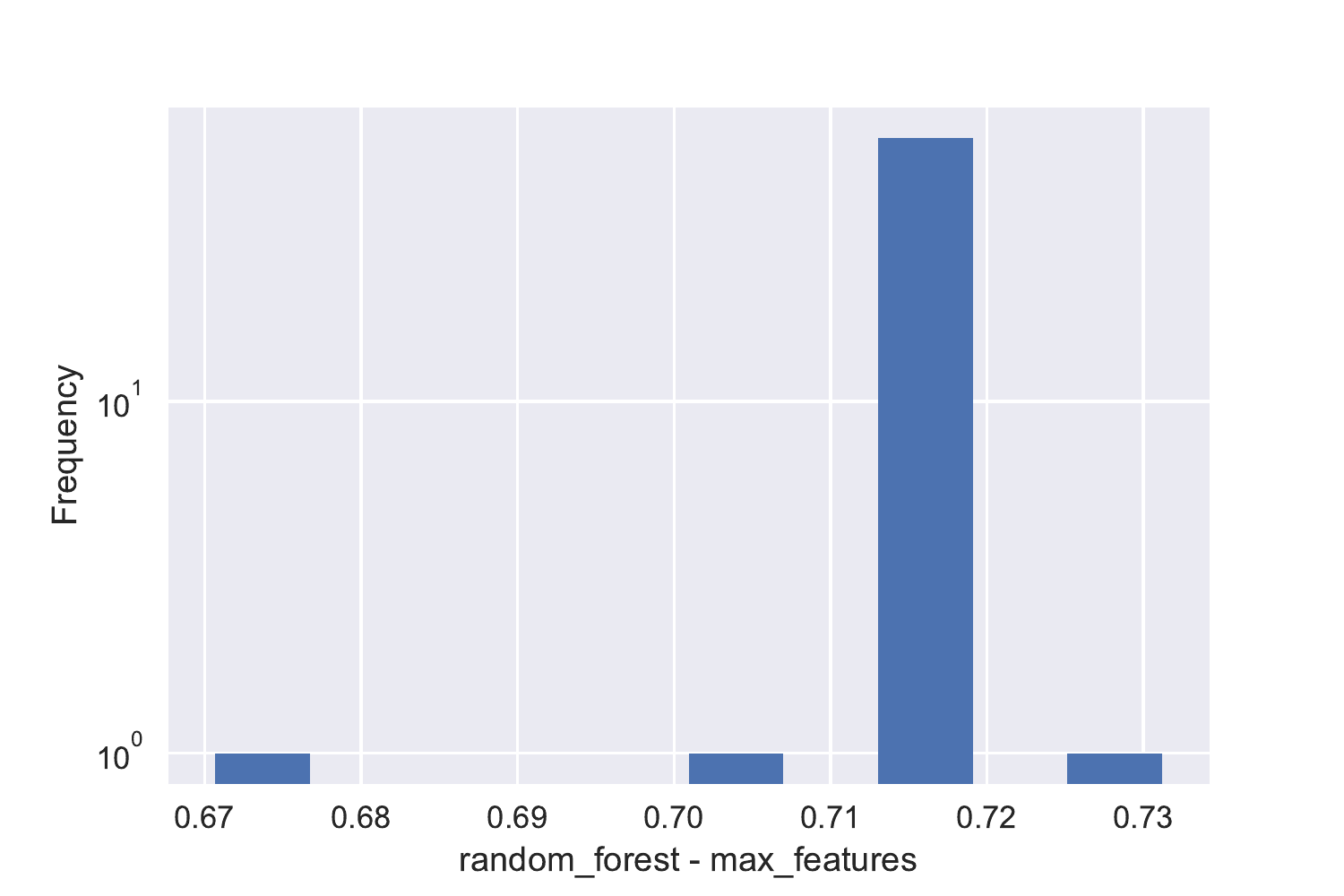}
    \caption{Random Forest - \texttt{max\_features}}\label{fig:histmaxfeaturesauc}
\end{subfigure}
    ~
\begin{subfigure}[t]{.3\linewidth}
    \centering
    \includegraphics[width=\textwidth]{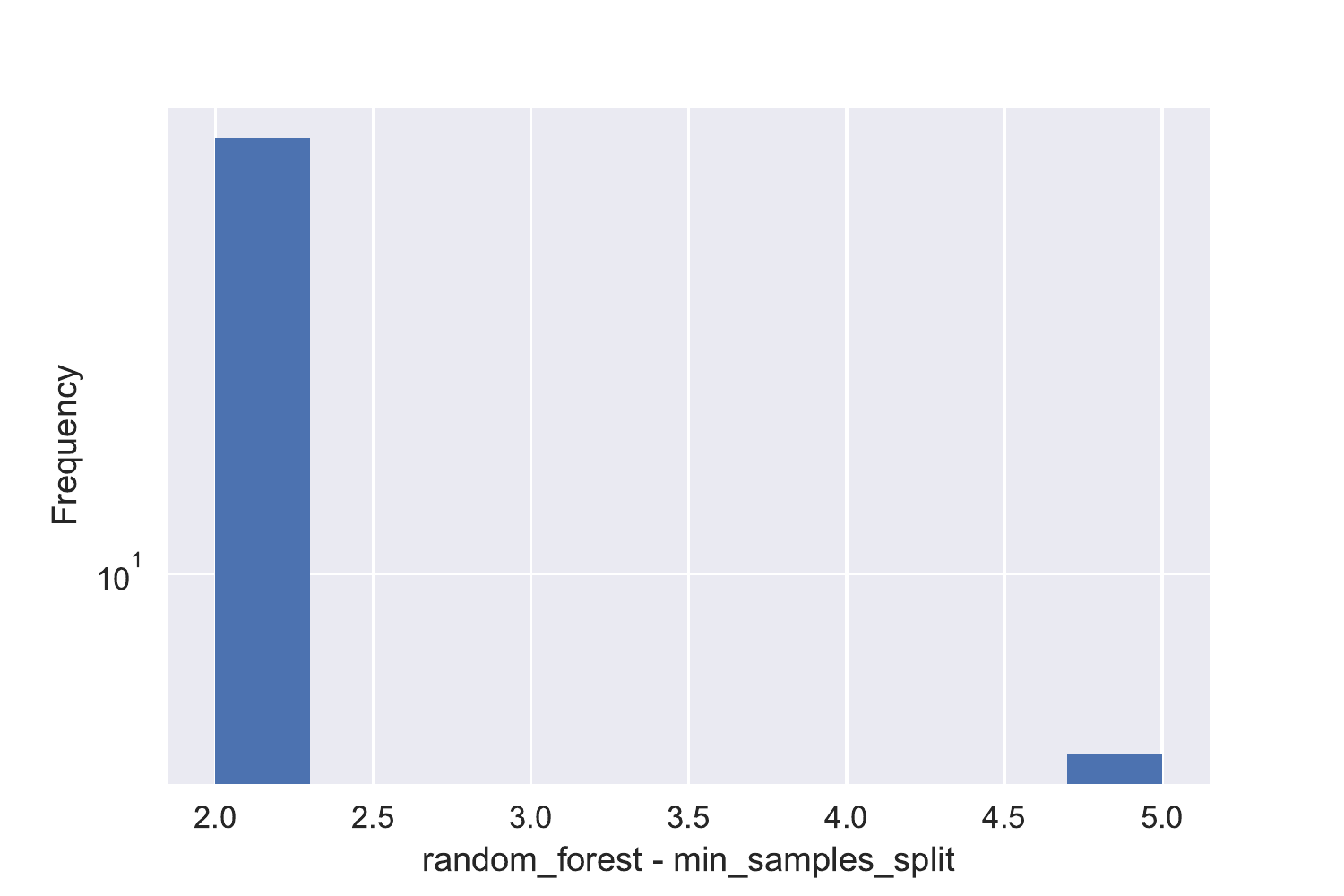}
    \caption{Random Forest - \texttt{min\_samples\_split}}\label{fig:histminsamplessplitauc}
    \end{subfigure}
    ~
\begin{subfigure}[t]{.3\linewidth}
    \centering
    \includegraphics[width=\textwidth]{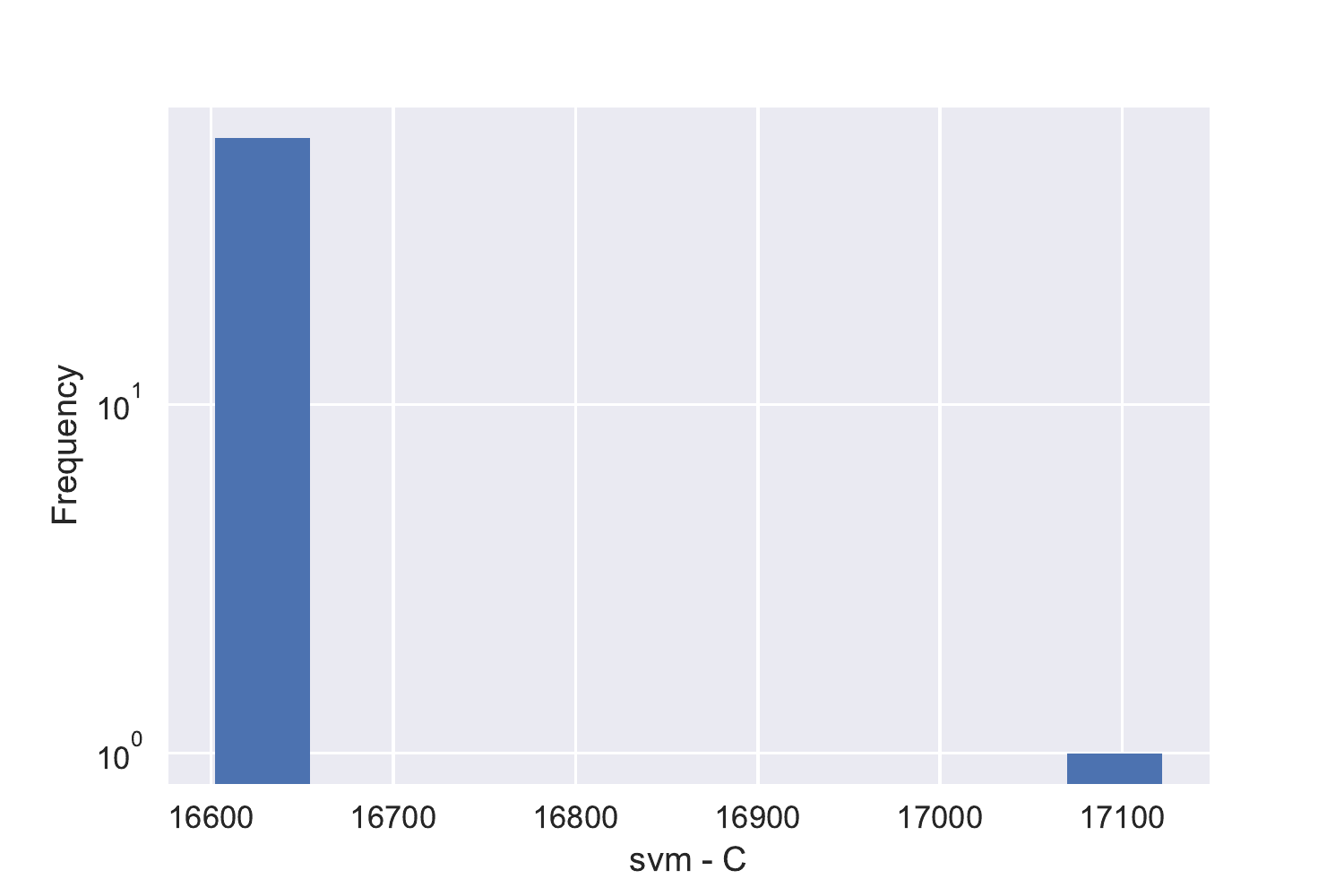}
    \caption{SVM - \texttt{criterion}}\label{fig:histCauc}
    \end{subfigure}
\bigskip
\begin{subfigure}[t]{.3\linewidth}
    \centering
    \includegraphics[width=\textwidth]{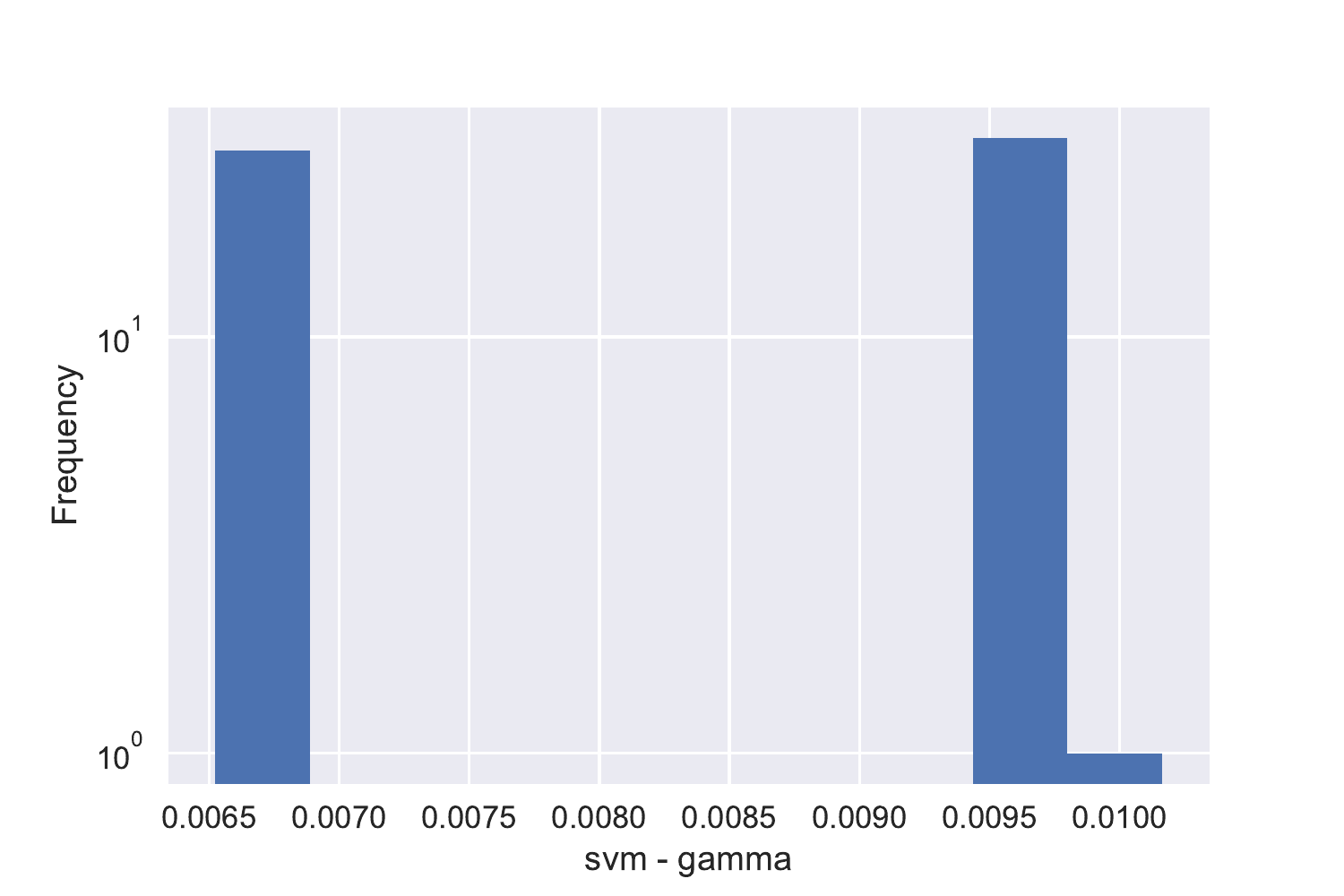}
    \caption{SVM - \texttt{gamma}}\label{fig:histgammaauc}
\end{subfigure}
~
\begin{subfigure}[t]{.3\linewidth}
    \centering
    \includegraphics[width=\textwidth]{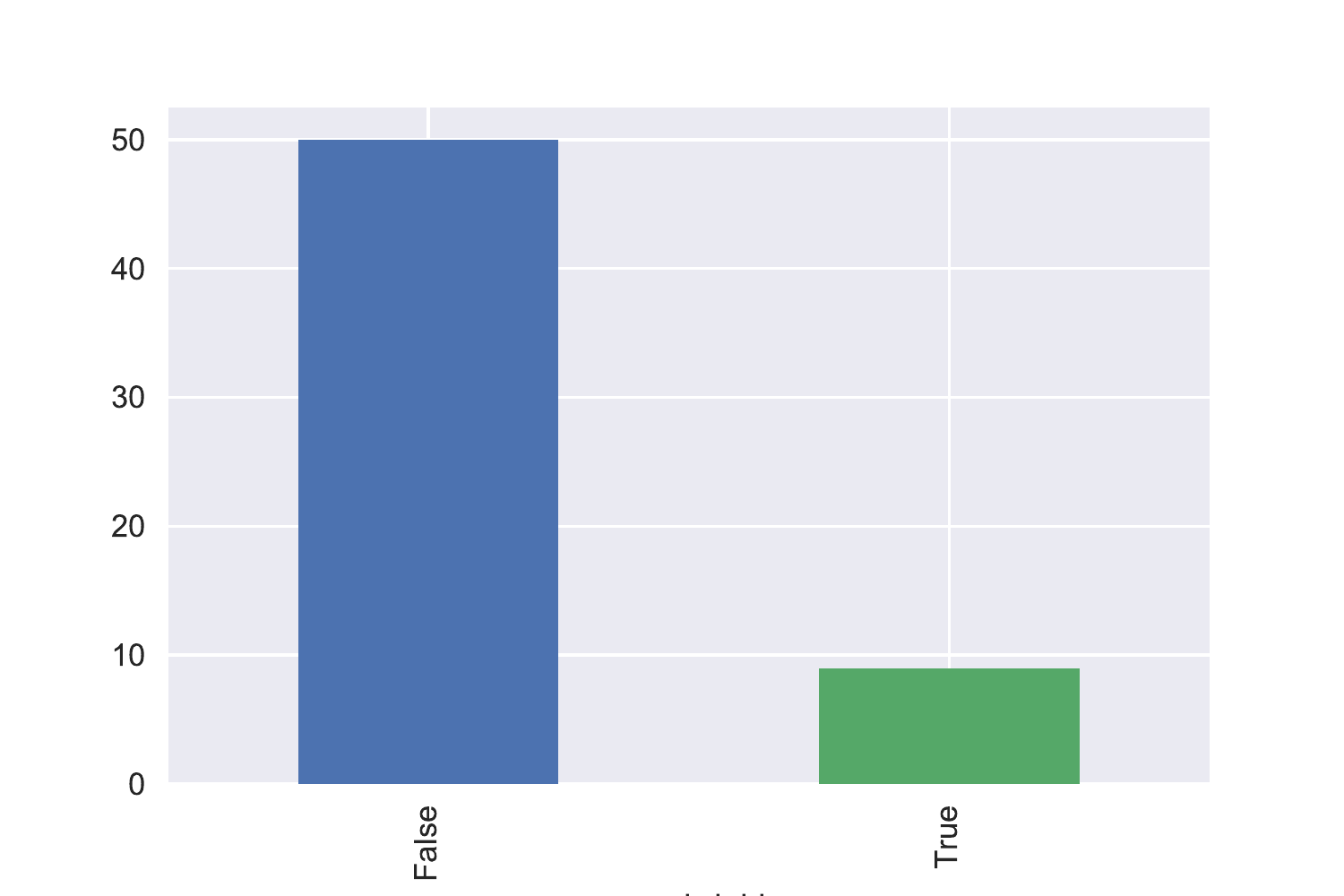}
    \caption{SVM - \texttt{shrinking}}\label{fig:histshrinkinguac}
\end{subfigure}
~
\begin{subfigure}[t]{.3\linewidth}
    \centering
    \includegraphics[width=\textwidth]{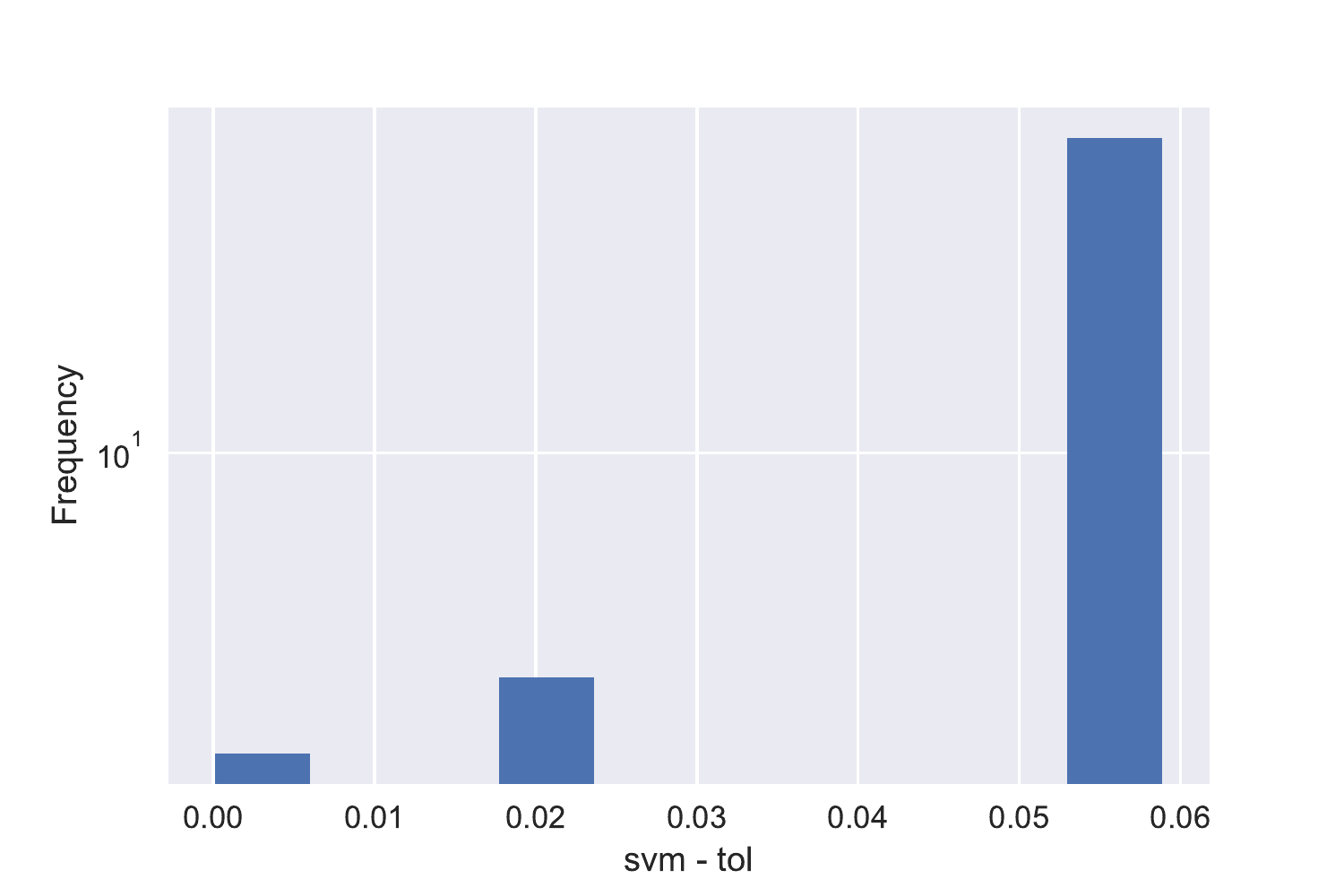}
    \caption{SVM - \texttt{tol}}\label{fig:histtolauc}
\end{subfigure}
\caption{ The distribution of the computed default values, using \textbf{AUC} based performance data, across different tasks for hyperparameters where the default values have a non-zero standard deviation.}
\label{fig:distrdefvalauc}
\end{figure}

The distributions of the computed default values for all hyperparameters for which the standard deviation of the default values is non-zero is depicted in Figure~\ref{fig:distrdefvalacc} and Figure~\ref{fig:distrdefvalauc} derived from respectively accuracy and AUC-based performance data.

\clearpage
\section{Average accuracy}
\label{app:accuracy}
\begin{figure}[ht]
\captionsetup[subfigure]{justification=centering}
\centering
    \begin{subfigure}{.3\linewidth}
        \centering
        \includegraphics[width=\textwidth]{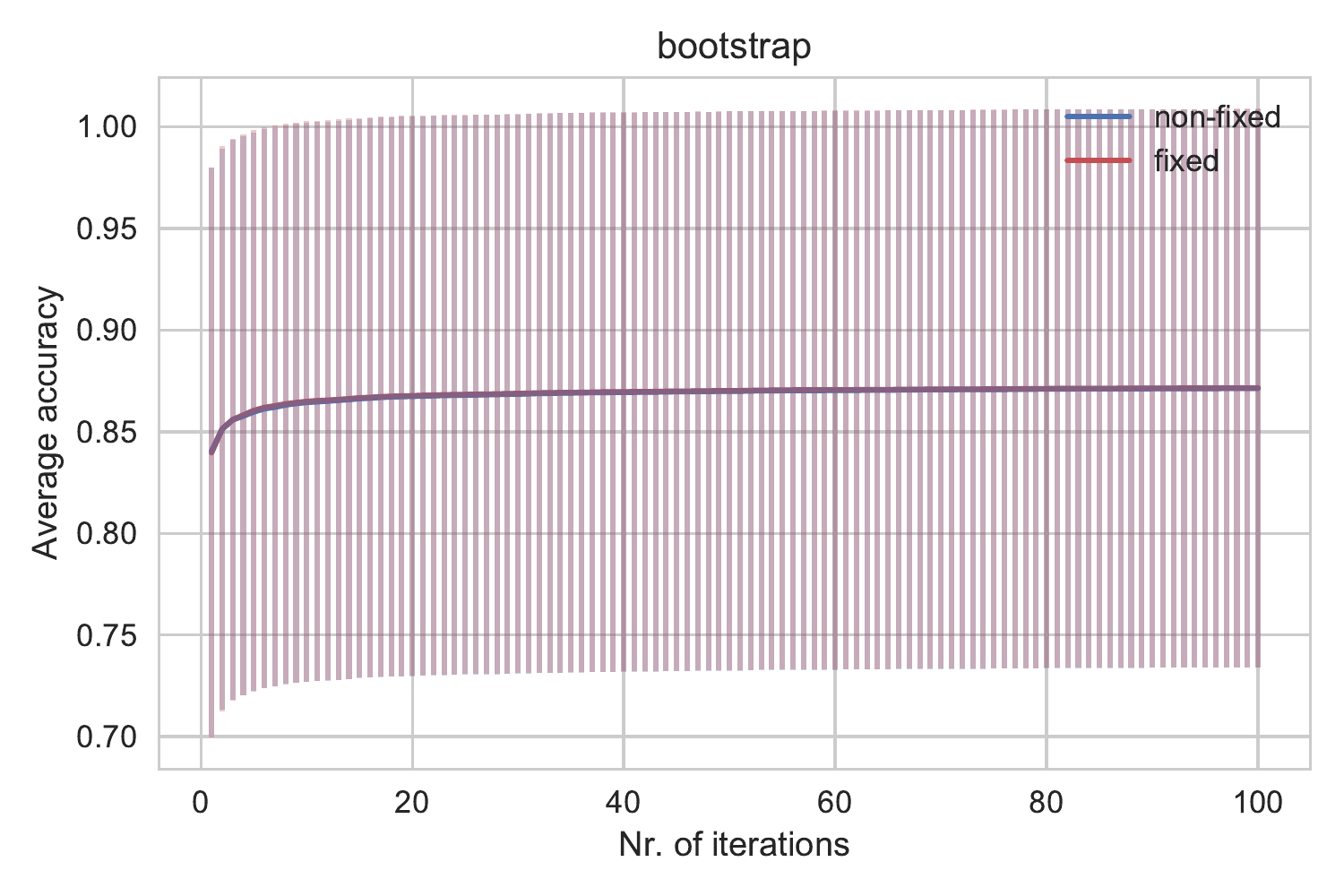}
        \caption{Random Forest - \texttt{bootstrap}}\label{fig:bootstrapaccuracy}
    \end{subfigure}
        \hfill
    \begin{subfigure}{.3\linewidth}
        \centering
        \includegraphics[width=\textwidth]{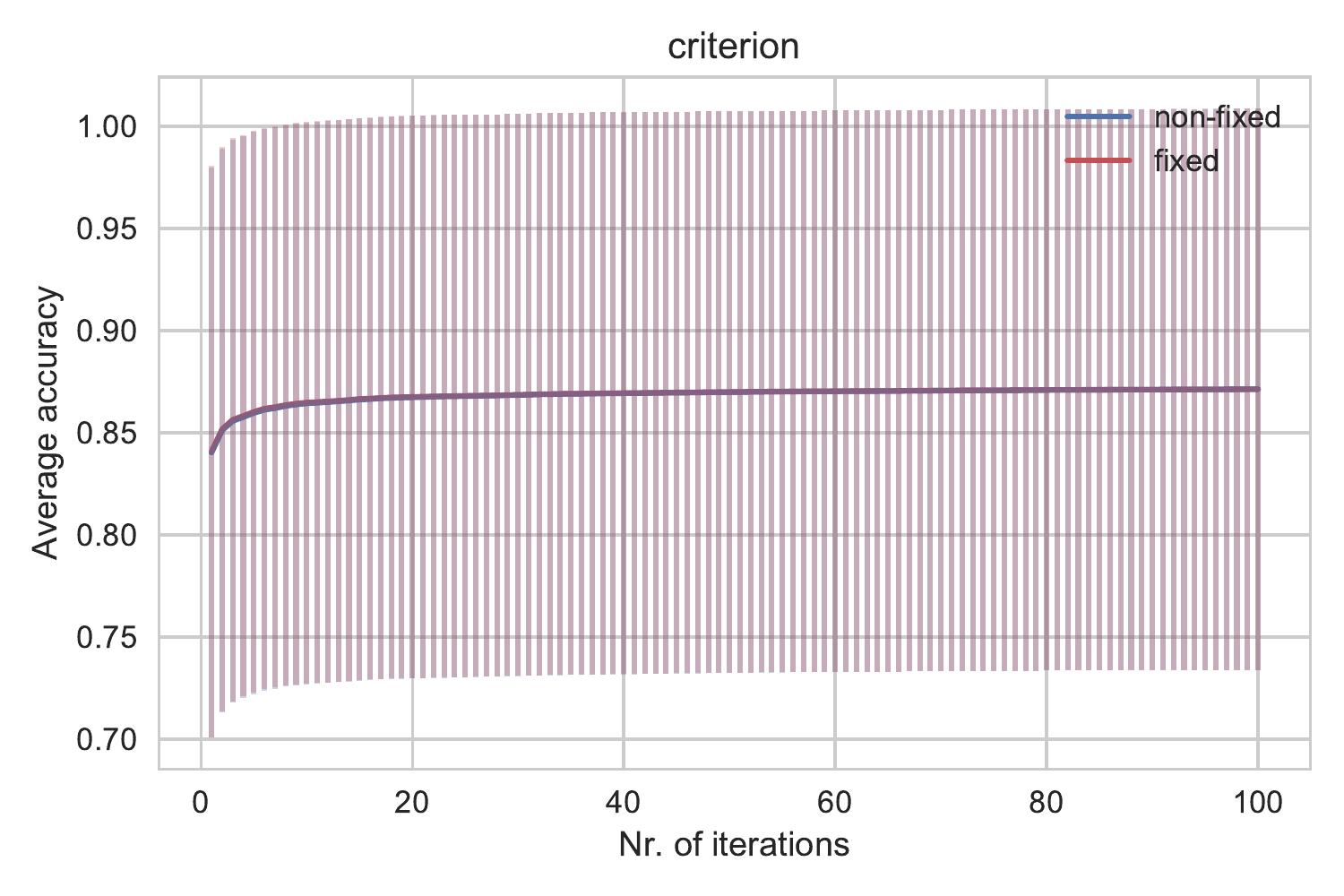}
        \caption{Random Forest - \texttt{criterion}}\label{fig:criterionaccuracy}
    \end{subfigure}
       \hfill
    \begin{subfigure}{.3\linewidth}
        \centering
        \includegraphics[width=\textwidth]{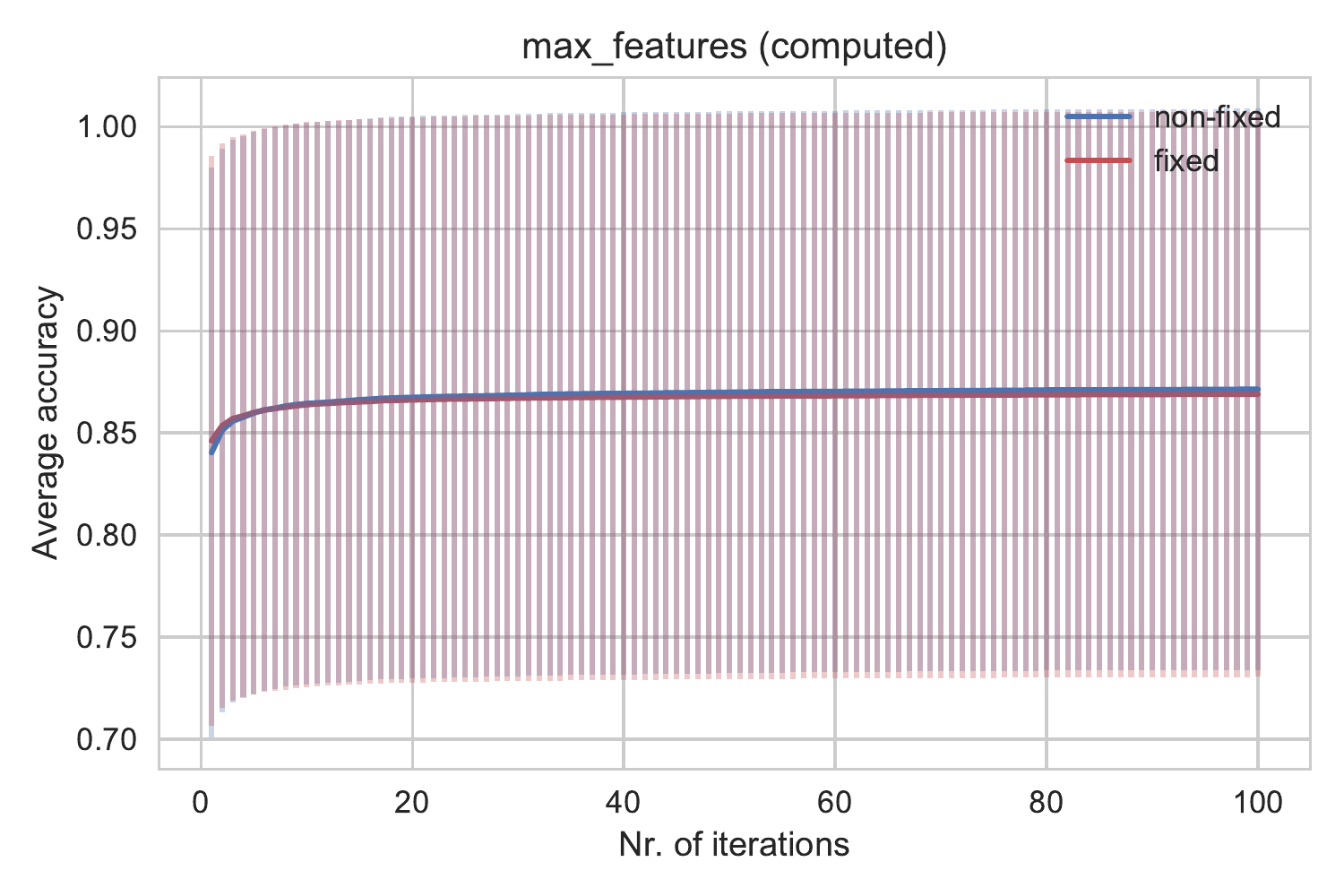}
        \caption{Random Forest - \texttt{max\_features} (computed default)}\label{fig:maxfeaturesaccuracy}
    \end{subfigure}

\bigskip
    
    \begin{subfigure}{.3\linewidth}
        \centering
        \includegraphics[width=\textwidth]{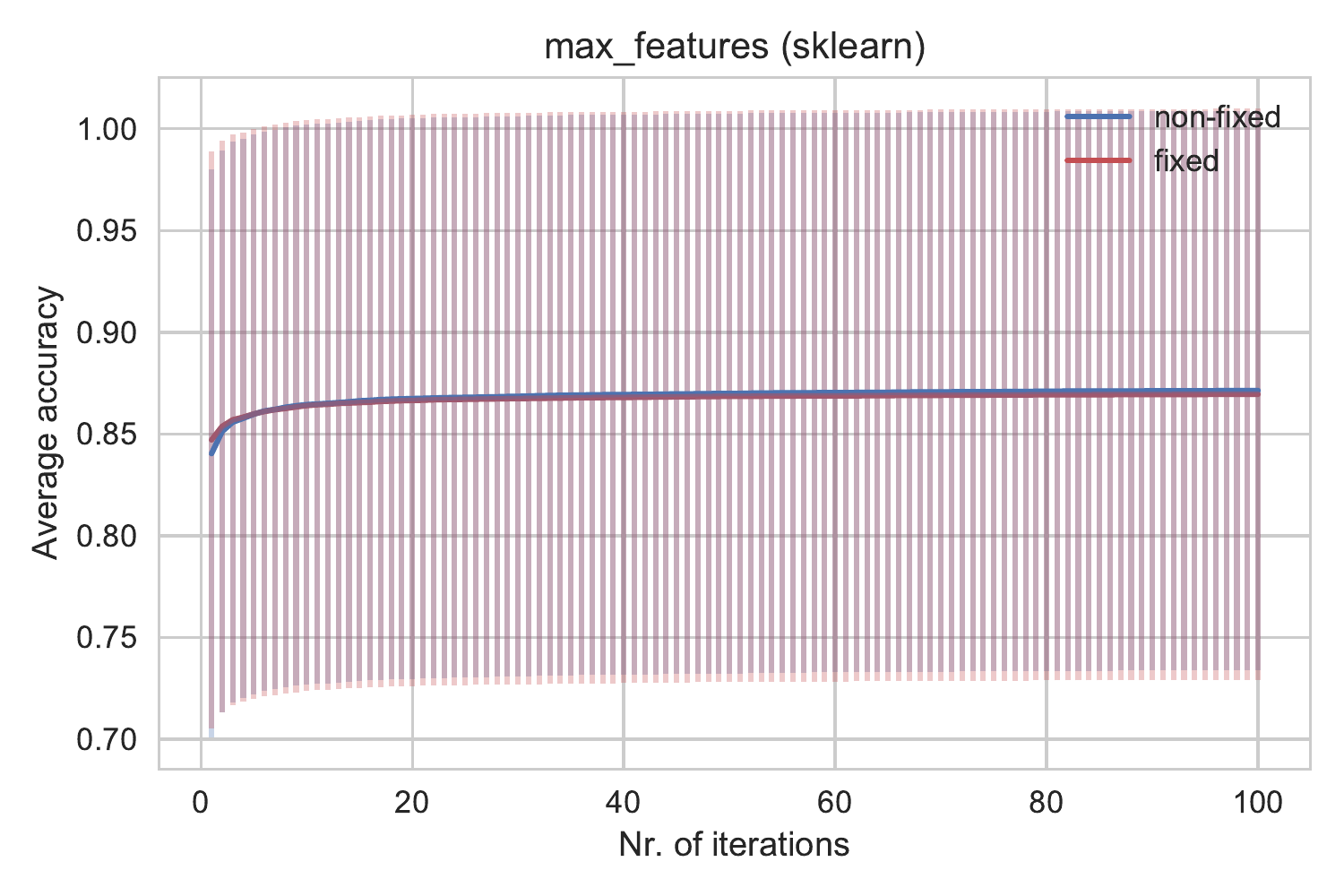}
        \caption{Random Forest - \texttt{max\_features} (scikit-learn default)}\label{fig:maxfeaturesaccuracyklearn}
    \end{subfigure}
\hfill
    \begin{subfigure}{.3\linewidth}
        \centering
        \includegraphics[width=\textwidth]{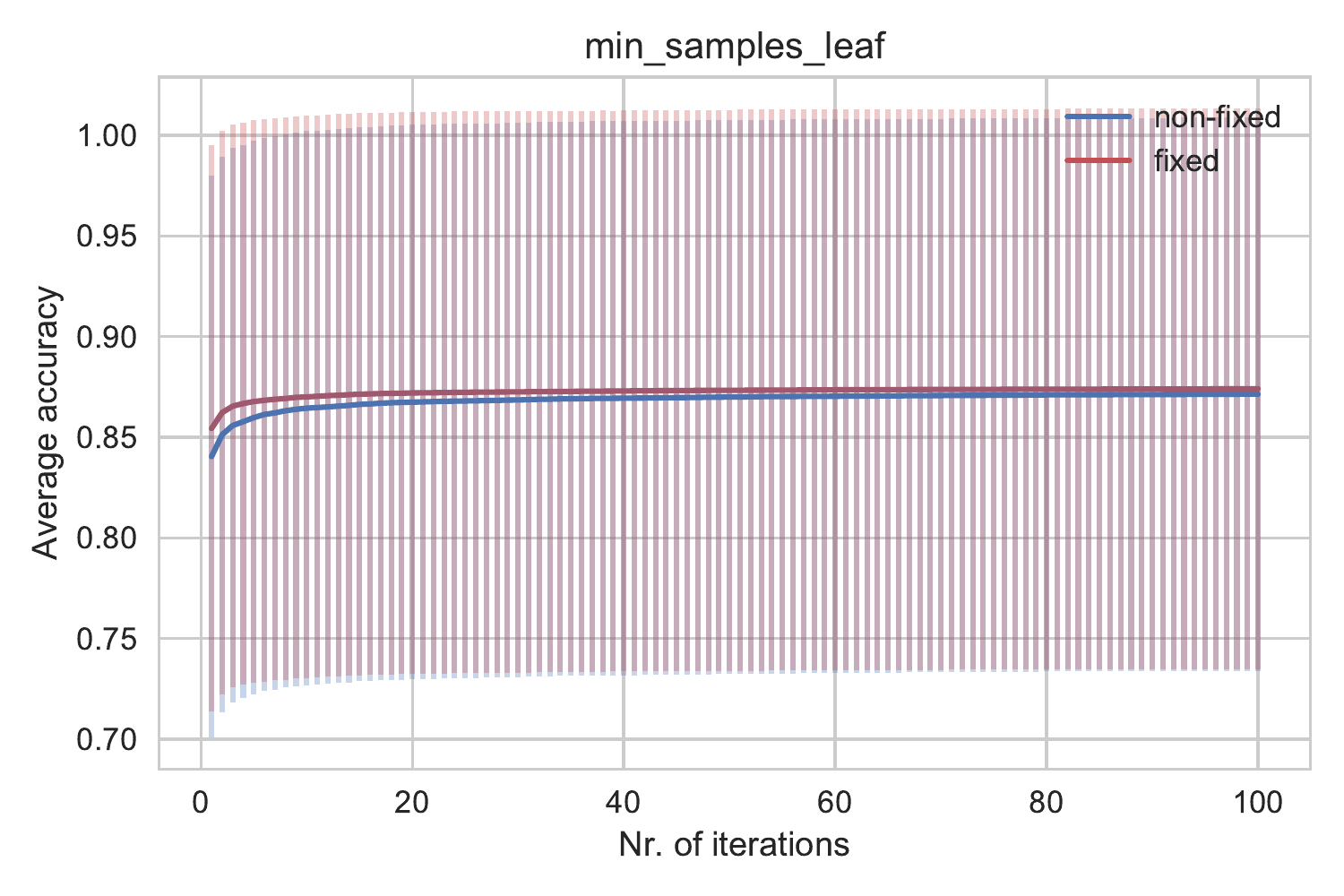}
        \caption{ Random Forest - \texttt{min\_samples\_leaf}}\label{fig:minsamplesleafaccuracy}
    \end{subfigure}
\hfill
    \begin{subfigure}{.3\linewidth}
        \centering
        \includegraphics[width=\textwidth]{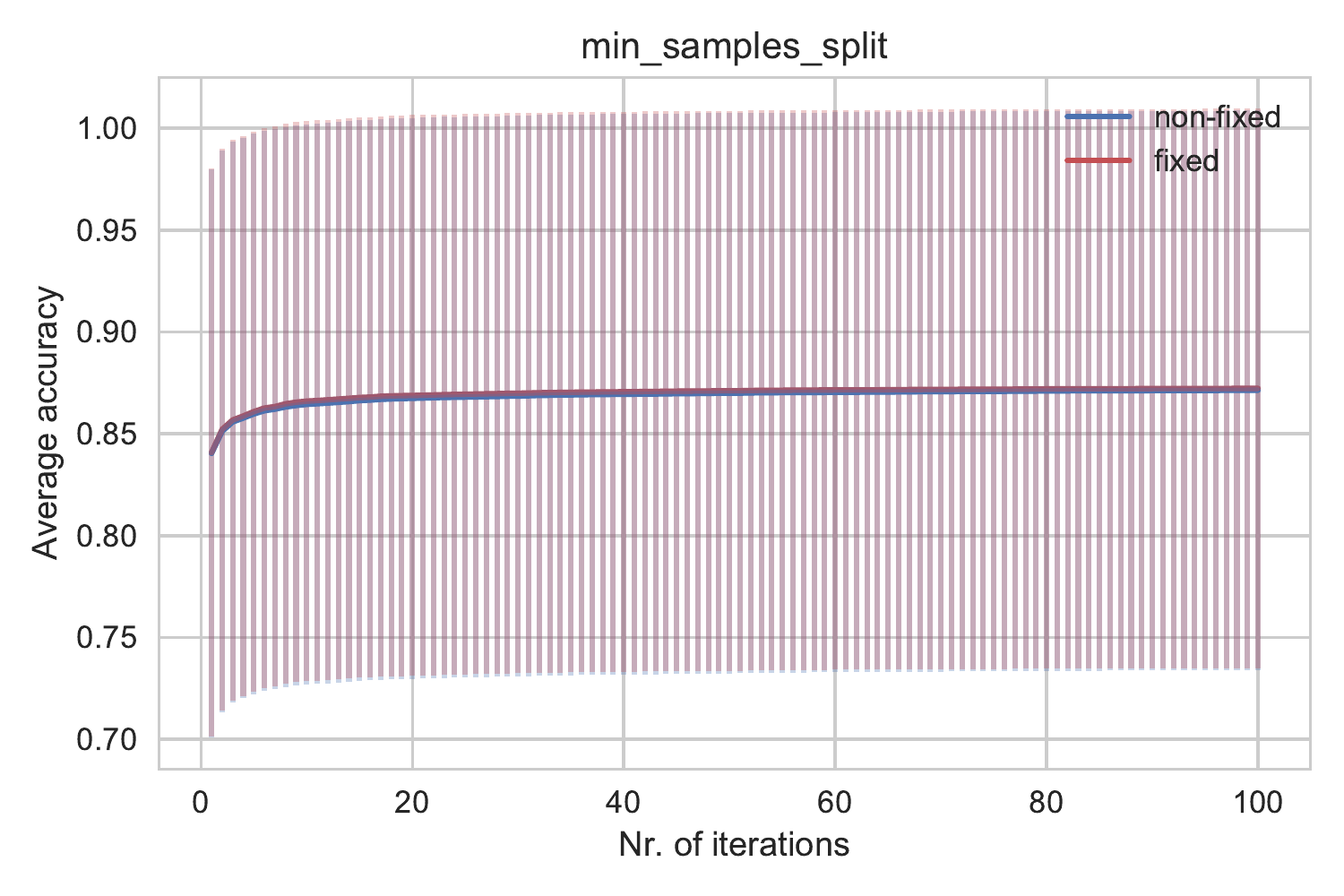}
        \caption{Random Forest - \texttt{min\_samples\_split}}\label{fig:minsamplessplitaccuracy}
    \end{subfigure}
    
\bigskip

    \begin{subfigure}{.3\linewidth}
        \centering
        \includegraphics[width=\textwidth]{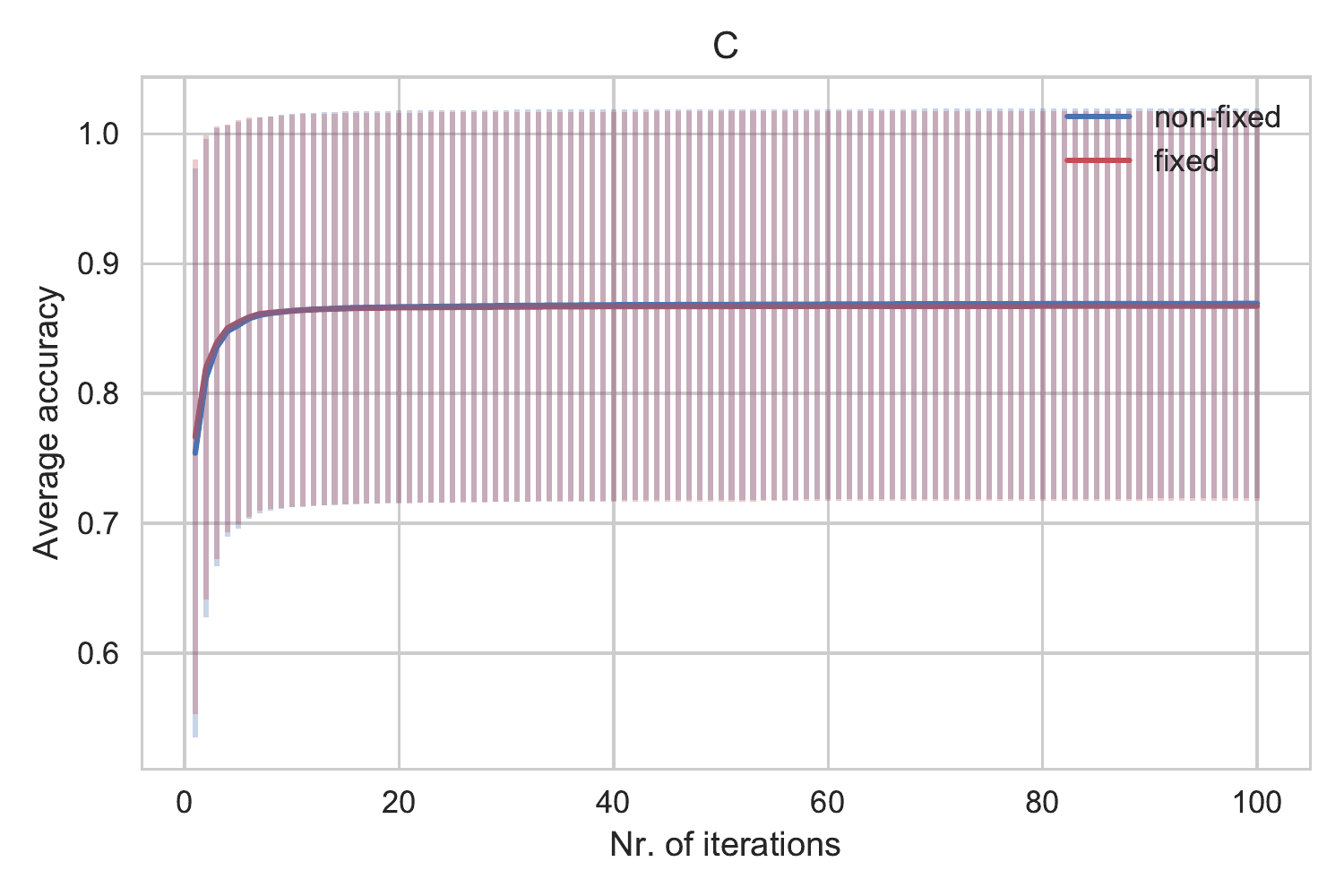}
        \caption{SVM - \texttt{C}}\label{fig:caccuracy}
    \end{subfigure}
\hfill
    \begin{subfigure}{.3\linewidth}
        \centering
        \includegraphics[width=\textwidth]{figures/accuracy__gamma__hyperimp.pdf}
        \caption{SVM - \texttt{gamma} (computed default)}\label{fig:gammaaccuracy}
    \end{subfigure}
\hfill
    \begin{subfigure}{.3\linewidth}
        \centering
        \includegraphics[width=\textwidth]{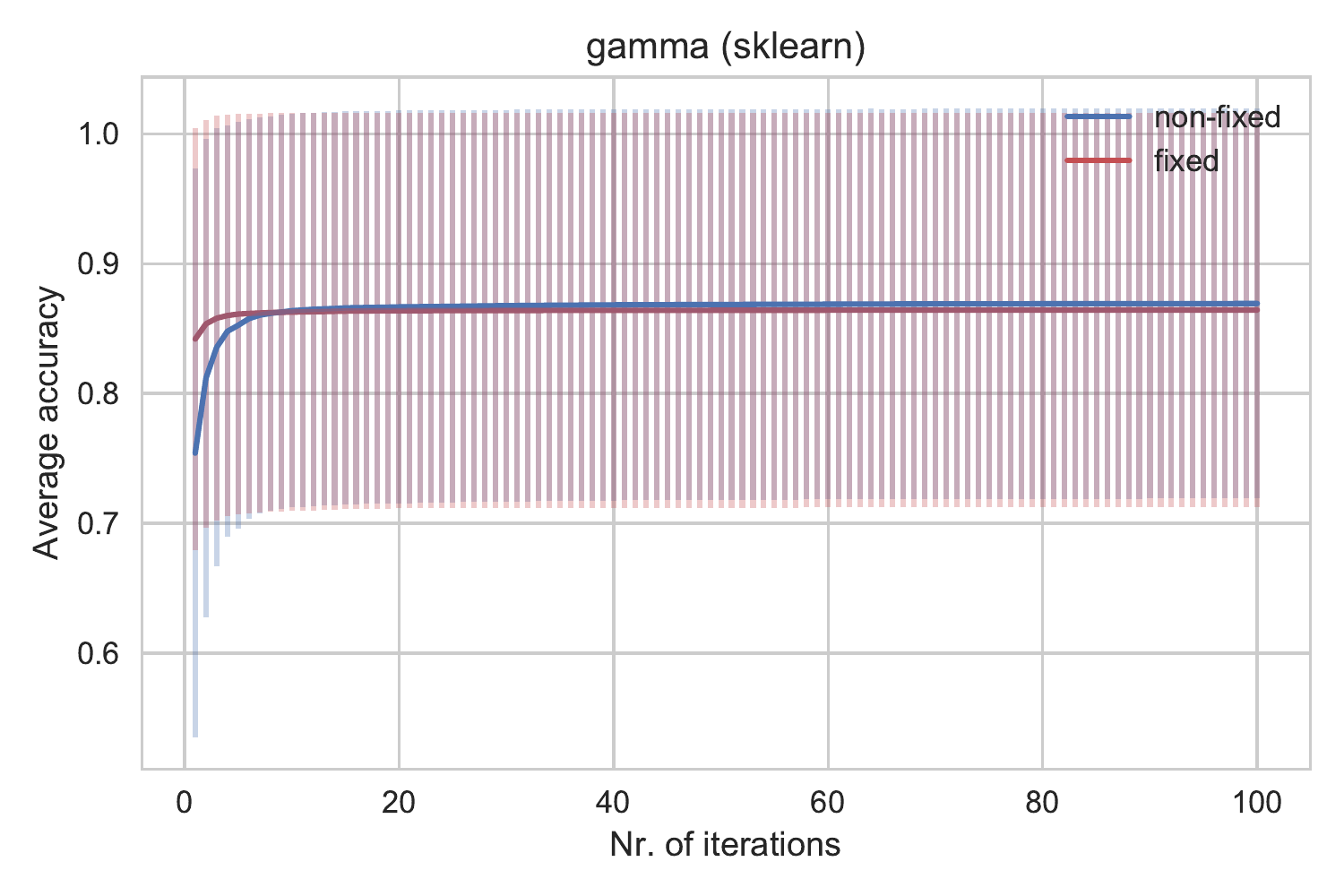}
        \caption{SVM - \texttt{gamma} (scikit-learn default)}\label{fig:gammaaccuracyklearn}
    \end{subfigure}
    
\bigskip
    \begin{subfigure}{.3\linewidth}
        \centering
        \includegraphics[width=\textwidth]{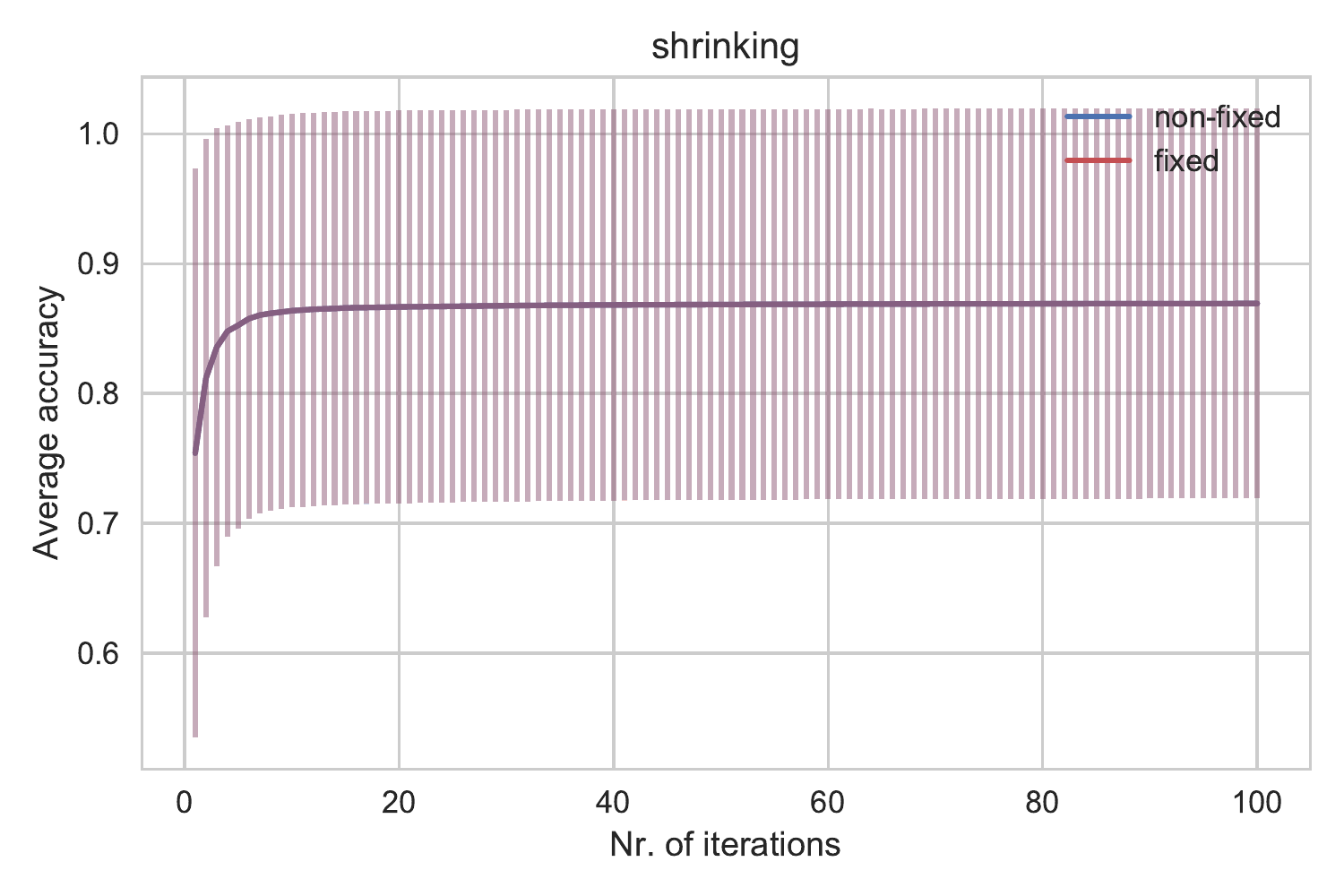}
        \caption{SVM - \texttt{shrinking}}\label{fig:shrinkingaccuracy}
    \end{subfigure}
~
\begin{subfigure}{.3\linewidth}
    \centering
    \includegraphics[width=\textwidth]{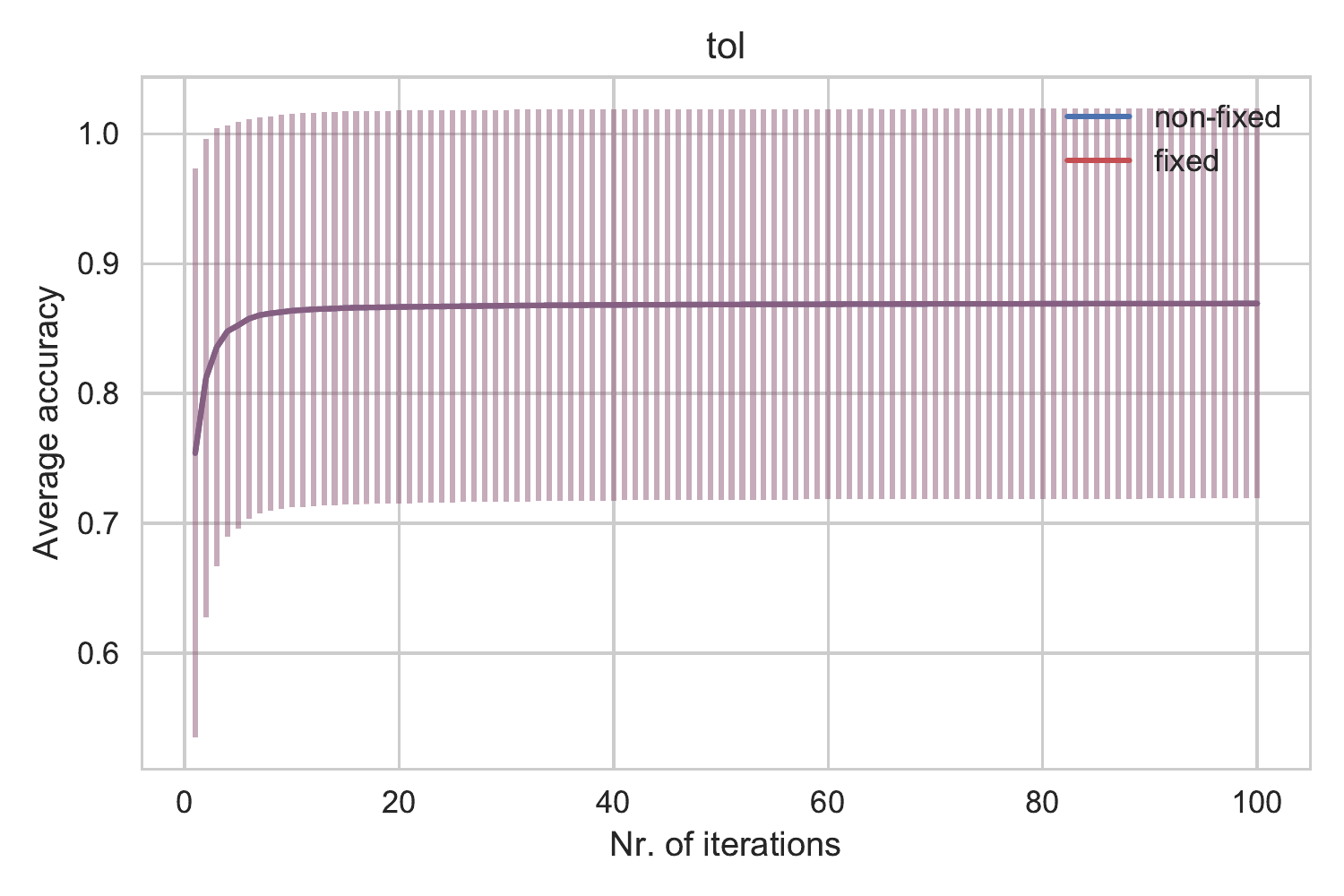}
    \caption{SVM - \texttt{tol}}\label{fig:tolaccuracy}
\end{subfigure}

\caption{ Average accuracy (+/- standard deviation) of the the fixed and non-fixed conditions across 59 datasets over number of iterations.}
\label{fig:accuracys}
\end{figure}

For each measurement, we computed the maximum average validation set accuracy up until a certain iteration. The average accuracy (+/- standard deviation) for both the fixed and non-fixed conditions is shown in Figure~\ref{fig:accuracys}.

\clearpage
\section{Average rank}
\label{app:ranks}
\begin{figure}[ht]
\captionsetup[subfigure]{justification=centering}
\centering
    \begin{subfigure}{.3\linewidth}
        \centering
        \includegraphics[width=\textwidth]{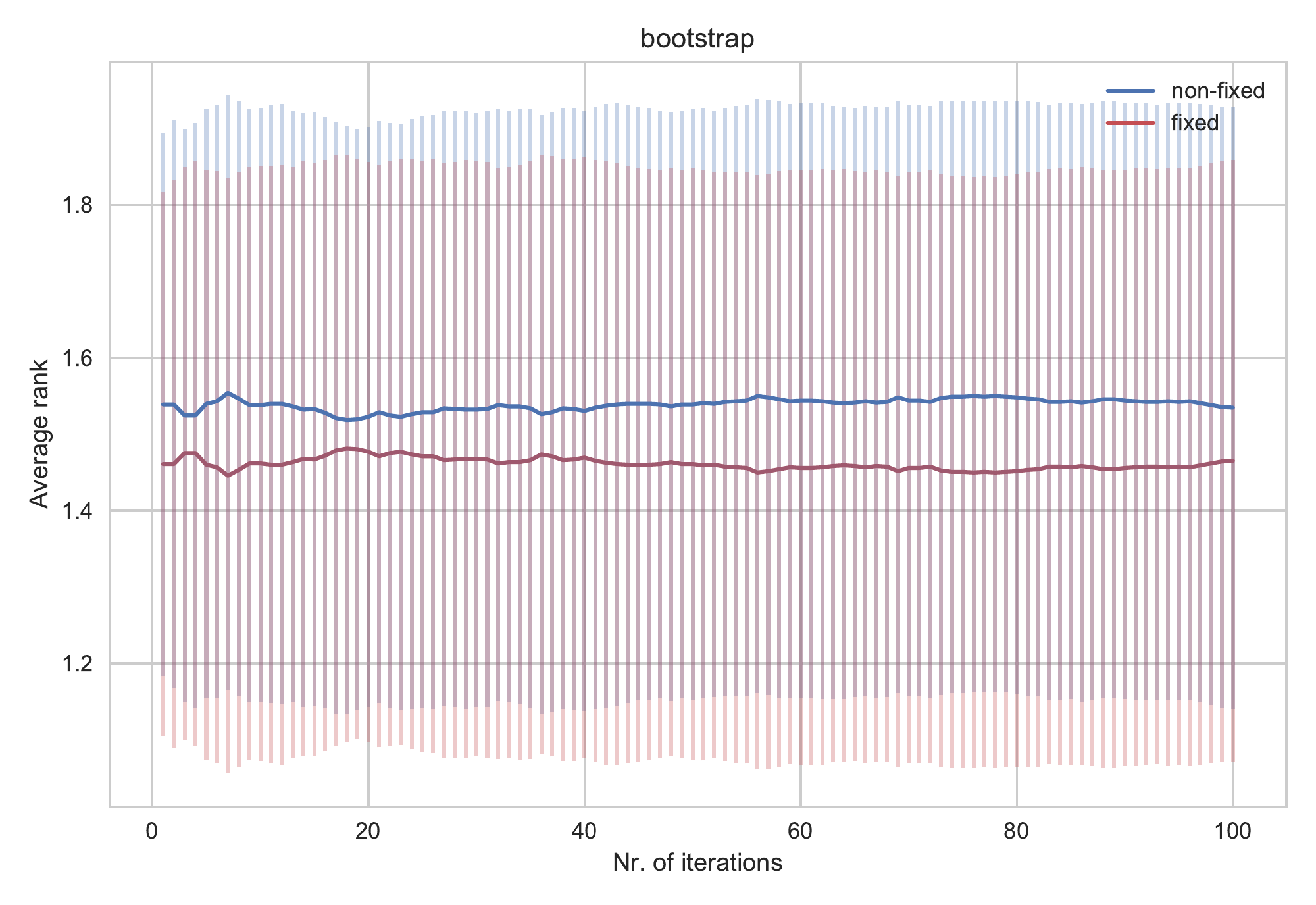}
        \caption{Random Forest - \texttt{bootstrap}}\label{fig:bootstraprank}
    \end{subfigure}
        \hfill
    \begin{subfigure}{.3\linewidth}
        \centering
        \includegraphics[width=\textwidth]{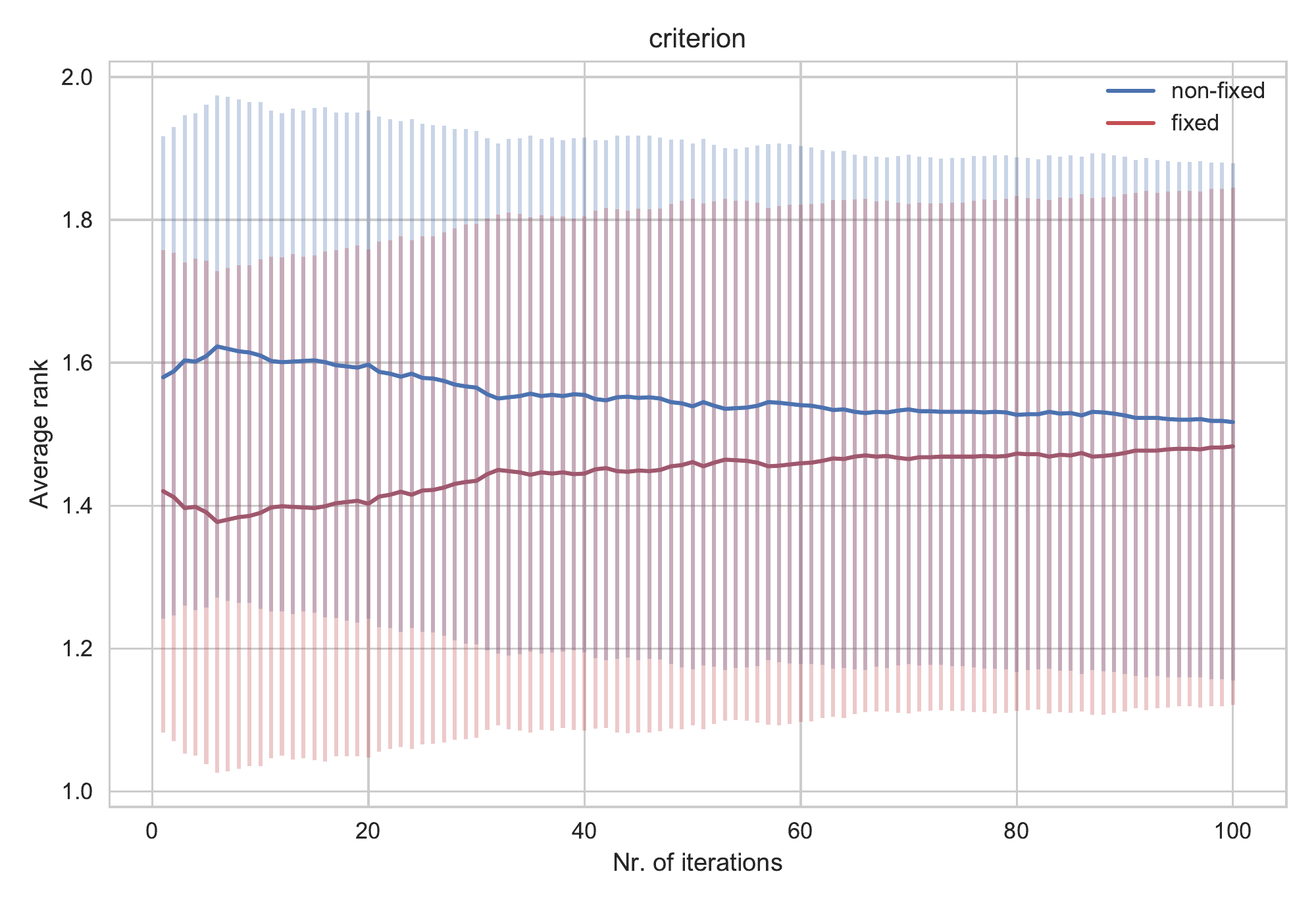}
        \caption{Random Forest - \texttt{criterion}}\label{fig:criterionrank}
    \end{subfigure}
       \hfill
    \begin{subfigure}{.3\linewidth}
        \centering
        \includegraphics[width=\textwidth]{figures/ranks__max_features__hyperimp.pdf}
        \caption{Random Forest - \texttt{max\_features} (computed default)}\label{fig:maxfeaturesrank}
    \end{subfigure}

\bigskip
    
    \begin{subfigure}{.3\linewidth}
        \centering
        \includegraphics[width=\textwidth]{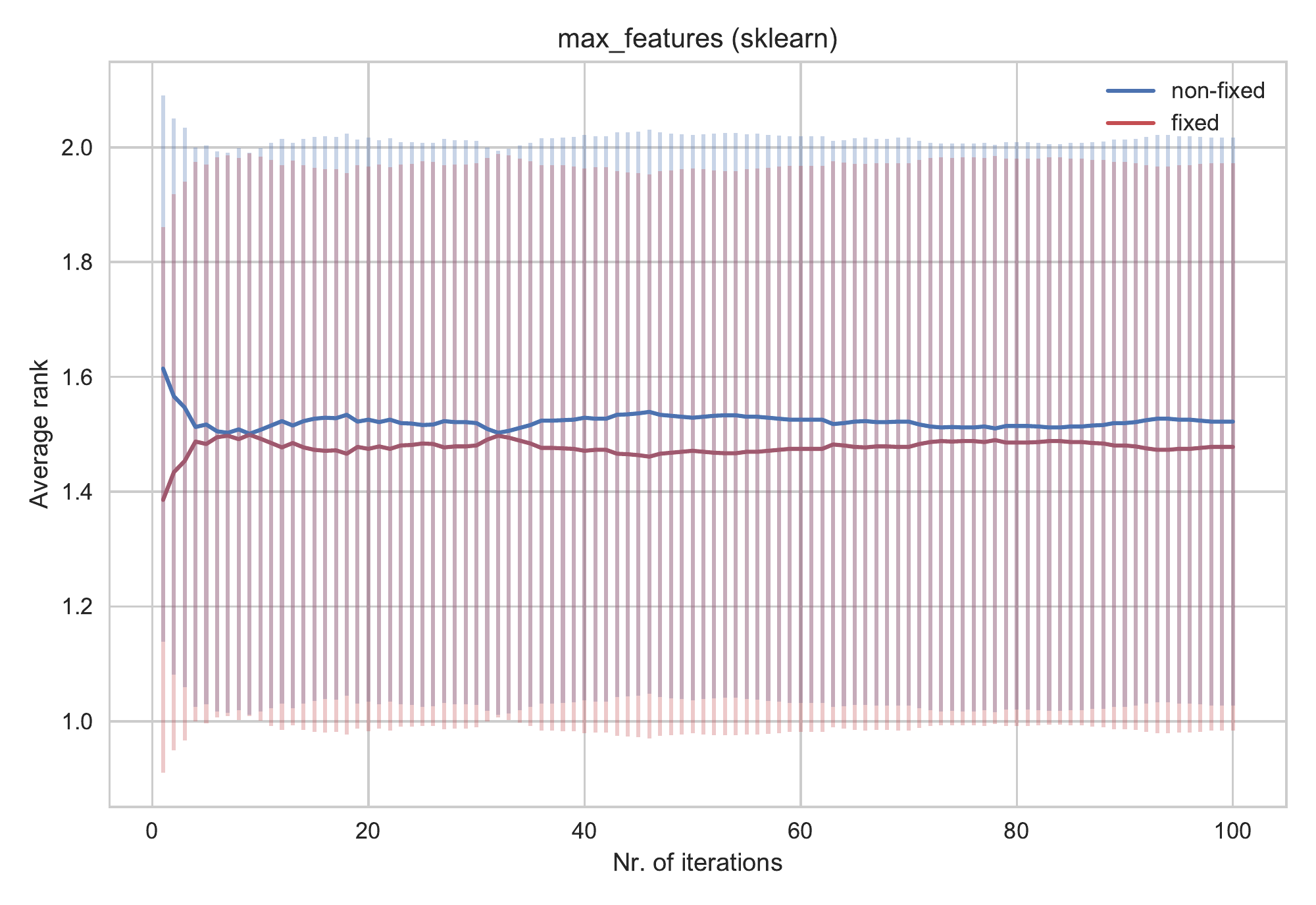}
        \caption{Random Forest - \texttt{max\_features} (scikit-learn default)}\label{fig:maxfeaturesranksklearn}
    \end{subfigure}
\hfill
    \begin{subfigure}{.3\linewidth}
        \centering
        \includegraphics[width=\textwidth]{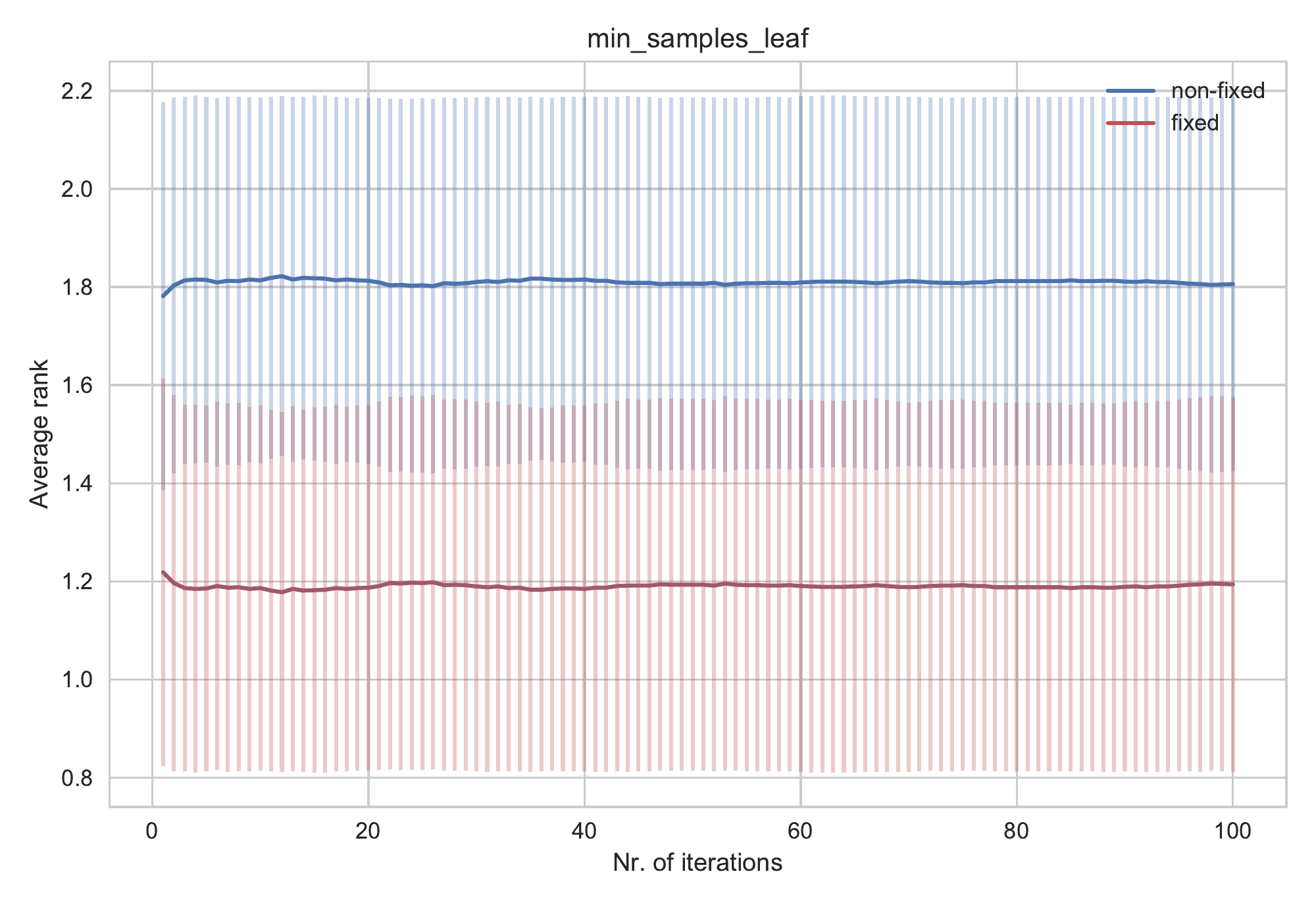}
        \caption{ Random Forest - \texttt{min\_samples\_leaf}}\label{fig:minsamplesleafranks}
    \end{subfigure}
\hfill
    \begin{subfigure}{.3\linewidth}
        \centering
        \includegraphics[width=\textwidth]{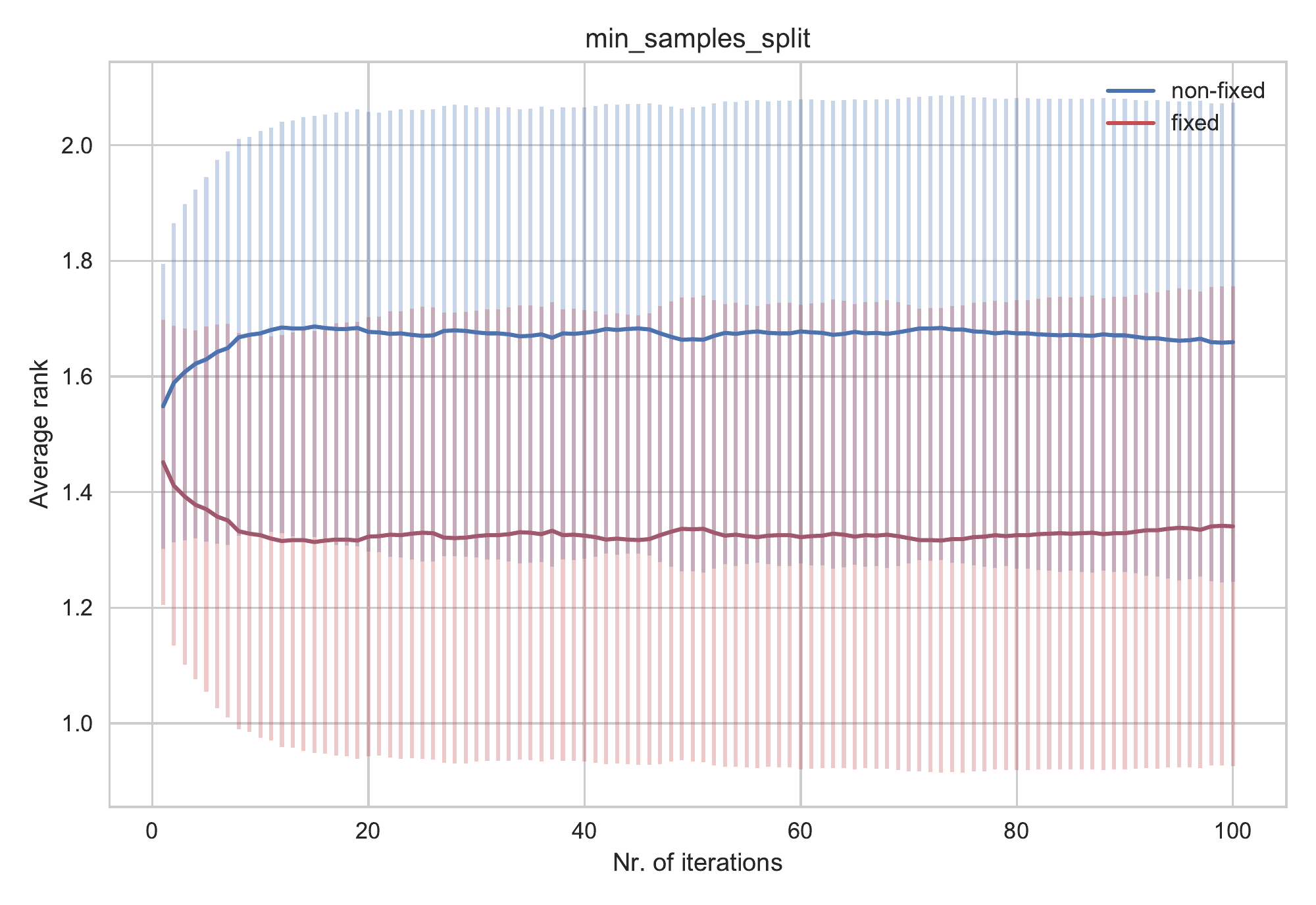}
        \caption{Random Forest - \texttt{min\_samples\_split}}\label{fig:minsamplessplitrank}
    \end{subfigure}
    
\bigskip

    \begin{subfigure}{.3\linewidth}
        \centering
        \includegraphics[width=\textwidth]{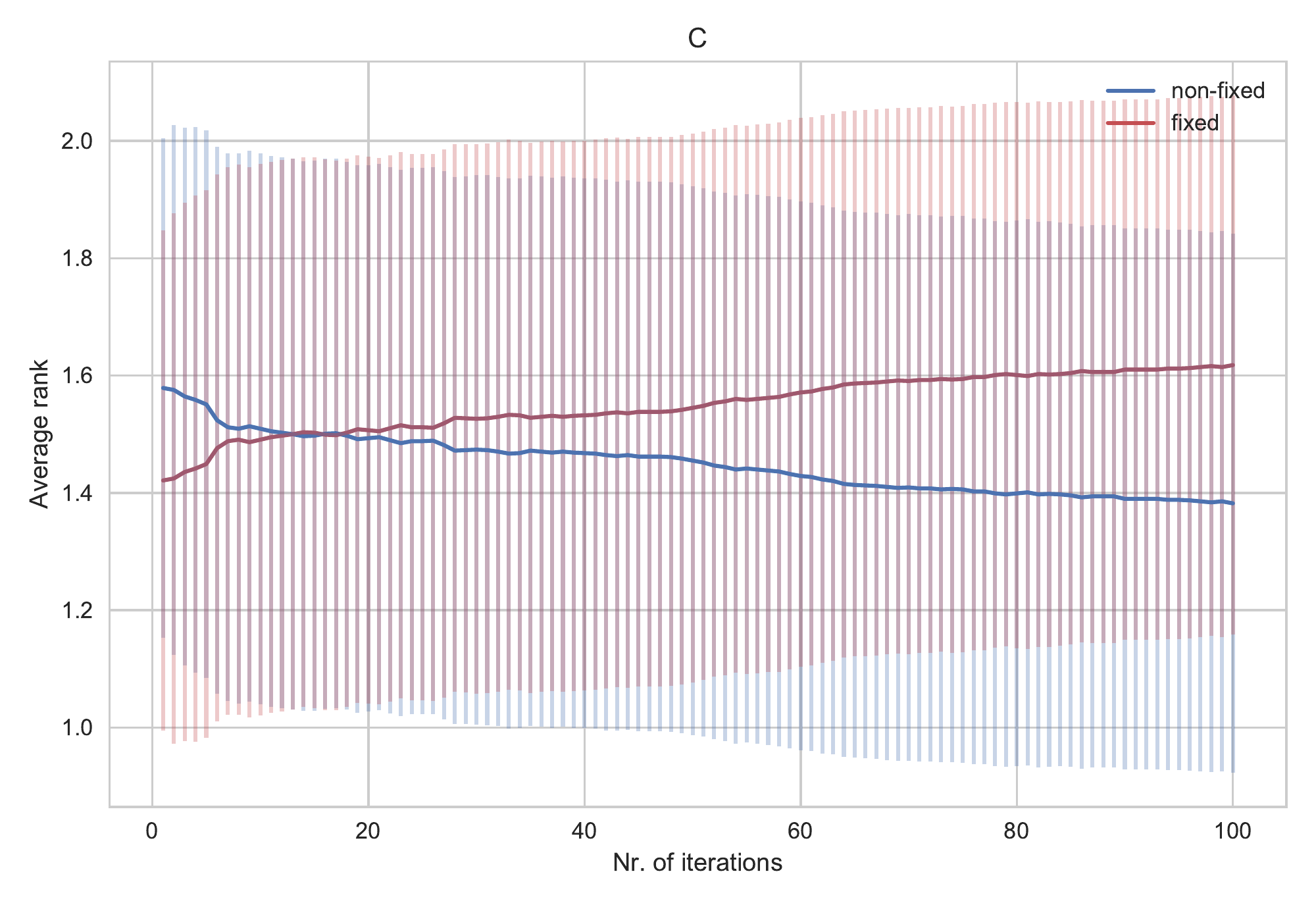}
        \caption{SVM - \texttt{C}}\label{fig:crank}
    \end{subfigure}
\hfill
    \begin{subfigure}{.3\linewidth}
        \centering
        \includegraphics[width=\textwidth]{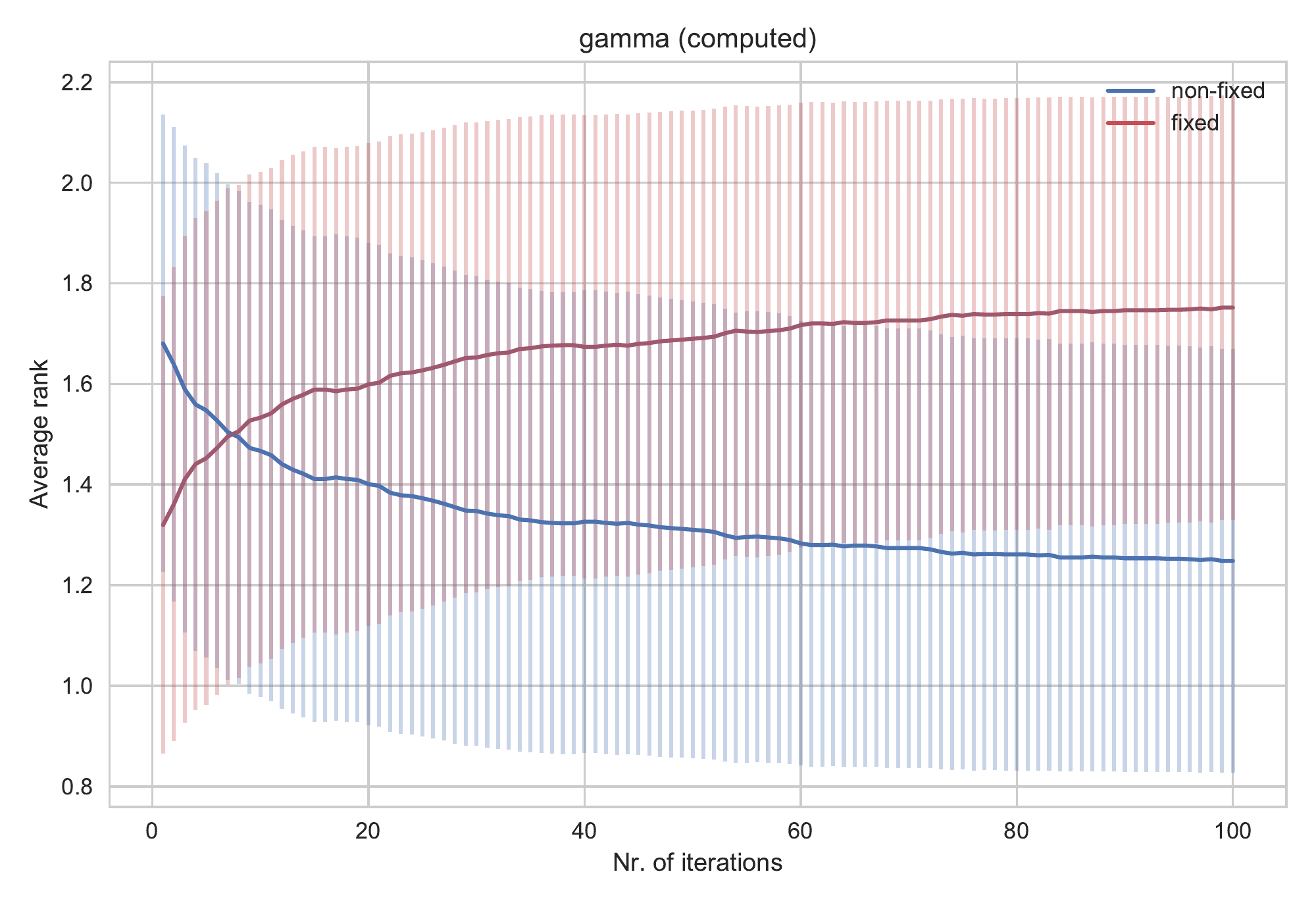}
        \caption{SVM - \texttt{gamma} (computed default)}\label{fig:gammarank}
    \end{subfigure}
\hfill
    \begin{subfigure}{.3\linewidth}
        \centering
        \includegraphics[width=\textwidth]{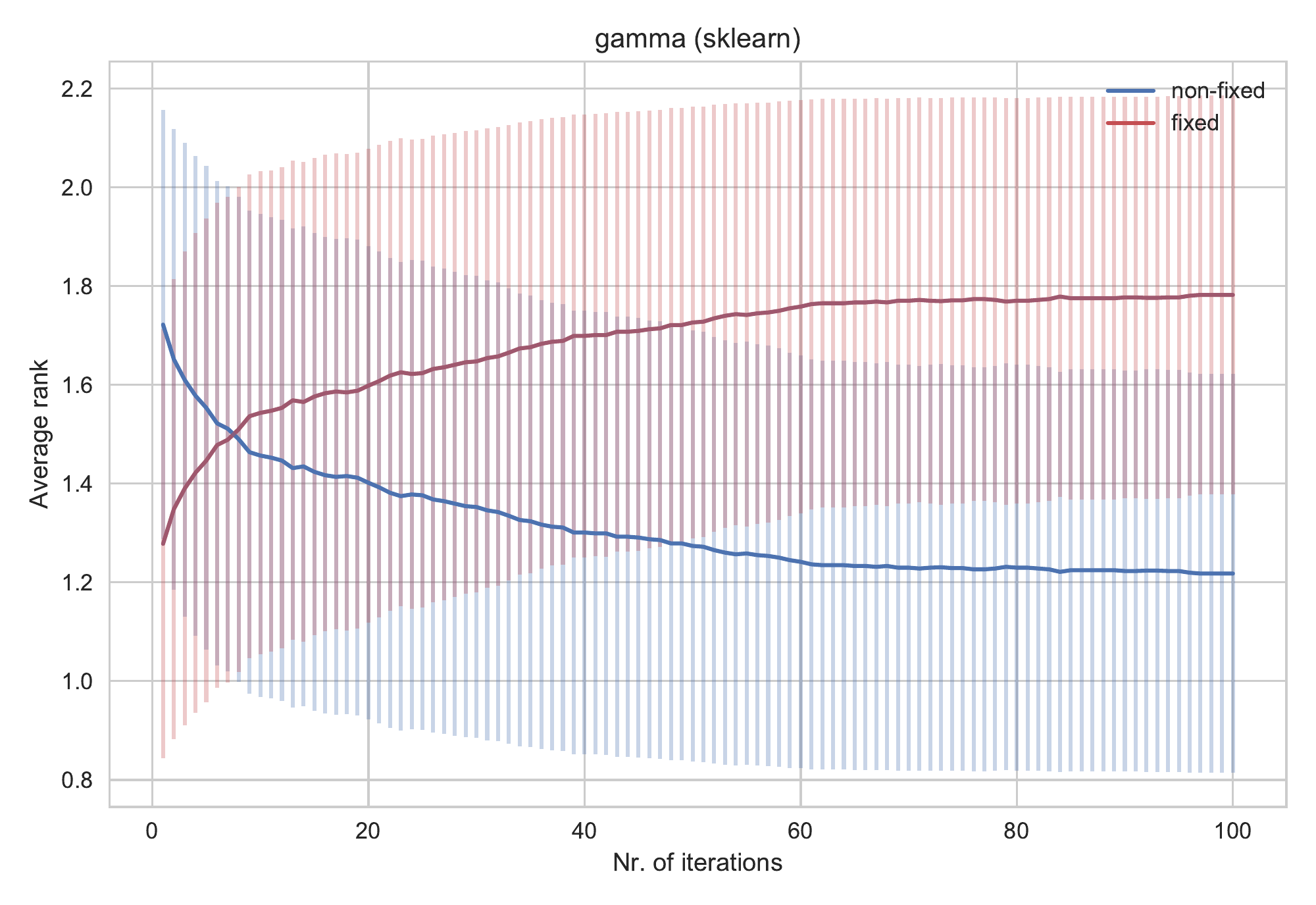}
        \caption{SVM - \texttt{gamma} (scikit-learn default)}\label{fig:gammaranksklearn}
    \end{subfigure}
    
\bigskip
    \begin{subfigure}{.3\linewidth}
        \centering
        \includegraphics[width=\textwidth]{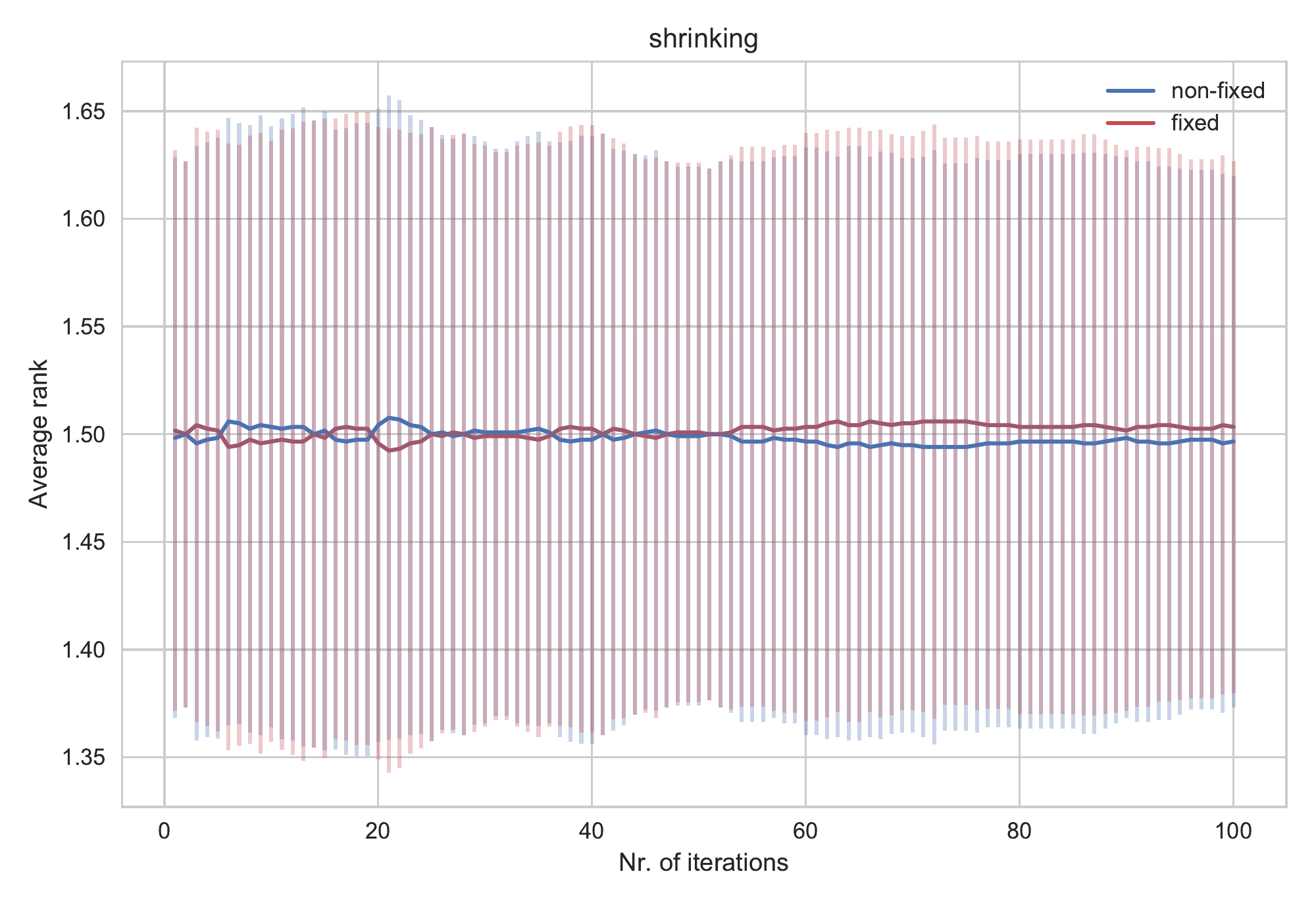}
        \caption{SVM - \texttt{shrinking}}\label{fig:shrinkingrank}
    \end{subfigure}
~
\begin{subfigure}{.3\linewidth}
    \centering
    \includegraphics[width=\textwidth]{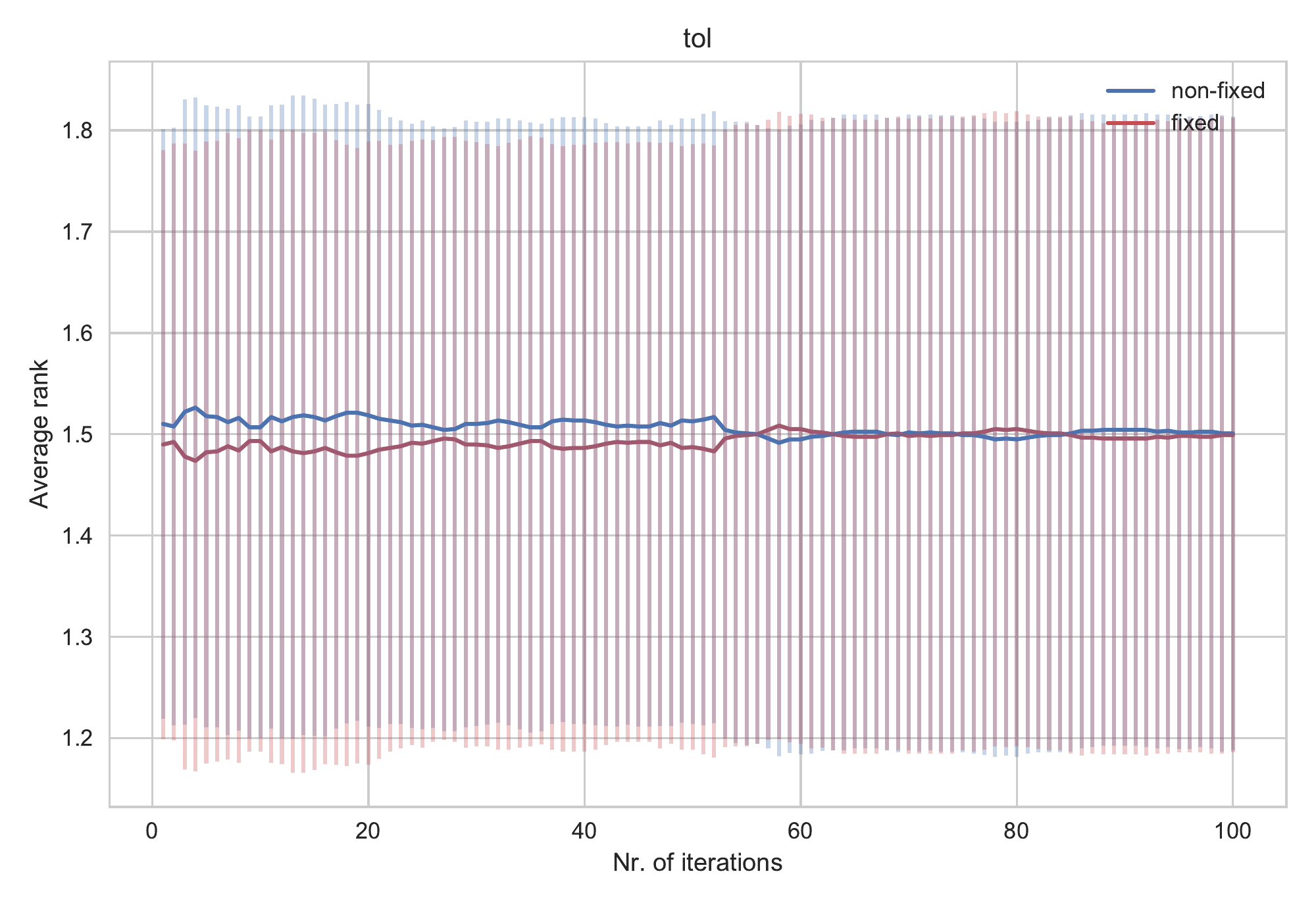}
    \caption{SVM - \texttt{tol}}\label{fig:tolrank}
\end{subfigure}

\caption{ Average rank (+/- standard deviation) of the the fixed and non-fixed conditions across 59 datasets over number of iterations. Ranks are based on the maximum average validation set accuracy. Note that ranks are a relative measure of performance and that smaller is better.}
\label{fig:ranks}
\end{figure}

For each measurement, we computed the maximum average validation set accuracy up until a certain iteration. This data was used to rank the fixed and non-fixed conditions. The average rank (+/- standard deviation) is shown in Figure~\ref{fig:ranks}.

\end{document}